\documentclass[twoside]{article}

\usepackage[accepted]{aistats2023}
\usepackage{amsfonts}
\usepackage{bm}
\usepackage{amsmath}
\usepackage{float}
\usepackage{comment}
\usepackage{multirow}
\usepackage[export]{adjustbox}

\usepackage{graphicx}
\usepackage{hyperref}       % hyperlinks
\usepackage{booktabs}       % professional-quality tables
\usepackage{float}
\usepackage{subcaption}
\usepackage{tabularx}
\usepackage{xcolor}         % colors
\usepackage[round]{natbib}

\newcommand{\ourtitle}{Energy-Based Models for Functional Data using Path Measure Tilting}

\begin{document}

% If your paper is accepted and the title of your paper is very long,
% the style will print as headings an error message. Use the following
% command to supply a shorter title of your paper so that it can be
% used as headings.
%
%\runningtitle{I use this title instead because the last one was very long}

% If your paper is accepted and the number of authors is large, the
% style will print as headings an error message. Use the following
% command to supply a shorter version of the authors names so that
% they can be used as headings (for example, use only the surnames)
%
%\runningauthor{Surname 1, Surname 2, Surname 3, ...., Surname n}

\twocolumn[

\aistatstitle{\ourtitle}

\aistatsauthor{Jen Ning Lim \And Sebastian J. Vollmer \And Lorenz Wolf \And  Andrew Duncan}

\aistatsaddress{University of Warwick \And RPTU of Kaiserslautern, DFKI \And  University College London \And Imperial College London}]

\begin{abstract}
Energy-Based Models (EBMs) have proven to be a highly effective approach for modelling densities on finite-dimensional spaces.  Their ability to incorporate domain-specific choices and constraints into the structure of the model through composition make EBMs an appealing candidate for applications in physics, biology and computer vision and various other fields. Recently, Energy-Based Processes (EBP) for modelling stochastic processes was proposed for \textit{unconditional} exchangeable data (e.g., point clouds).
In this work, we present a novel subclass of EBPs, called $\mathcal{F}$-EBM for \textit{conditional} exchangeable data, which is able to learn distributions of functions (such as curves or surfaces) from functional samples evaluated at finitely many points. Two unique challenges arise in the functional context. Firstly, training data is often not evaluated along a fixed set of points.  Secondly, steps must be taken to control the behaviour of the model between evaluation points, to mitigate overfitting.
The proposed model is an energy based model on function space that is decomposed spectrally, where a Gaussian Process path measure is used to reweight the distribution to capture smoothness properties of the underlying process being modelled. The resulting model has the ability to utilize irregularly sampled training data and can output predictions at any resolution, providing an effective approach to up-scaling functional data.
We demonstrate the efficacy of our proposed approach for modelling a range of datasets, including data collected from Standard and Poor's 500 (S\&P) and UK National grid.
\end{abstract}

\section{INTRODUCTION}

The problem of generative modelling is concerned with learning distributions from samples. This challenge arises naturally in various applications of  machine learning for which a number of well-established methods has been proposed including Variational Autoencoders \citep{kingma2013auto}, Generative Adversarial Networks \citep{gan} and Energy-Based Models (EBM) - which are intimately connected \citep{che2020your}.

Broadly, these methods all assume that the samples are intrinsically finite dimensional. When the underlying data is continuous, generative models are usually trained on a discretization of the data; ultimately, it learns a probability distribution on a finite dimensional space whose dimension scales with the resolution of the data \citep{ramsay1982data}. For example, in the context of images, the energy of an EBM is often defined on the same resolution of the data \citep{NEURIPS2019_378a063b}. A  key distinguishing feature between classical finite dimensional generative models and the proposed formulation is the ability to generate predictions along any mesh and at any resolution.   This naturally leads to important applications, including up-scaling the resolution of functional data, as well as data imputation over irregularly sampled datasets.

In this work, we propose a novel class of generative model for data which is assumed to live within an infinite dimensional space of functions $\mathcal{F}$. In this setting, the data will consist of a finite set of function discretizations; each evaluated over a finite set of points. Based on this data, we seek to learn a generative model for the associated distribution  over the  function space $\mathcal{F}$ directly. The functional data context poses unique challenges. Firstly, it is often the case that the evaluation points along which the functions are discretized is not uniform across samples, i.e., each function in the sample can be evaluated over a different set of points.  We address this challenge by constructing a likelihood which can accommodate observations along irregular meshes.  The second challenge relates to how to inform the characteristics of the learned functional distribution between evaluation points. For this, we use a Gaussian process path measure ``tilt'' the model towards functions with certain characteristics. We note that the key difference between functional data (or \textit{conditional} exchangeable data) and \textit{unconditional} exchangeable data (as in \citet{yang2020energy}, e.g., point clouds) is that the evaluation points are available in functional settings, and so do not explicitly share the same challenges.

\begin{comment}
An impediment in the deployment of EBMs on function space is the fact that there is no generalization of the Lebesgue measure on an infinite-dimensional separable Hilbert space \citep{feldman1966nonexistence}.
This implies that there is no canonical reference measure on $\mathcal{F}$ with respect to which a density of the form $\exp(-E(x))$ can be formulated.
Instead, the proposed generative model is assumed to be absolutely continuous with respect to a probability measure associated to the latent Gaussian process taking values in $\mathcal{F}$.  This measure effectively acts as a prior on $\mathcal{F}$,  capturing intrinsic properties of the functional data such as regularity and integrability. This base measure can also be informed by data, either through hyperparameter tuning or through preprocessing.
\end{comment}

In summary, our contributions include introducing a novel class of energy-based models that is fully flexible and well adapted to functional data for generative modelling and regression. We compare our proposed model with other popular models such as the Neural Process, Gaussian Process and Variational Implicit Processes for generative modelling and as well as function inference on a range of synthetic and real-life datasets. Specifically, we evaluate whether each model can accurately interpolate between points, and upscale the underlying function, as well as, its ability to ``trick'' a two-sample test.
We demonstrate the benefits and abilities of our proposal for modelling FashionMNIST \citep{xiao2017/online} as a resolution independent model that can freely upscale and downscale images.

\section{BACKGROUND}

Our proposed methodology builds upon the energy-based modelling (EBM) paradigm \citep{hinton2002training} and is a subclass of energy-based processes (EBP) \citep[Section 3.2]{yang2020energy} that provides a generalization of EBMs to  infinite dimensional spaces.  In the classical setting, EBMs seek to learn a density proportional to $\exp(-E(x))$ over  sample space. The normalization constant  $C = \int \exp(-E(x))\,dx$ (known as the partition function) is typically unknown and must be approximated. The energy function $E(\cdot)$, often parametrized using a neural network, seeks to assign low-energy values to inputs $x$ in the data distribution and high-energy values to others. The flexibility and expressibility of the energy function give rise to several advantages of EBMs over other generative models, such as those based on transformation of noise.  One key advantage is its compositionality property that allows for incorporation of domain-specific knowledge through, e.g., summing up two or more energy functions which represent different goals or constraints \citep{mnih2005learning}. This makes EBMs promising candidates for modelling real life phenomena \citep{Du2020Energybased, matsubara2020deep} and thus have found applications in physics \citep{noe2019boltzmann} and biology \citep{ingraham2019learning} to name a few.

A latent variable EBM has the form $p_\theta (Y, Z) = \frac{1}{C_\theta}\exp(-E_\theta(Y, Z))$ where $Y$ is observed, $Z$ is a latent variable with energy function $E_\theta : \mathcal{Y} \times \mathcal{Z} \rightarrow \mathbb{R}$, and $C_\theta = \int_{\mathcal{Y},\mathcal{Z}} \exp(-E_\theta(y, z)) dydz$ is the normalizing constant. We will focus on these kinds of EBM.

Generating samples from an EBM is itself a challenging problem. Common approaches of generating these samples is via a Markov Chain Monte Calo  (MCMC) algorithm, such as Hamiltonian Monte Carlo \citep{neal2011mcmc}. In our work, we utilize Langevin Monte Carlo without an accept/reject step \citep[Section 5.3]{neal1993probabilistic} with the following transition:
\begin{align}
    Y_{t+1} &= Y_t - h_t \nabla_{Y_t} E_\theta(Y_t, Z_t)  + \sqrt{2h_t}\omega_t, \\
    Z_{t+1} &= Z_t - h_t \nabla_{Z_t} E_\theta(Y_t, Z_t)  + \sqrt{2h_t}\omega'_t,
    \label{eq:langevin}
\end{align}
where $h_t$ is the step size and $\omega_t, \omega'_t \sim \mathcal{N}(0, \bm{I})$. Samples from the conditional distribution $p_\theta(Z\,|\,Y = y)$ can be generated using Eq. \ref{eq:langevin} for a fixed $Y_t = y$. It can be shown that as $t \rightarrow \infty$ and $\lambda_t \rightarrow 0$ then we have $(Y_t, Z_t) \sim p_\theta(Y,Z)$ \citep{roberts1996exponential}. In practice, the chain is run for a finite number of iterations with a fixed step size, which is sufficient to produce samples close to its stationary distribution $p_\theta(Y,Z)$ \citep{Teh2016,Vollmer2016}. 

Given $n$ samples from the data distribution $\mathbf{Y} := \{ Y_i\}_{i=1}^n \overset{i.i.d}{\sim} p(Y)$, the parameters of an EBM can be obtained by maximizing the log marginal likelihood $\mathcal{L}(\theta) = \frac{1}{n}\sum_{i=1}^n\log \int_\mathcal{Z} p_\theta (Y_i, z) dz$. However, directly optimizing $\mathcal{L}$ is infeasible due to the intractability of the normalizing constant. Contrastive divergence is a method for approximately maximizing the marginal likelihood by approximating the gradient update. The derivative of the log marginal likelihood can be written as 
\begin{align}
    \frac{\partial\mathcal{L}(\theta)}{\partial\theta} &= \mathbb{E}_{Y_i \sim p(\bm{Y})}\mathbb{E}_{Z \sim p_\theta(Z\,|\,Y_i)}\left [\frac{\partial E_\theta(Y_i, Z)}{\partial\theta} \right ] \\
    &- \mathbb{E}_{(Y, Z) \sim p_\theta(Y, Z)}\left [\frac{\partial E_\theta(Y, Z)}{\partial\theta} \right ],
\end{align}
where $p(\mathbf{Y})$ is the empirical distribution of $\mathbf{Y}$. The derivation can be found in Appendix \ref{sec:derivation_gradient_ml}. This quantity can be estimated using samples from the conditional and joint distribution for the respective terms. However, generating these samples from most models is a difficult problem. Instead, Contrastive divergence uses samples drawn approximately from the required distributions by running short run MCMC chains and so defined as
\begin{align}
\mathrm{CD}(\theta) &= \mathbb{E}_{Y_i \sim p(\bm{Y})}\mathbb{E}_{Z \sim \Lambda [p^k_\theta(Z\,|\,Y_i)]} E_\theta(Y_i, Z) \\
    &- \mathbb{E}_{(Y, Z) \sim  \Lambda [ p^k_\theta(Y, Z)]}E_\theta(Y, Z),
    \label{eq:cd}
\end{align}
where $p^k_\theta(Z\,|\,Y_i)$ and $p^k_\theta(Y, Z)$ is the conditional and joint distribution respectively after running $k$ steps of the Langevin dynamics for instance, and $\Lambda$ is the stop gradient operator as in \citet{pmlr-v139-du21b}.
Intuitively, minimizing Equation \ref{eq:cd} results in a decrease of the energy of samples from the data distribution (and posterior) and increase in the synthetic samples, which follows  ``analysis by synthesis'' scheme \citep{grenander2007pattern}.

\section{PROPOSAL: $\mathcal{F}$-EBM}

\begin{figure*}[h]
    \centering
    \begin{subfigure}[b]{0.32\linewidth}
        \includegraphics[width=\linewidth]{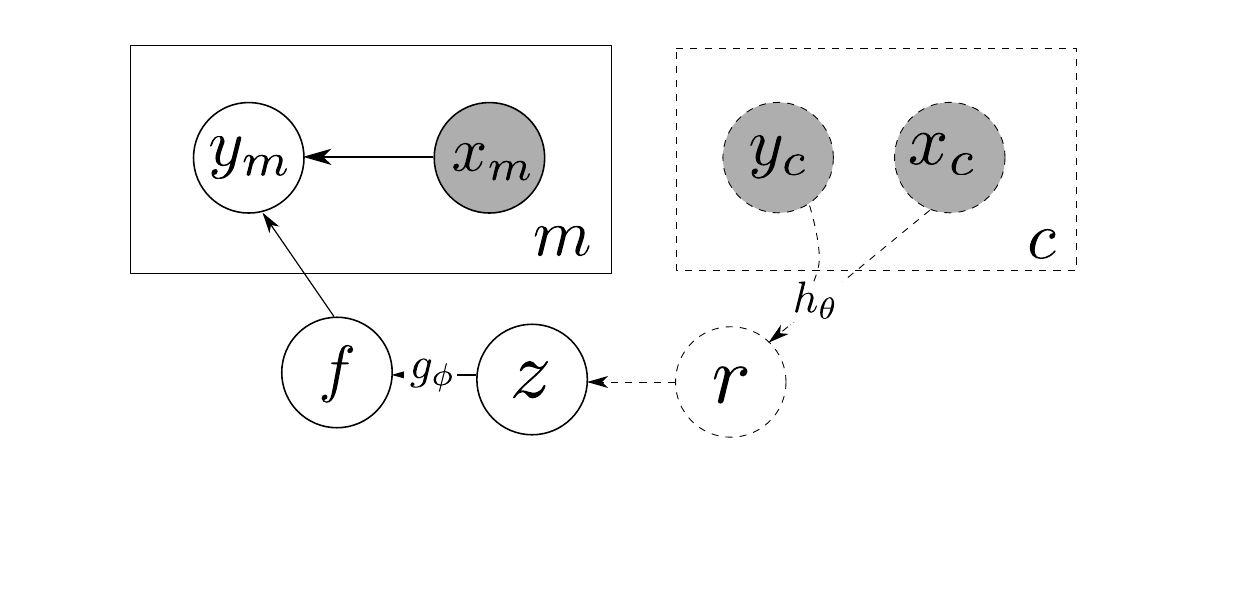}
        \caption{Neural Process}\label{fig:gm_np}
    \end{subfigure}
    \begin{subfigure}[b]{0.32\linewidth}
        \includegraphics[width=\linewidth]{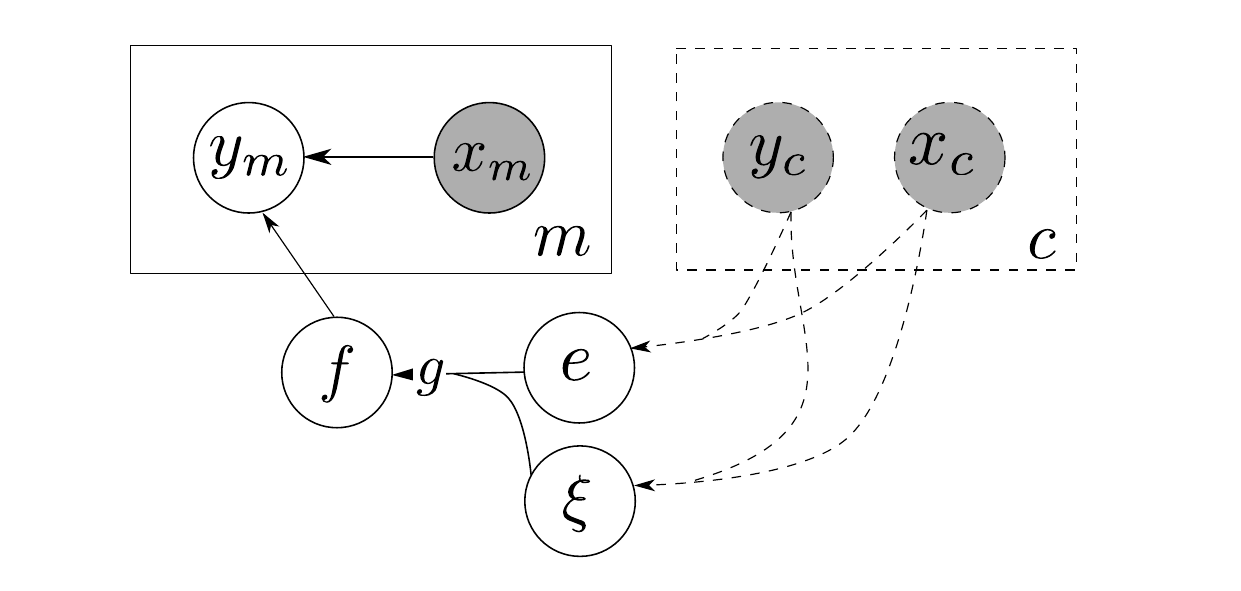}
        \caption{Gaussian Process (weight-view)}\label{fig:gm_gp}
    \end{subfigure}
    \begin{subfigure}[b]{.32\linewidth}
        \includegraphics[width=\linewidth]{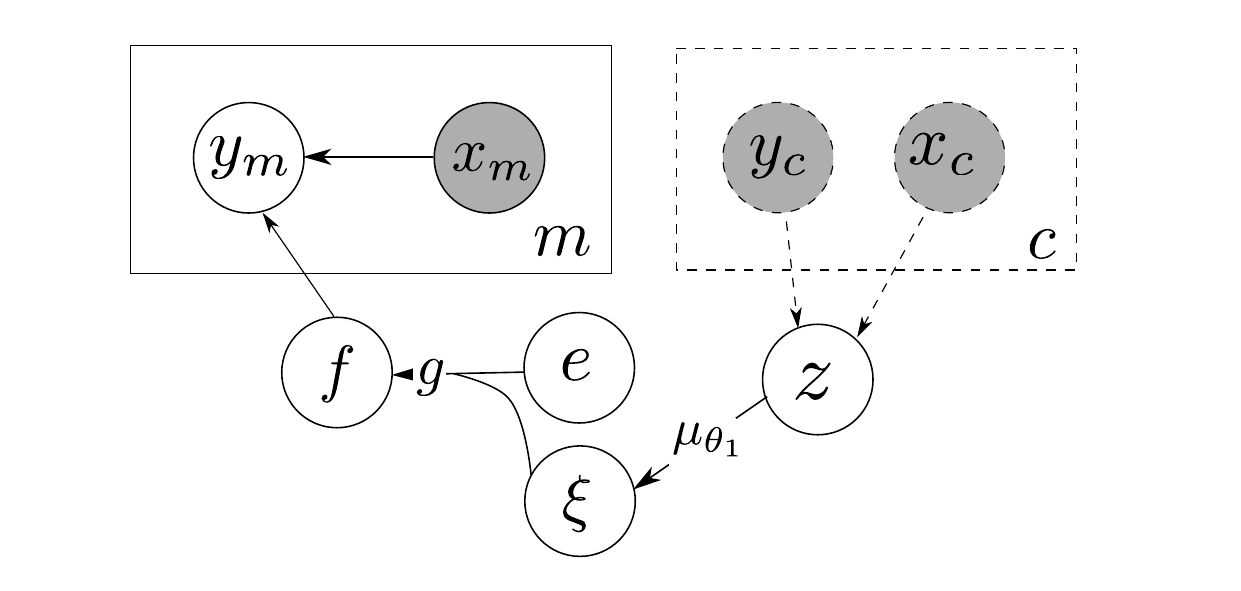}
        \caption{$\mathcal{F}$-EBM (ours)}\label{fig:gm_febm}
    \end{subfigure}
    \caption{Graphical models of (\subref{fig:gm_np}) Neural process, (\subref{fig:gm_gp}) Gaussian process, and (\subref{fig:gm_febm}) $\mathcal{F}$-EBM. Grey nodes indicate the variable is observed. Dashed lines indicate inference (given $c$ context points). Continuous lines indicate the generative process, and dashed lines the inference.}
    \label{fig:gm}
\end{figure*}

\textbf{Setting}.
We assume that we observe $N$ independent realizations $f_1,\ldots, f_N \in \mathcal{F}$ of an (unknown) probability measure $\mathbb{P}$ supported on the space $\mathcal{F}:\mathcal{X} \rightarrow \mathcal{Y}$. For an arbitrary function $f_i$, it is evaluated on $M$ (possibly distinct) points $X_i:=(x_m)_{m=1}^M \subset \mathcal{X}^M$ taking values $Y_i:= \left(f\left(x_m \right)\right)_{m=1}^M \subset \mathcal{Y}^M$.   In this setting, the dataset  takes the form $(\bm{X}, \bm{Y})$, where $\bm{X} = \{X_n\}_{n=1}^N$ and $\bm{Y} = \{Y_n\}^N_{n=1}$. Note that, for notational simplicity, we shall assume that each representation has the same number of evaluation points. The extension to the more general setting is straightforward.   Given the tuple $(\bm{X},\bm{Y})$, our objective is to learn a representation of the unknown probability distribution $\mathbb{P}$.

\textbf{Gaussian Process (weight-view)}. Consider a mean-zero Gaussian process (GP) $\mathrm{GP}(0, k)$ with positive-definite kernel $k$, the Karhunen-Loeve expansion \citep[Section 11.1]{sullivan2015introduction} provides a method for sampling from the GP via its eigensystem $\{(\lambda_i,e_i)\}_{i=1}^\infty$. In other words, through Mercer's theorem, we have $k(x,t) = \sum_{i=1}^\infty \lambda_i e_i(x) e_i(t)$ where the eigenvalues $\lambda_i$ and eigenfunctions $e_i(\cdot)$ are solutions to the eigenvalue problem $\lambda_i e_i(t) = \int_\mathcal{X} k(t, x)e_i(x)\,dx$. Then, Karhunen-Loeve expansion states that samples from the GP can be expressed as the following infinite sum
$
f(\cdot) = \sum_{i=1}^\infty \xi_i  \sqrt{\lambda_i}e_i(\cdot)
$
where $\{\xi_i\}_{i=1}^\infty$ is a sequence of independent $\mathcal{N}(0, 1)$ random variables.
Given $\xi = (\xi_i)_{i=1}^\infty\sim \Pi := \prod_{i=1}^\infty \mathcal{N}(0, 1)$, the map $g(\xi)(\cdot) = \sum_{i=1}^\infty \xi_i \sqrt{\lambda_i} e_i(\cdot)$ defines a GP path measure as the push forward of measure $\Pi$. In other words, for $\xi \sim \Pi$, we have $g(\xi)\sim \mathrm{GP}(0,k)$. Figure \ref{fig:gm_gp} shows the graphical model of a GP in weight-space view.

\textbf{Proposal: $\mathcal{F}$-EBM}. The Karhunen-Loeve weight-space interpretation provides an inspiring perspective for constructing an energy-based model that can be informed by the kernel. We propose a distribution of functions which is tilted by a GP path measure: the path measure plays the role of reweighting the eigenfunctions depending on its smoothness.

Given a kernel and its (approximate) eigensystem $(\{\lambda_i\}_{i=1}^{d_\xi},\{e_i(\cdot)\}_{i=1}^{d_\xi})$, we define the map $\tilde{g}_{\theta_{gen}} (Z)(\cdot) := \sum_{i=1}^{d_{\xi}}\mu_{\theta_{gen}} (Z)_i\sqrt{\lambda_i}e_i(\cdot)$
where $\mu_{\theta_{gen}}: \mathbb{R}^{d_Z} \rightarrow \mathbb{R}^{d_{\xi}}$ is a parameterized function. The eigenvalues play the crucial role of reweighting the eigenfunction to bias the model towards smoother functions; the level of which is determined by the kernel. Each sample takes the form
$f(\cdot) = \tilde{g}_\theta (Z)(\cdot)$ where $Z$ be a $\mathbb{R}^{d_z}$-valued r.v.\ with density $p_{\theta_{prior}}$. The function $\mu_{\theta_{gen}}$ can be thought of as a ``generator'' that synthesizes samples in the weight space instead of generating directly in the data space.

As for the distribution of $Z$, denoted by $p_{\theta_{prior}}(Z)$, there are a variety of choices. We  utilize energy-based modelling and choose a prior density $p_{\theta_{prior}}(Z) \propto \exp(-E_{\theta_{prior}}(Z))p_0(Z)$ where $E_{\theta_{prior}}:\mathbb{R}^{d_Z}\rightarrow \mathbb{R}$ and $p_0 := \prod_{i=1}^{d_{Z}}\mathcal{N}(0, \sigma^2_0)$ is the reference or base distribution. This technique is known as exponential tilting (see \citet{xiao2020exponential, pang2020learning} in the context of generative modelling). We found that a Gaussian prior for $p_{\theta_{prior}}$ was too simple which resulted in poor fit of the model, and using an energy-based prior (with no base distribution) resulted in a distribution that is hard to sample from (see Section \ref{sec:exp_prior}).

 Given evaluations of a function $(X,Y)$, we relate the observations to an underlying function $\tilde{g}_{\theta_{gen}}(Z)$ (induced the random variable $Z$) using a likelihood function $p_{\theta_{gen}}(Y |\, Z, X)$. The likelihood is constructed by assuming that each point is sampled independently of each other. In order words, the likelihood $p_{\theta_{gen}}(Y |\, Z, X)$ takes the form $\Pi_{i=1}^M p_{\theta_{gen}}(y_i\,|,\,Z, x_i)$. There are many choices for the (point-wise) likelihood. In this work, we focus on the case where $p_{\theta_{gen}}(y_i\,|\,Z, x_i)$ is Gaussian and (continuous) Bernoulli (see \citet{NEURIPS2019_f82798ec}). More precisely, for the Gaussian case, we have
 $
 p_{\theta_{gen}}(y_i\,|\,Z, x_i) \propto \exp(-\|y_i - \tilde{g}_{\theta_{gen}}(Z)(x_i)\|^2/2\sigma^2).
 $

Combining the Karhunen-Loeve interpretation and (tilted) energy-based prior, we can construct a generative model over function space. In particular, we propose the following energy-based model:
$$
p_{\theta}(Y, Z|\, X) = p_{\theta_{gen}}(Y|\, Z, X)p_{\theta_{prior}}(Z),
$$
where $\theta = (\theta_{gen},\theta_{prior})$. The difference between $\mathcal{F}$-EBM, and Neural processes, and Gaussian processes can be seen from the graphical model perspective as in Figure \ref{fig:gm_febm}. In Appendix \ref{sec:math_background}, we show that $\mathcal{F}$-EBM defines a valid stochastic process using Kolmogorov extension theorem.

One method of training the model is by minimizing contrastive divergence (CD) \citep{hinton2002training}. However, we found that directly applying contrastive divergence as in Eq.\ \ref{eq:cd} did not produce sufficiently satisfying results. Instead, following in \citet{pang2020learning}, one can improve the objective by returning to the gradient of the maximum likelihood and applying further reductions to avoid approximations until the last step. The gradients of the marginal likelihood can be written as
\begin{align*}
    \nabla_{\theta_{gen}} \mathcal{L}(\theta) &=  - \mathbb{E}_{p(\bm{X}, \bm{Y})} \mathbb{E}_{p_\theta (Z|X, Y)}\nabla \log p_\theta(Y| Z, X), \\
    \nabla_{\theta_{prior}} \mathcal{L}(\theta) &= \mathbb{E}_{p(\bm{X}, \bm{Y})}\mathbb{E}_{p_\theta(Z|X, Y)} \nabla E_\theta(Z)  \\
    &- \mathbb{E}_{p_{\theta}(Z)} \nabla E_\theta(Z).
\end{align*}
Its derivation can be found in Appendix \ref{sec:derivation_gradient}. Now, we can apply approximation techniques and estimated the gradient using samples generated from Langevin MC targeting the prior $p_{\theta_{prior}}(Z)$ and the conditional distribution $p_\theta(Z\,|\,Y, X)$. In other words, we minimize the following loss:
\begin{align}
\begin{split}
\label{eq:loss}
\mathcal{L}(\theta) := - \mathbb{E}_{p(\bm{X}, \bm{Y})} \mathbb{E}_{\Lambda(p^k_\theta (Z|X, Y))} \log p_\theta(Y| Z, X), \\
 + \mathbb{E}_{p(\bm{X}, \bm{Y})}\mathbb{E}_{\Lambda(p^k_\theta(Z|X, Y))} E_\theta(Z) 
    - \mathbb{E}_{\Lambda(p^k_{\theta}(Z))} E_\theta(Z).
\end{split}
\end{align}
where $p^k(\cdot)$ denotes the distribution after $k$ steps of the Langevin algorithm, and $\Lambda$ is the stop gradient operator.
The loss yields a simple interpretation. When minimizing this quantity, the parameter $\theta_{gen}$ will be updated such the observation $(X,Y)$ and the posterior latent variable $Z$ has a higher likelihood. In the case of a Gaussian likelihood, this will correspond to minimizing the mean squared error between $Y$ and $\tilde{g}_\theta (Z)(X)$. As for the prior parameter $\theta_{prior}$, the update is akin to CD where we are decreasing the energy of the posterior latent variable and increasing the energy of the model latent samples.

\textbf{Estimating eigensystem of a kernel}. For a given kernel $k$, we typically do not have analytical expressions for the associated Mercer eigensystem and must resort to numerical approximation. To this end, we employ the Nystr\"om method to approximate the eigenvalue problem, i.e., we have 
\begin{equation}
    \lambda_i e_i(t) = \int_\mathcal{X} k(t, x)e_i(x)dp(x) \approx \frac{1}{l}\sum_{j=1}^l k(t, x_j)e_i(x_j),
    \label{eq:approx_eigenproblem}
\end{equation}
for some choice of $X :=\{x_j\}_{j=1}^l \subset \mathcal{X}$. Substituting $t =x_k$ for $k=1, \ldots, l$ results in an eigenvalue problem $\frac{1}{l} K({X}, {X}) \hat{\bm{e}}_i(X) = \hat{\lambda}_i \hat{\bm{e}}_i({X})$ where $K(X, X) = (k(x_i,x_j)_{ij}) \in \mathbb{R}^{l \times l}$ is the gram matrix, and $\hat{\bm{e}}_i(X) = [\hat{e}_i(x_1), \ldots, \hat{e}_i(x_l)]^\top \in \mathbb{R}^l$. To obtain $\{(\hat{\lambda}_i, \hat{\bm{e}}_i(X))\}_{i=1}^l$, we solve for the eigenspectrum of the scaled gram matrix $\frac{1}{l} K(X, X)$ with eigenvectors normalized to have $\ell^2$ norm equal to $\sqrt{l}$ yields the desired result. The eigenvalues $\hat{\lambda}_i$ converge to $\lambda_i$ in the limit $l \rightarrow \infty$ \citep[Theorem 3.4]{Baker1979}. The eigenvectors correspond to the eigenfunctions $\{\hat{\bm{e}}_i(X)\}^l_{i=1}$ evaluated at $X$ but we require eigenfunctions to be evaluated at arbitrary locations. One common estimator is to solve for $e(t)$ in Equation \ref{eq:approx_eigenproblem} and using estimates of $\{\hat{\lambda}_i, \hat{\bm{e}}_i(X)\}_{i=1}^l$ (for instance, see \citet[Eq. 9]{williams2001using}). However, we found that the estimates of eigenfunctions with small eigenvalues were not accurate. Other methods of accurately interpolating between evaluated points can be used such as kernel ridge regression (see Appendix \ref{sec:interpolation}).

\textbf{Inference of $f$}. Given a context pair $(X^c,Y^c)$, one may want to infer the underlying latent function $f$. To do this, we first obtain samples from the conditional distribution $p(Z\,|\,Y,X)$, which is then passed through the map $\tilde{g}$ to induce a distribution over the functions (see Figure \ref{fig:gm_febm}). We write $p(f\,|\,Y,X)$ to denote the distribution of $\tilde{g}_\theta (Z)\,|\,Y,X$. Inference of a function (corresponding to a potentially infinite-dimensional vector) is cast as inference of $Z$ with a much lower dimensionality and allows for efficient use of MCMC methods.

\textbf{Kernel Choice}. The kernel can be used to capture any prior belief about the underlying function and plays a crucial role in the learnt model. One method is to construct kernels from the product and sums of well-studied existing kernels, such as the Gaussian and Mat\'ern kernel \citep[see Section 4.2]{williams2006gaussian}. Other choices include constructing an explicit feature map with a neural network, or obtain a data-driven basis via functional Principal component analysis. A combination of all the aforementioned methods can also be used in conjunction.

\section{RELATED WORK}
\label{sec:related}
Stochastic processes provide an elegant method for defining distributions of functions. A popular example is the Gaussian processes (GP) \citep{williams2006gaussian}. A significant search effort has been made to improve expressivity of GP by learning complex kernel functions (see \citet{damianou2013deep, wilson2016stochastic} to name a few). An alternative research direction has been to construct expressive stochastic processes using neural networks. These include neural processes (NP) and its variants
\citep{garnelo2018conditional, garnelo2018neural, al2017learning}, EBP \citep{yang2020energy}, $\pi$-VAE \citep{mishra2022pi} and Variational implicit processes (VIP) \citep{ma2019variational}. NP and VIP has been proposed to learn an approximation of a stochastic process by combining the perks of GPs and neural networks for scalable inference while quantifying its uncertainty. While VIP, GPs and NPs can be used as generative models for functional data, the focus has been on uncertainty quantification and prediction rather than generative modelling. \citet{yang2020energy} proposed the form of EBM for stochastic processes called energy-based process (EBP) taking the form
$p_\theta(Y, Z; X) \propto \exp (-f_\theta (Y,Z; X))p(Z)$
where particular forms of the likelihood $f_\theta$ and $p(Z)$ recovers popular models such as Gaussian processes and Neural processes.

Our proposal $\mathcal{F}$-EBM is a particular instantiation of both EBP and VIP. Energy-based processes \citep[Section 3.2]{yang2020energy} have the form $p_\theta(Y, Z; X) \propto \exp (-f_\theta (Y,Z; X))p(Z)$, and so choosing $f_\theta (Y,Z; X)) = -\log \left [ p_{\theta_{gen}}(Y| Z, X) \right ] +E_{\theta_{prior}}(Z)$ and $p(Z) = p_0(Z)$, it is clear that $\mathcal{F}$-EBM is a particular EBP. As for (noiseless) Variational implicit processes \citep[Definition 1]{ma2019variational}, they have the form $f=h_\theta(\cdot, Z)$ and $Z \sim p(Z)$ and choosing $h_\theta (\cdot, Z) = \tilde{g}_\theta (Z)(\cdot)$ and $Z$ to be sampled from a distribution with density $p_{\theta_{prior}}(Z)$, we have shown that $\mathcal{F}$-EBM is a VIP.

The recently proposed $\pi$-VAE \citep{mishra2022pi} extends the VAE formalism to function classes and effectively providing a generative model for stochastic processes. This is achieved by performing a projection of the data onto a finite number of basis of functions.  The encoder/decoder pair are then trained to learn the distribution of the associated basis coefficients, from which realizations of the learned  stochastic process can be readily generated. This is similar, in spirit, to our approach which relies on a set of basis functions and transforms the problem into density estimation on the weight space domain. However, our approach can be seen as defining a prior for this transformation resulting in a more stable transformation, particularly for irregular meshes which is not the case for $\pi$-VAE as seen in Table \ref{tab:sse_loss}.

There are other related works that focus on learning implicit representations of images as functions \citep{pmlr-v151-dupont22a, anokhin2021image} and physics-informed modelling \citep{yang2020physics, meng2022learning}. 
\citet{pmlr-v151-dupont22a} proposed a GAN approach for learning a generative model for images as functions. Concurrently, PI-GAN \citep{yang2020physics}  proposed as a GAN approach for learning an approximation of a stochastic process, focusing on endowing the generator with prior knowledge in the form of stochastic differential equation. \citet{meng2022learning} extended it further by using DeepONets \citep{lu2021learning} to incorporate physical knowledge and proposed a method for inference of the latent function by using HMC \citep{neal2011mcmc}. Note that the form of the PI-GAN requires some extensions to accommodate for learning functions with different evaluation points whilst our proposal does not.

\section{EXPERIMENTS}

\begin{table*}[ht]
\tiny
\centering
\caption{Visualisation of $100$ samples drawn from the dataset, $\mathcal{F}$-EBM (ours), $\pi$-VAE, Neural Process (NP), Energy-based processes (EBP), and Variational Implicit processes (VIP).}
\makebox[\textwidth]{
    \begin{tabular}{@{}ccccccc@{}}
    \toprule
     & \centering{Dataset} & $\mathcal{F}$-EBM (ours) & $\pi$-VAE & NP & EBP & VIP\\ \midrule
    \texttt{Quadratic} 
    & \includegraphics[width=.12\linewidth,valign=m]{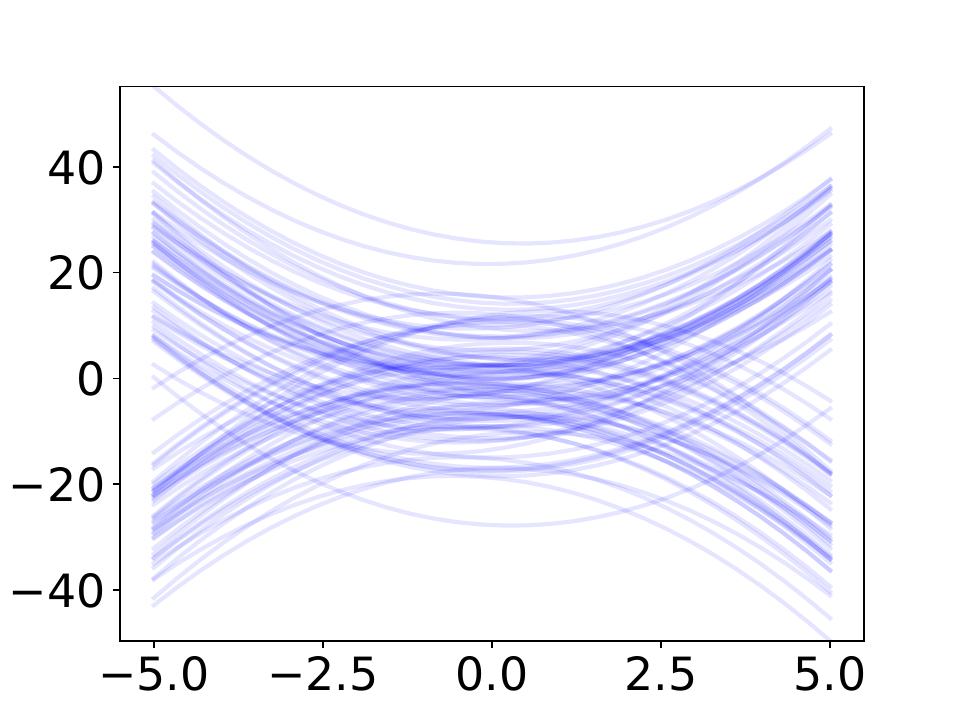} 
    & \includegraphics[width=.12\linewidth,valign=m]{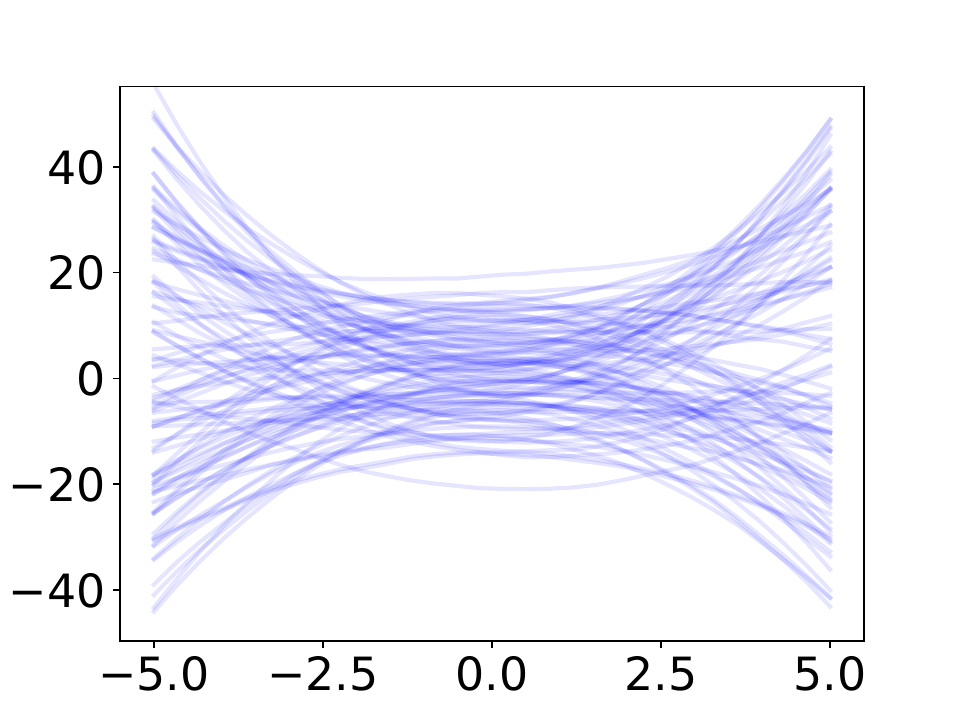} 
    & \includegraphics[width=.12\linewidth,valign=m]{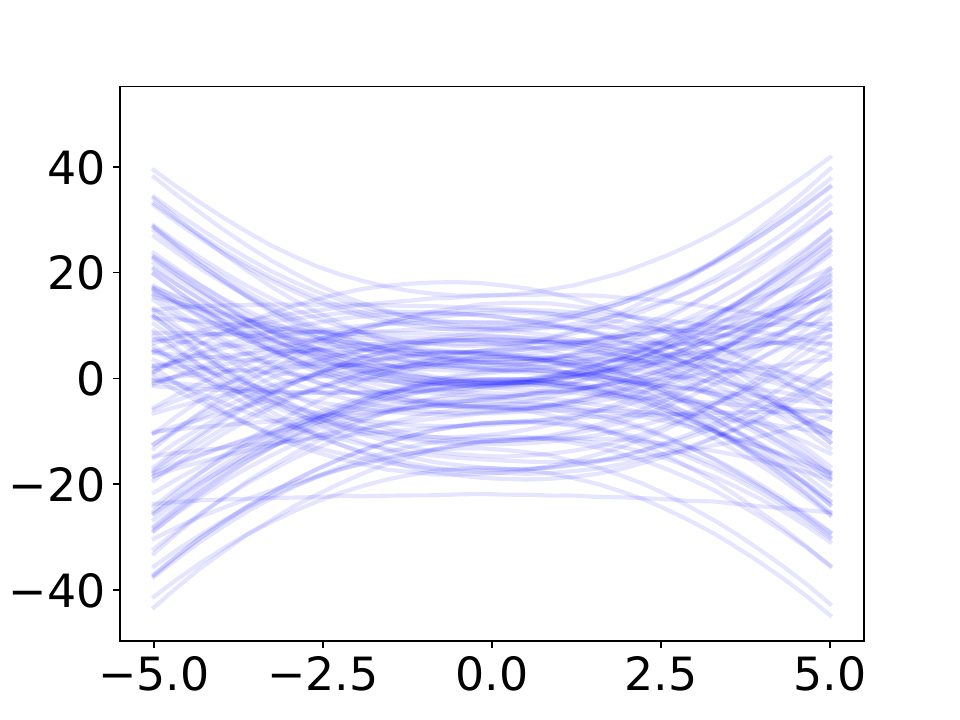}
    & \includegraphics[width=.12\linewidth,valign=m]{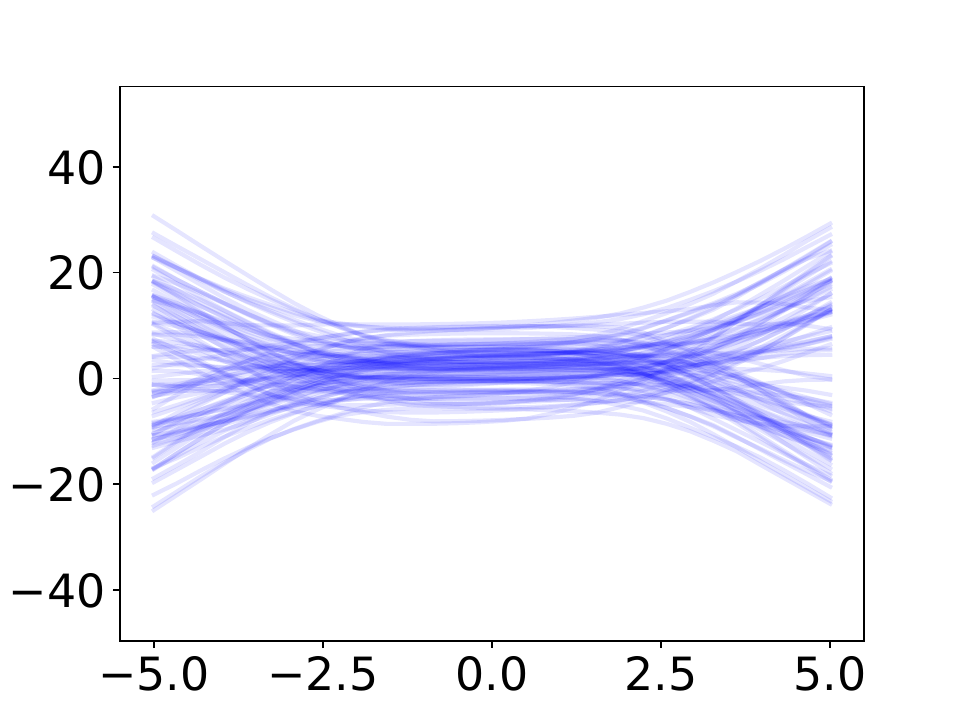}
    & \includegraphics[width=.12\linewidth,valign=m]{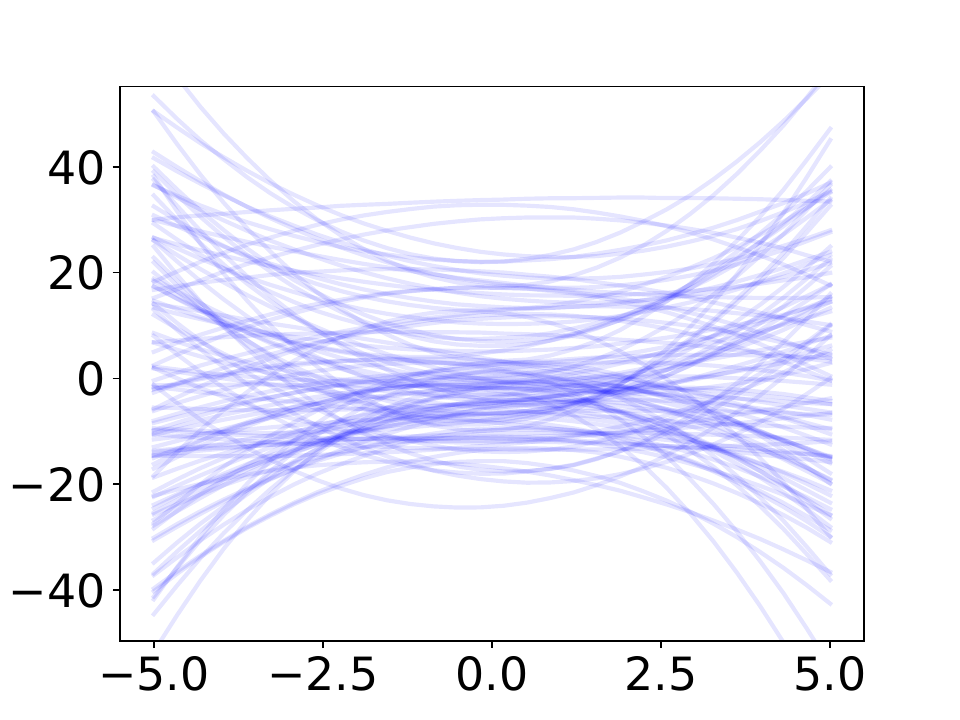}
    & \includegraphics[width=.12\linewidth,valign=m]{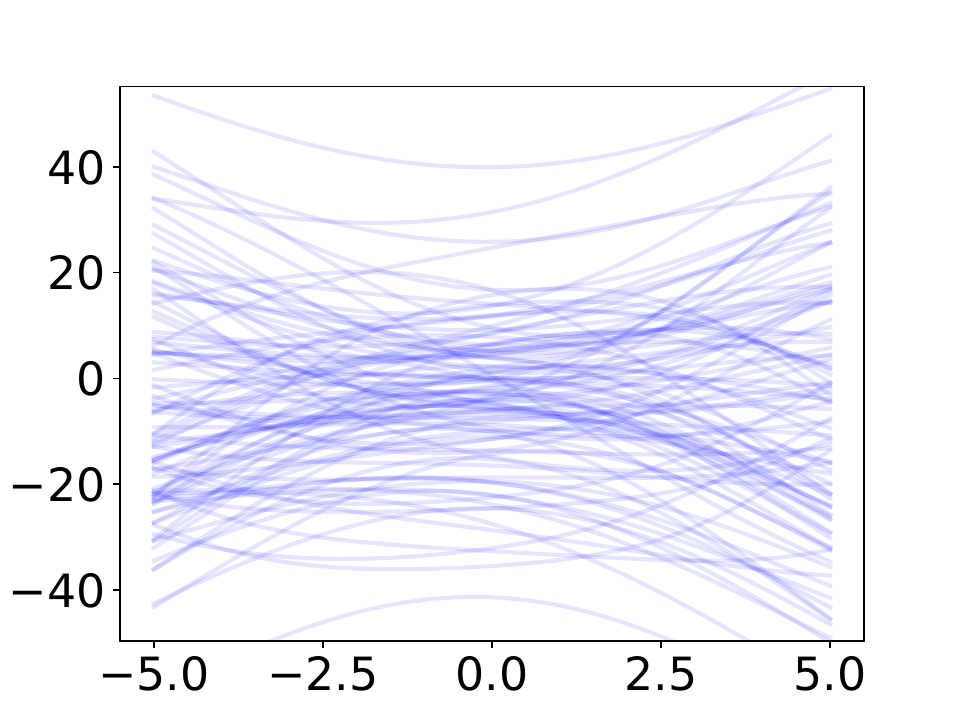}
    \\
    \texttt{Melbourne} 
    & \includegraphics[width=.12\linewidth,valign=m]{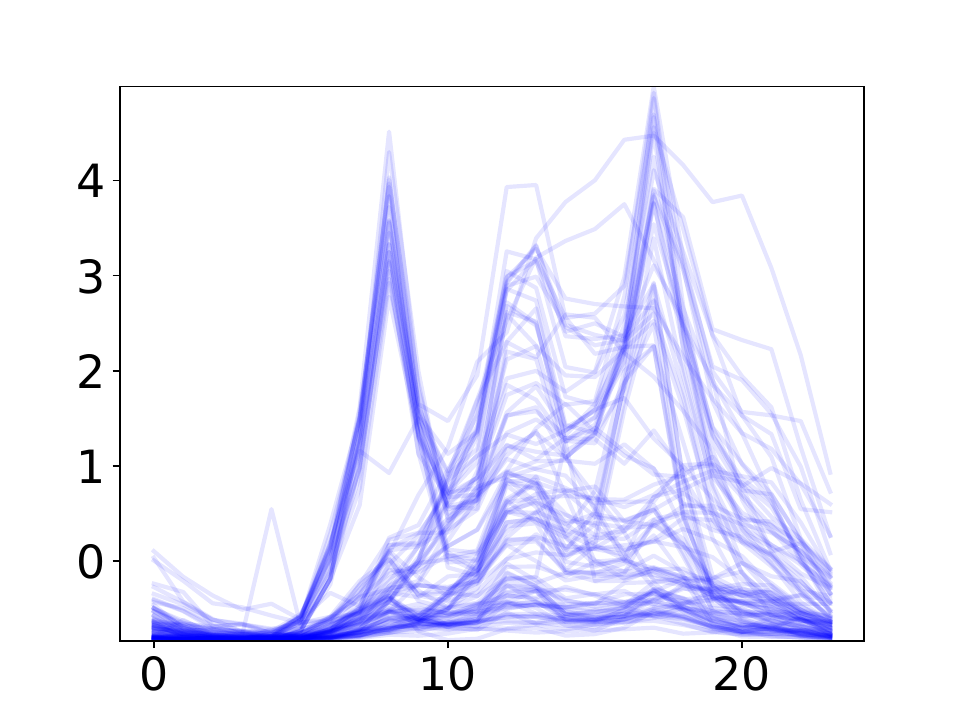}
    & \includegraphics[width=.12\linewidth,valign=m]{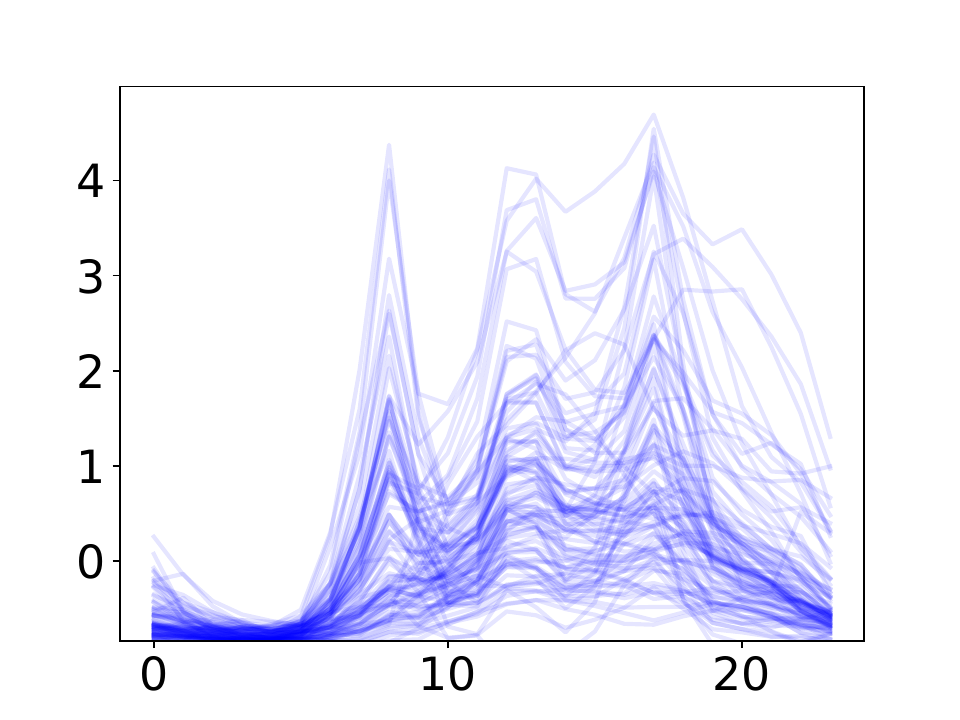} 
    & \includegraphics[width=.12\linewidth,valign=m]{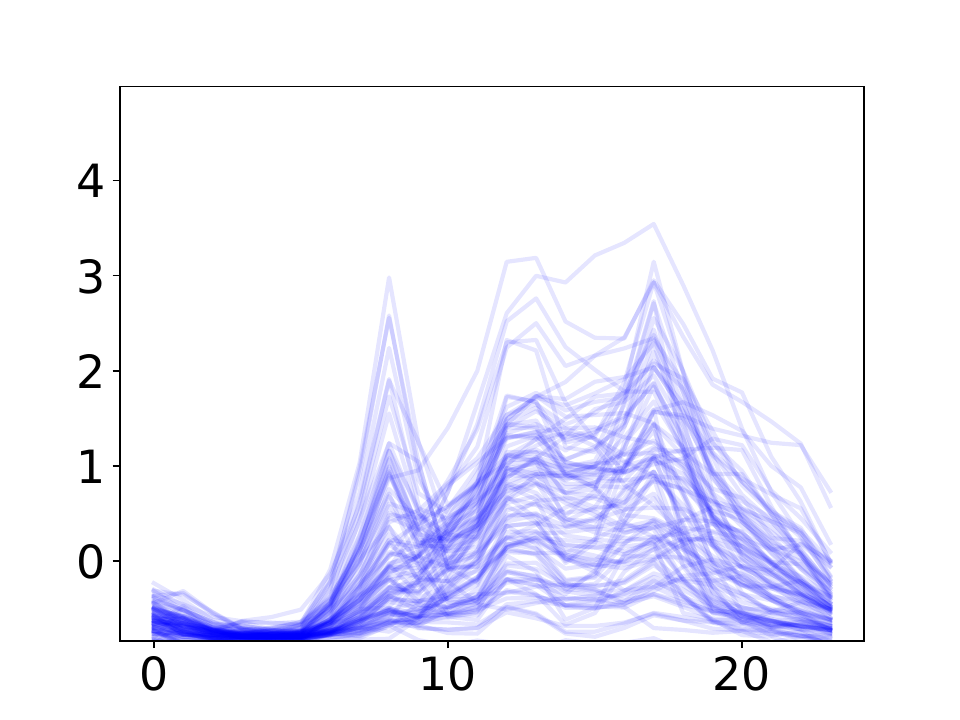}
    & \includegraphics[width=.12\linewidth,valign=m]{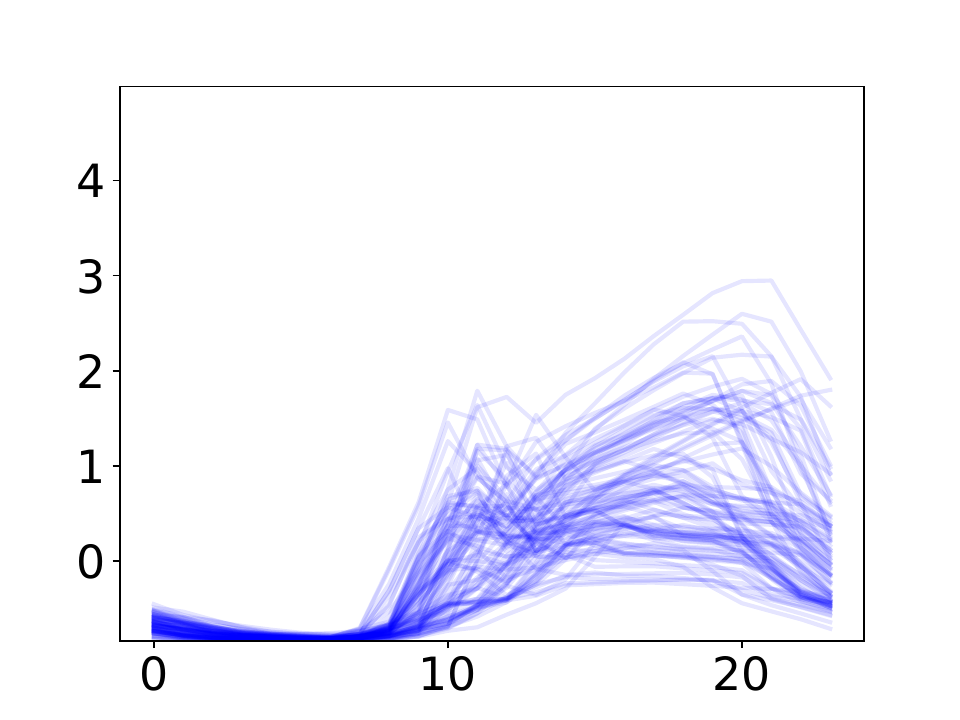}
    & \includegraphics[width=.12\linewidth,valign=m]{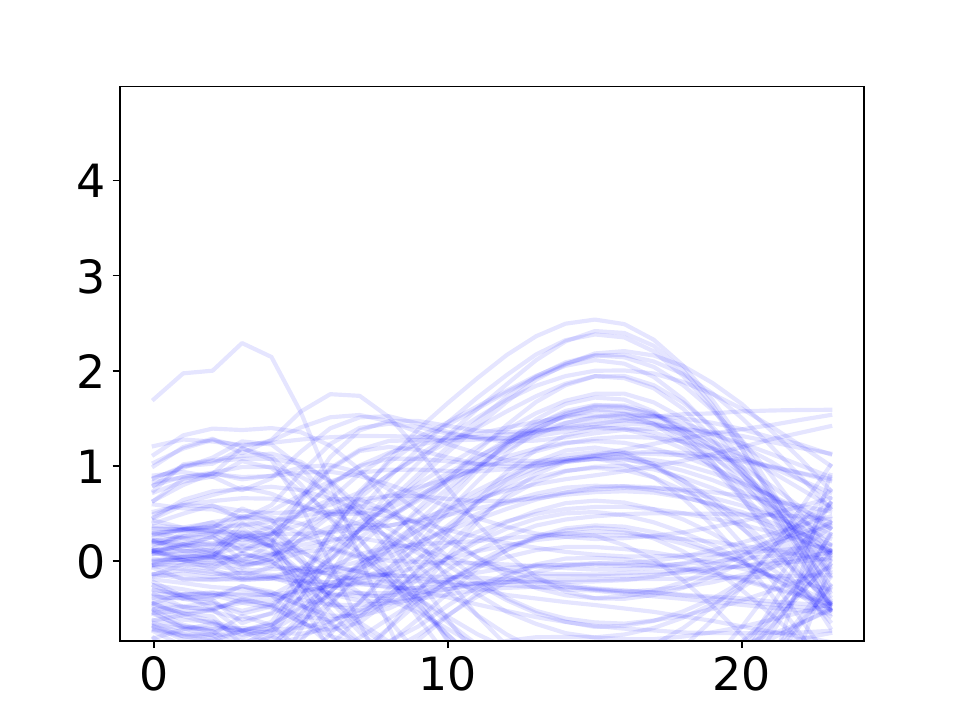}
    & \includegraphics[width=.12\linewidth,valign=m]{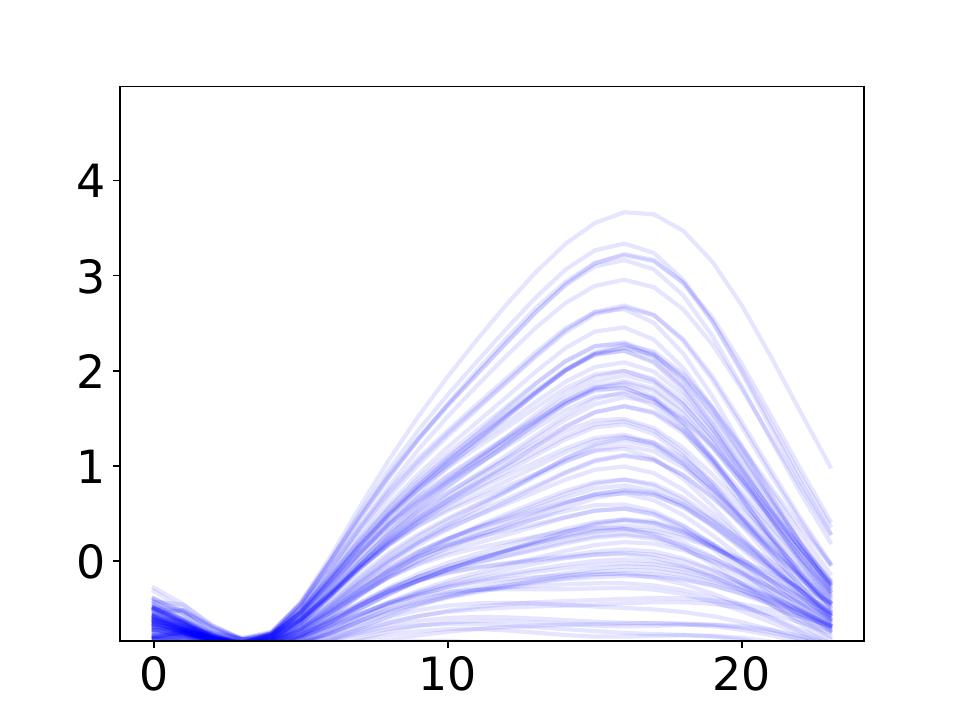}\\
    \texttt{Stocks} 
    & \includegraphics[width=.12\linewidth,valign=m]{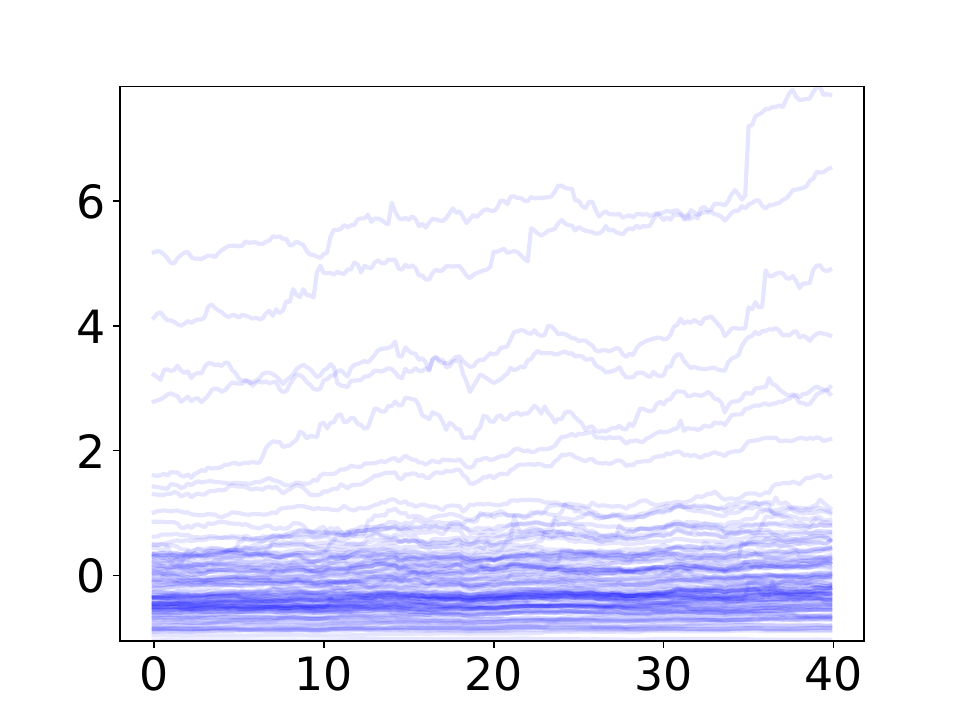} 
    & \includegraphics[width=.12\linewidth,valign=m]{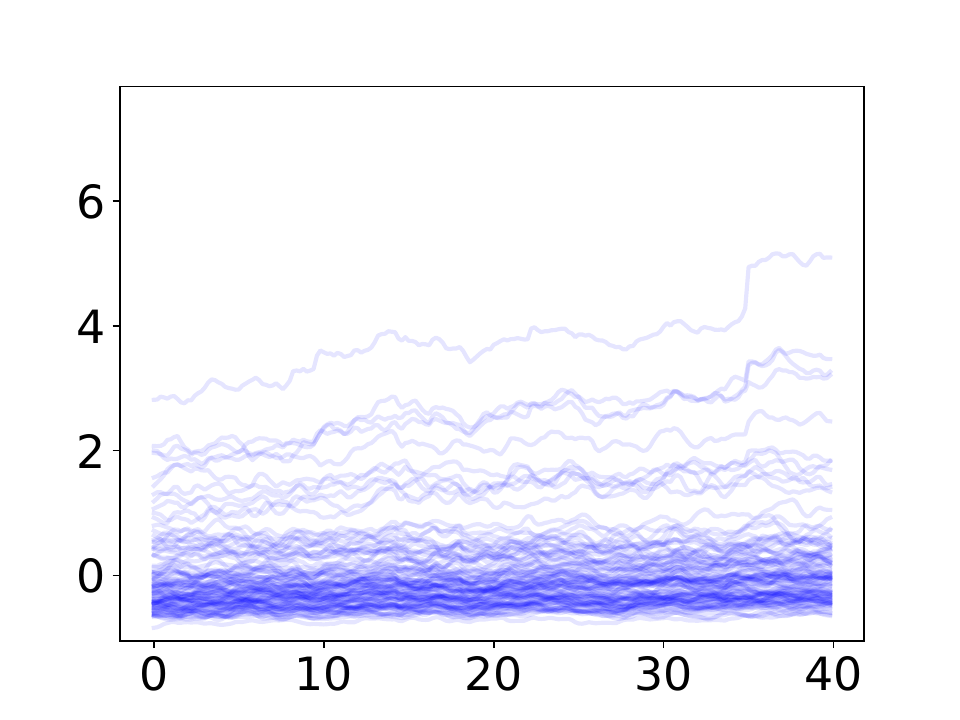} 
    & \includegraphics[width=.12\linewidth,valign=m]{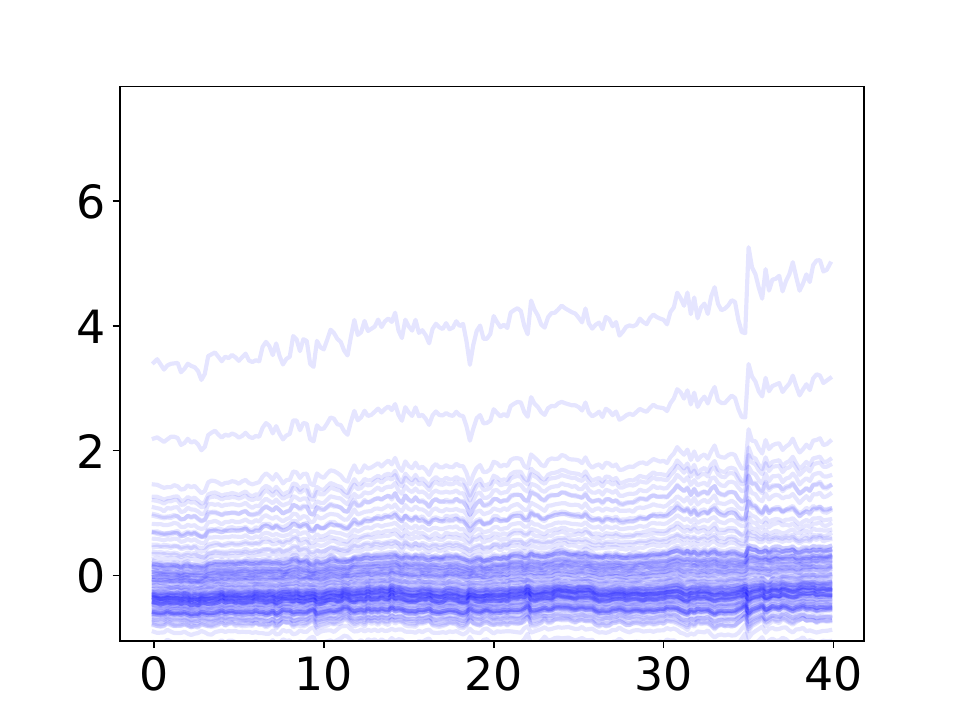}
    & \includegraphics[width=.12\linewidth,valign=m]{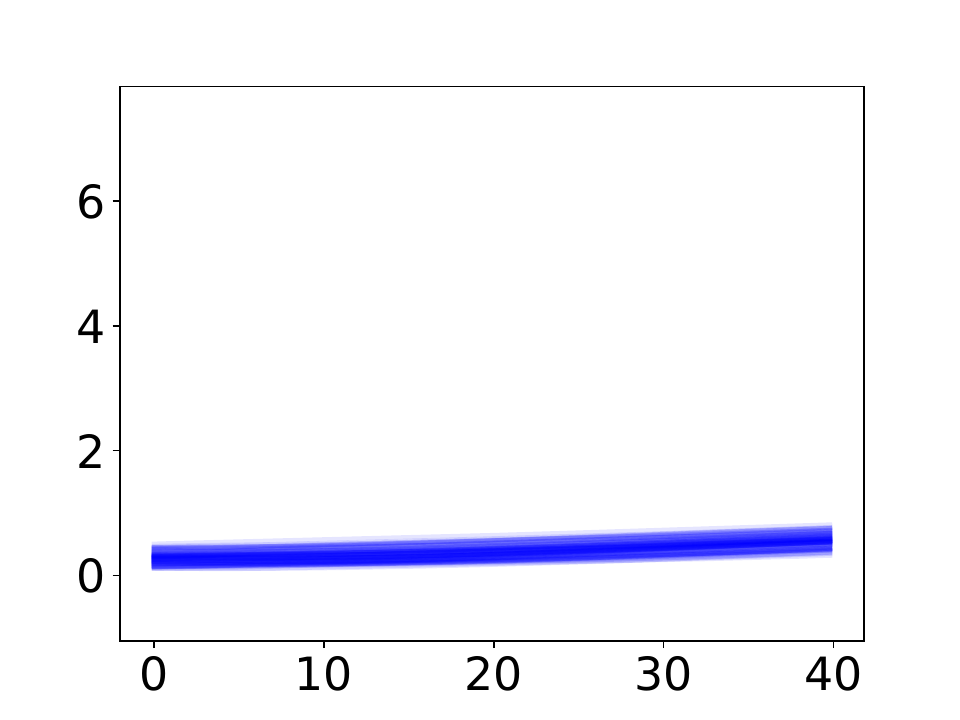}
    & \includegraphics[width=.12\linewidth,valign=m]{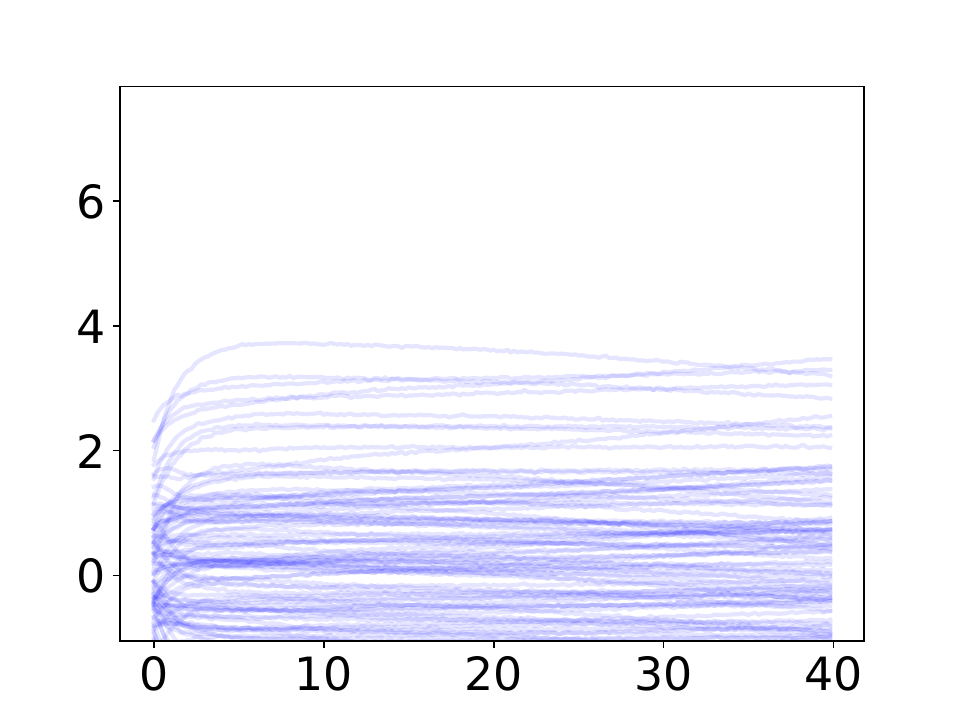}
    & \includegraphics[width=.12\linewidth,valign=m]{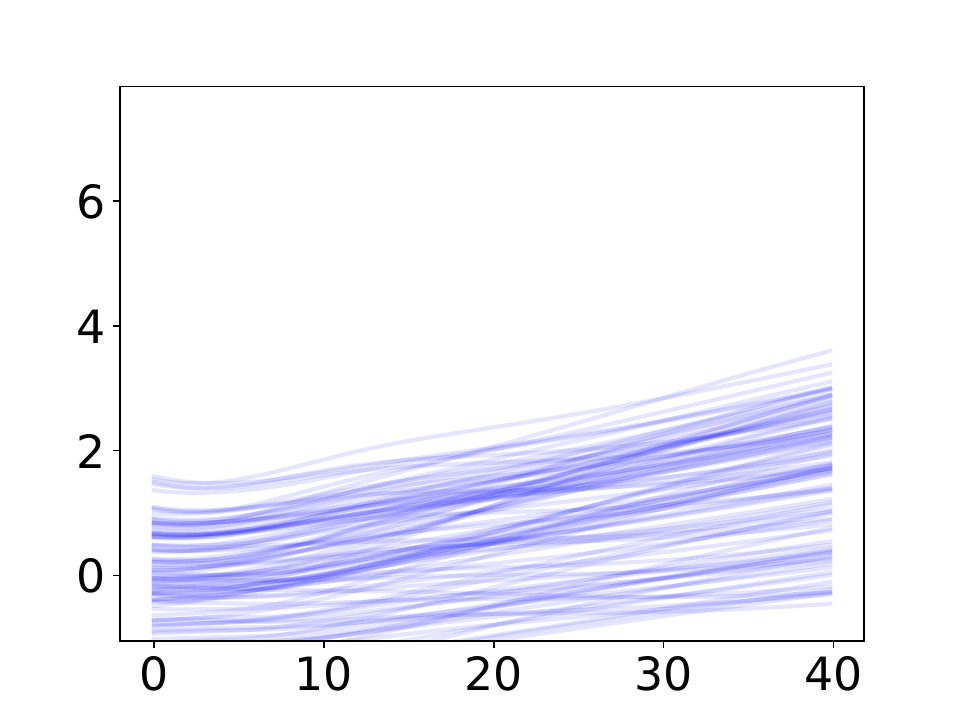}\\
    \texttt{GridWatch} 
    & \includegraphics[width=.12\linewidth,valign=m]{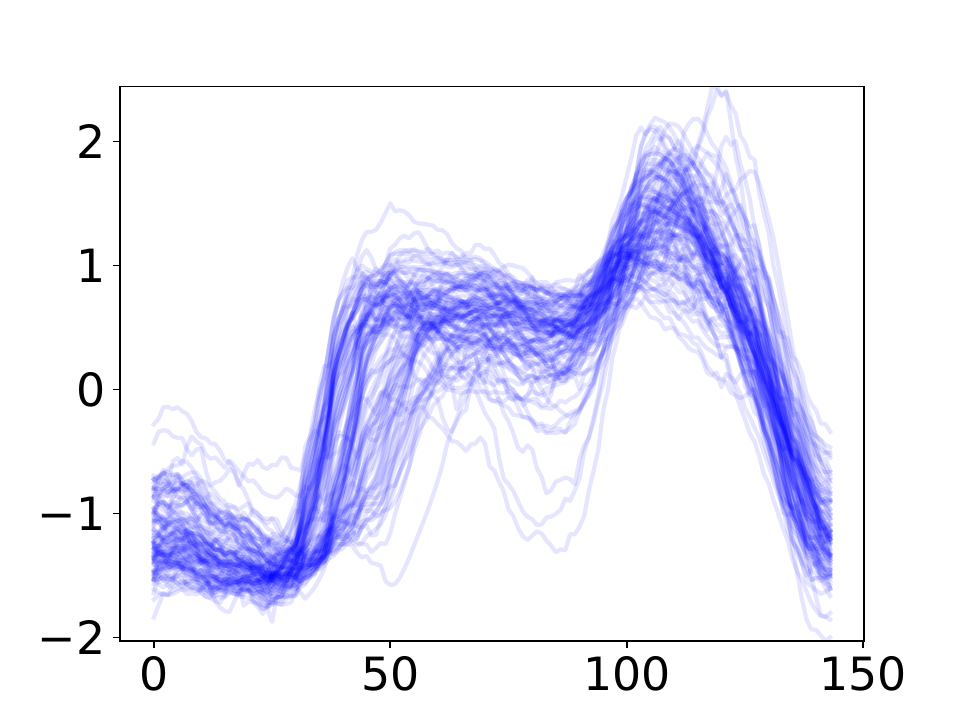} 
    & \includegraphics[width=.12\linewidth,valign=m]{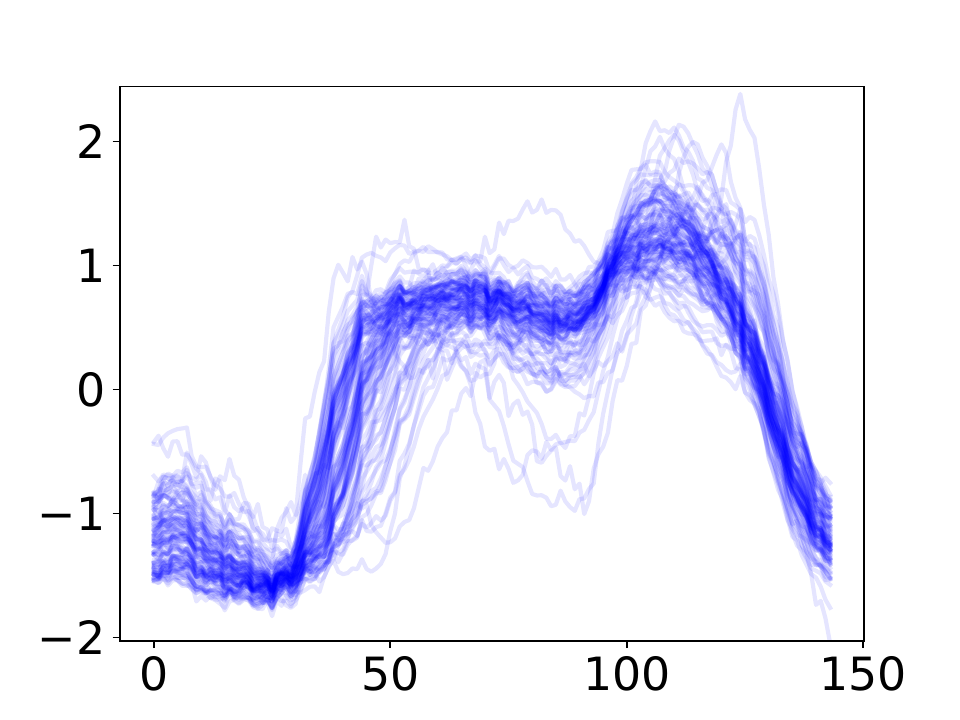}
    & \includegraphics[width=.12\linewidth,valign=m]{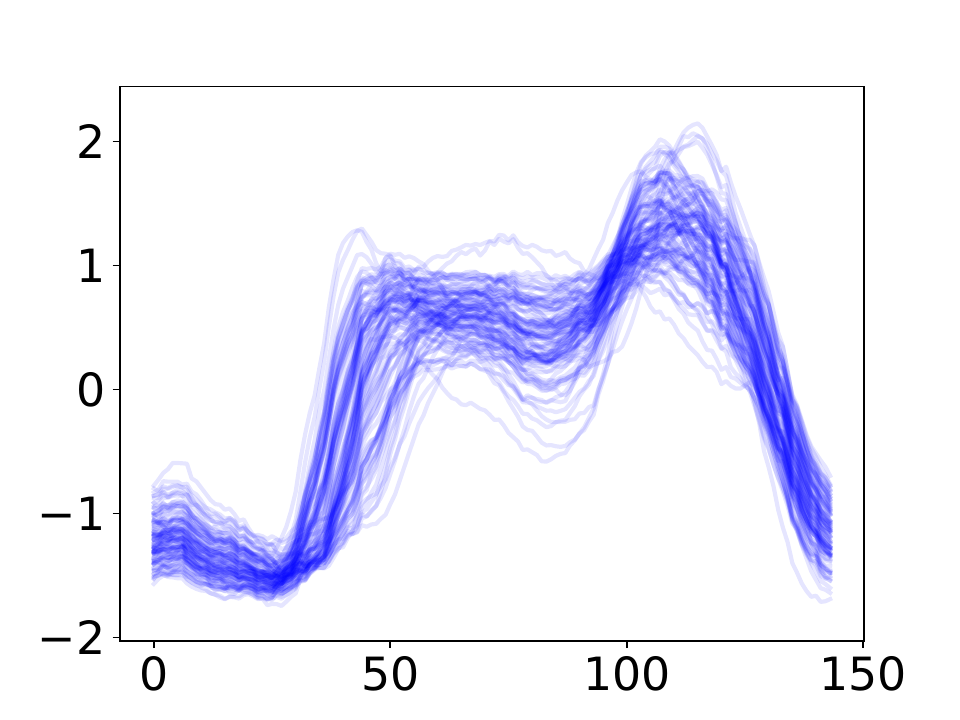}
    & \includegraphics[width=.12\linewidth,valign=m]{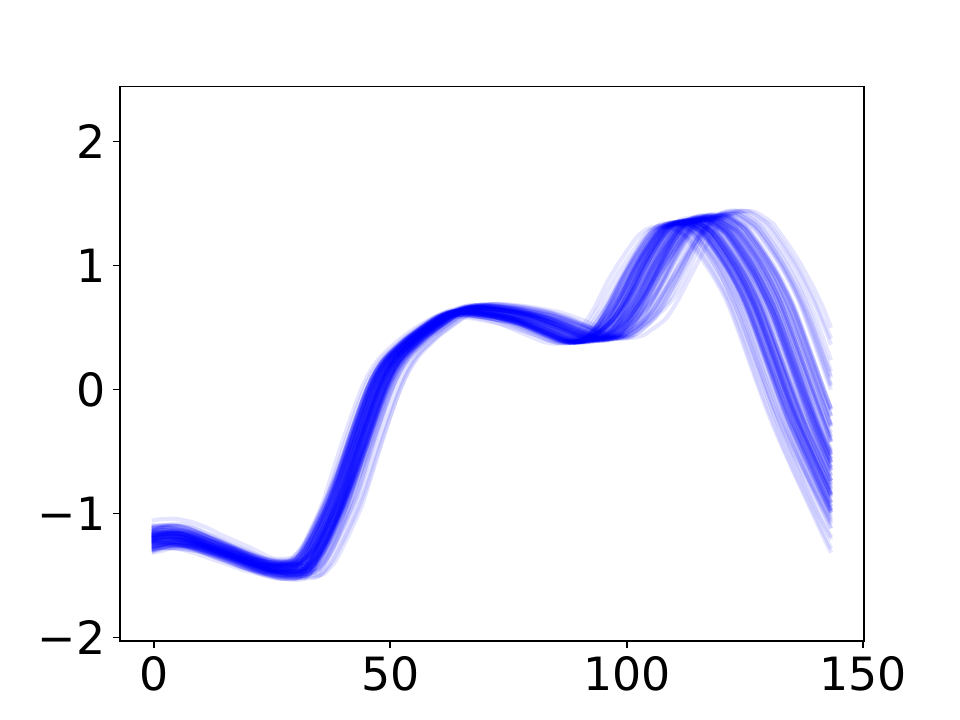}
    & \includegraphics[width=.12\linewidth,valign=m]{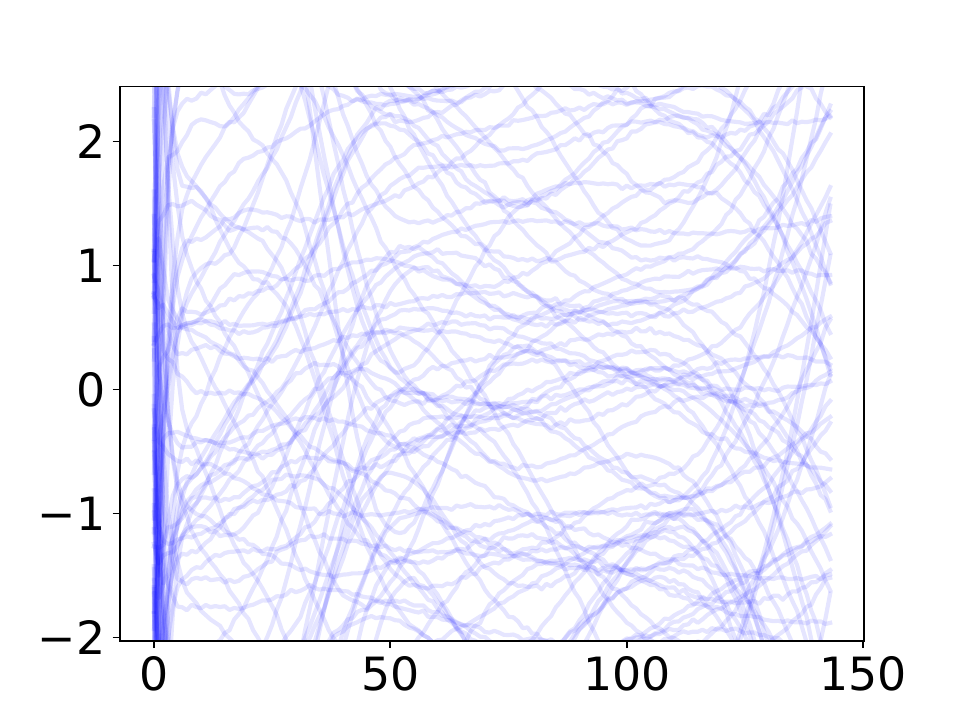}
    & \includegraphics[width=.12\linewidth,valign=m]{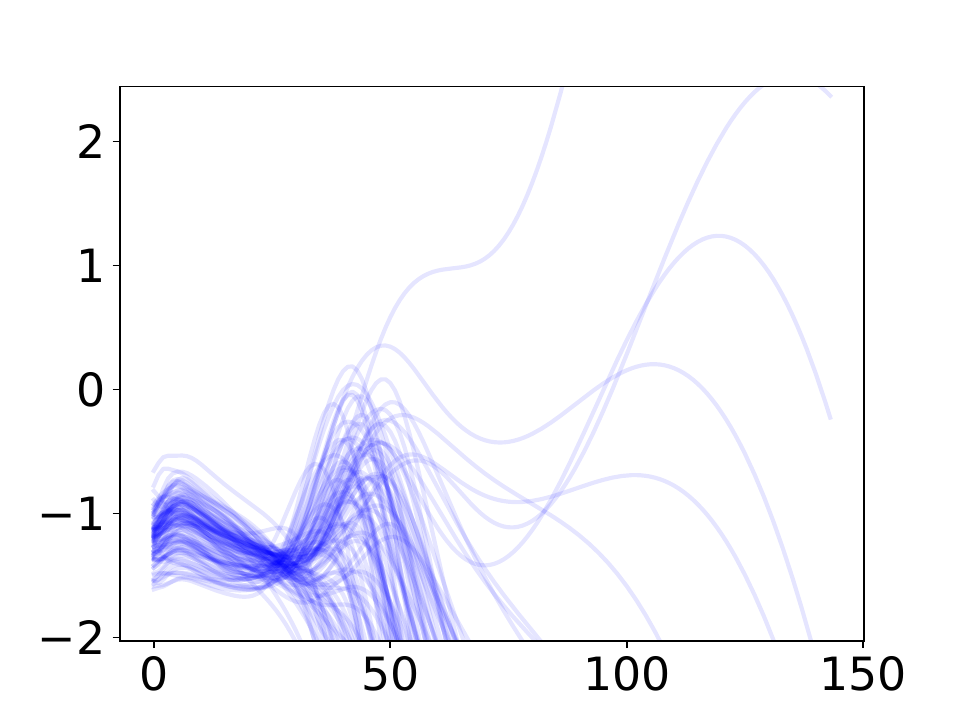}\\
    \midrule
    \end{tabular}
}
\label{tab:samples-visualised}
\end{table*}

\begin{table*}[!h]
\centering
\tiny
\caption{Test power of two-sample test comparing samples of the data with samples from the model. \textbf{Bold} indicates the best score (lower is better). The test power is averaged over $200$ trials with $10$ samples drawn from the dataset and model at significance level $\alpha = 0.05$.}
\begin{tabular}{@{}ccccc@{}}
                         \toprule
Dataset                         & \texttt{Quadratic} & \texttt{Melbourne} & \texttt{Stocks} & \texttt{GridWatch} \\ \midrule
$\mathcal{F}$-EBM (ours) & $\mathbf{0.09}$                               & $\bm{0.09}$                         & $\mathbf{0.08}$                  & $\mathbf{0.09}$                     \\
$\pi$-VAE                & $0.22$                               & $0.32$                              & $0.11$                  & ${0.16}$                            \\
Neural Process           & $0.59$                               & $0.95$                              & $1.00$                  & $1.00$                              \\
EBP                      & $0.322$                              & $0.52$                             & $0.31$                          & $0.56$                             \\
VIP                      & $\mathbf{0.09}$                      & $0.11$                              & $0.86$                           & $1.00$                                \\
Gaussian Process         & $1.00$                               & $0.77$                              & $0.67$                           & $1.00$                              \\ \bottomrule
\end{tabular}
\label{tab:test_power}
\end{table*}

First, in Section \ref{sec:calibration}, we begin by looking at the calibration of our proposal on draws from a noiseless Gaussian process. Then, in Section \ref{ref:benchmark}, we benchmark our model against existing models such as $\pi$-VAE, GP using the Mat\'{e}rn, NP, VIP (with a neural sampler, see \citet[Example 1]{ma2019variational}) and EBP (Gaussian model, see \citet[Section 3.2]{yang2020energy}) on synthetic and real data. Following that, in Section \ref{sec:exp_prior}, we compare the between different choices of $p_{\theta_{prior}}(Z)$. Then, in Section \ref{sec:generative_model}, we show how our proposal can be used as a resolution-independent generative model that can freely upscale and downscale images.The details for the experiments can be found in the Appendix \ref{sec:exp_settings}. The code can be found online at \url{https://github.com/jenninglim/functional-ebm}.

\begin{table*}[ht!]
\centering
\tiny
\caption{Comparison between the performance of $\mathcal{F}$-EBM for different priors on \texttt{Quadratic} dataset.}
\begin{tabular}{@{}cccllllllll@{}}
\toprule
\multicolumn{1}{l}{} & \multicolumn{1}{l}{} & \multicolumn{3}{c}{Downsample}                                                              & \multicolumn{3}{c}{Middle}                                                                                      & \multicolumn{3}{c}{Random}                                                                                      \\ \midrule
Prior Type           & Test Power           & $p=\frac{1}{4}$ & \multicolumn{1}{c}{$p=\frac{1}{3}$} & \multicolumn{1}{c}{$p=\frac{1}{2}$} & \multicolumn{1}{c}{$p=\frac{1}{4}$} & \multicolumn{1}{c}{$p=\frac{1}{2}$} & \multicolumn{1}{c}{$p=\frac{3}{4}$} & \multicolumn{1}{c}{$p=\frac{1}{4}$} & \multicolumn{1}{c}{$p=\frac{1}{2}$} & \multicolumn{1}{c}{$p=\frac{3}{4}$} \\ \midrule
Gaussian ($d_z=100$) & $1.00$               & ${0.006}$         & ${0.005}$                             & ${0.004}$                             & ${0.256}$                             & $\bm{0.006}$                             & $\bm{0.005}$                             & $\bm{0.006}$                             & $\bm{0.003}$                             & $\bm{0.003}$                             \\
EBM ($d_z=10$)       & $0.10$               & $0.029$         & $0.026$                             & $0.029$                             & $3.32$                              & $0.057$                             & $0.066$                             & $0.048$                             & $0.040$                             & $0.043$                             \\
EBM ($d_z=100$)      & $0.65$               & $3.337$          & $2.933$                              & $0.924$                             & $83.374$                             & $72.046$                             & $30.640$                             & $95.854$                             & $45.024$                             & $7.528$                              \\ \midrule
Tilted               & $\bm{0.08}$               & $\bm{0.005}$         & $\bm{0.004}$                             & $\bm{0.003}$                             & $\bm{0.044}$                             & $0.008$                             & $0.006$                             & $0.008$                             & $0.005$                             & $0.004$                             \\ \bottomrule
\end{tabular}
\label{table:prior}
\end{table*}

\subsection{Calibration}
\label{sec:calibration}
We are interested in examining the calibration of our proposed method. This is performed by visually inspecting the confidence intervals and its average predictive CDF as recommended in \citet[Section 3.2] {gneiting2007probabilistic}.

The dataset is composed of $100$ draws from a \textit{noiseless} GP with a Gaussian kernel where each function evaluated at $50$ distinct irregular points randomly sampled from $U(-5, 5)$. The baseline is a \textit{noisy} GP that shares the same kernel as the underlying ground-truth noiseless GP, but assumes a Gaussian observational noise in the likelihood. This baseline model provides a glimpse of the performance when you know the underlying functional process, but makes in incorrect assumption about the noise.

\begin{figure*}[!ht]
    \centering
    \includegraphics[width=\linewidth]{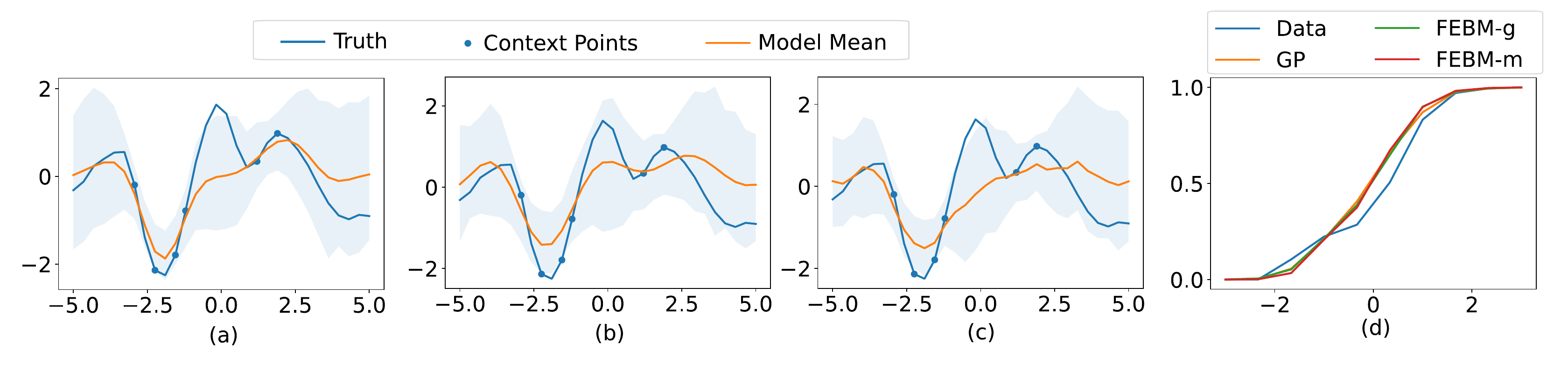}
    \caption{Calibration plots. We show the $95\%$ confidence interval generated from $300$ samples of (a) Gaussian Process, $\mathcal{F}$-EBM with (b) Gaussian, and (c)  Mat\'{e}rn kernel. In (d), we show the average predictive CDF of various models in comparison with the data.}
    \label{fig:uncertainty}
\end{figure*}

In Figure \ref{fig:uncertainty}, it can be seen that the confidence intervals and average predictive CDF (see Section $3.2$, Gneiting et al.\ 2007) of the proposed methods are similar in calibration to the GP. The average predictive CDF also hints the cost of assuming noisy observations.

\subsection{Benchmark}
\label{ref:benchmark}
 We utilize four datasets composed of one synthetic and three real-world datasets. The samples from the dataset can be seen in Table \ref{tab:samples-visualised}. The first \texttt{Quadratic} ($n=400$, $m=30$) is a bi-modal dataset composed of quadratic functions that was generated by sampling the sign of the quadratic term from a uniform distribution on $\{-1,1\}$. The dataset was designed to test the model's capacity to express bi-modality. We use $200$ samples for training and the remainder for evaluation. The second \texttt{Melbourne} ($n=1138$, $m=24$) is a real-life dataset where each sample is the number of pedestrians on a certain street in Melbourne recorded throughout different times of the day. We use $796$ samples for training and the remainder for evaluation. The third dataset \texttt{GridWatch} ($n=532$, $m=144$) is another real-life dataset where each sample is the demand of energy on the UK National Grid throughout different times of the day. We use $372$ samples for training and the remainder for evaluation. Finally, we use \texttt{Stocks} ($n=400$, $m=200$) where each sample corresponds to a stock highest performance on that day, recorded over $200$ days starting from February $2013$ (when data is available). The data is taken from S\&P 500. We use $200$ samples for training and the remainder for evaluation. See Appendix \ref{sec:preprocessing} for preprocessing details. These datasets are ``simple'', in the sense, that they are low dimensional, but they exhibit complex properties such as bimodality, multi-modality and large support. For these datasets, we chose to use a Gaussian likelihood.

\textbf{Evaluation.} We measure the quality of the trained model by considering its \textit{goodness-of-fit} and \textit{predictive error}. We compare our proposed model to a Gaussian process (GP) implemented in GPyTorch \citep{gardner2018gpytorch}, a Neural process (NP) which follows recommended practices \citep{le2018empirical}, and our own implementation of $\pi$-VAE, EBP, VIP (see Appendix \ref{sec:baseline} for details).

\begin{table*}[ht!]
\tiny
\caption{The mean of squared errors to measure the mismatch between the mean function and the true function on unseen points. The mean function is averaged from $100$ samples drawn from $p(f\,|\,{X}^c,{Y}^c)$. The error is computed on the evaluation pair and averaged over the evaluation dataset. \textbf{Bold} indicates the lowest score (lower is better). High indicates a value greater than $100$.}
\makebox[\textwidth]{
\begin{tabular}{@{}lcccccccccc@{}}
\toprule
                                    &                   & \multicolumn{3}{c}{Downsample}                      & \multicolumn{3}{c}{Middle}                          & \multicolumn{3}{c}{Random}                          \\ \midrule
Dataset                             & Model             & $p=\frac{1}{4}$ & $p=\frac{1}{3}$ & $p=\frac{1}{2}$ & $p=\frac{1}{4}$ & $p=\frac{1}{2}$ & $p=\frac{3}{4}$ & $p=\frac{1}{4}$ & $p=\frac{1}{2}$ & $p=\frac{3}{4}$ \\ \midrule

\texttt{Quadratic} & $\mathcal{F}$-EBM & $\bm{0.005}$    & $\bm{0.004}$    & $\bm{0.003}$    & $\bm{0.044}$    & $\bm{0.008}$    & ${0.006}$    & $\bm{0.008}$    & $\bm{0.005}$    & $\bm{0.004}$    \\
                                    & $\pi$-VAE         & $0.144$         & $0.167$         & $0.092$         & High            & $7.402$         & $0.025$         & $1.001$         & $0.330$         & $0.027$         \\
                                    & NP                & $7.431$         & $10.491$        & $9.447$         & $34.785$        & $15.596$        & $3.112$         & $8.048$         & $7.686$         & $7.599$         \\
                                    & EBP               & $0.039$         & $0.040$         & $0.034$         & ${0.474}$         & ${0.012}$         & $\bm{0.004}$         & $0.042$         & $0.016$         & $0.016$         \\
                                    & VIP               & $3.752$         & $5.070$         & $4.296$         & $2.710$         & $2.973$         & $2.611$         & $3.961$         & $4.422$         & $3.803$         \\
                                    & GP                & $28.089$        & $36.523$        & $19.934$        & High            & $75.350$        & $25.556$        & $70.742$        & $39.231$        & $8.275$         \\ \midrule

\texttt{Melbourne} & $\mathcal{F}$-EBM & $\bm{0.054}$       & $\bm{0.048}$       & $\bm{0.024}$    & $\bm{0.773}$    & $\bm{0.421}$    & $0.110$         & $0.275$         & $\bm{0.068}$       & $\bm{0.042}$         \\
                                    & $\pi$-VAE         & High            & $0.393$         & $0.195$         & High            & $32.294$        & $\bm{0.082}$    & $12.288$        & $0.860$         & ${0.093}$    \\
                                    & NP                & ${0.194}$    & ${0.179}$    & $0.174$         & $1.271$         & $1.016$         & $0.452$         & $\bm{0.231}$    & ${0.278}$    & $0.266$         \\
                                    & EBP               & $0.344$         & $0.400$         & $0.280$         & $0.961$         & $0.782$         & $0.421$         & $0.500$         & $0.474$         & $0.447$         \\
                                    & VIP               & $0,286$         & $0.410$         & $0.297$         & ${0.850}$         & $0.781$         & $0.446$         & $0.460$         & $0.491$         & $0.392$         \\
                                    & GP                & $0.424$         & $0.420$         & $0.213$         & $1.526$         & $1.450$         & $0.723$         & $1.047$         & $0.648$         & $0.508$         \\ \midrule
\texttt{Stocks}    & $\mathcal{F}$-EBM & $\bm{0.008}$    & $\bm{0.007}$    & $\bm{0.006}$    & ${0.023}$    & ${0.007}$    & ${0.006}$    & $\bm{0.008}$    & $\bm{0.006}$    & $\bm{0.005}$         \\
                                    & $\pi$-VAE         & $0.030$         & $0.030$         & $0.022$         & High            & High            & High            & High            & $0.023$         & ${0.011}$    \\
                                    & NP                & $0.039$         & $0.037$         & $0.037$         & $0.092$         & $0.040$         & $0.018$         & $0.038$         & $0.039$         & $0.041$         \\
                                    & EBP               & ${0.010}$             & ${0.009}$             & ${0.007}$             & $\bm{0.009}$             & $\bm{0.006}$             & $\bm{0.004}$             & ${0.011}$             & ${0.007}$             & ${0.006}$             \\
                                    & VIP               & $0.082$         & $0.103$         & $0.105$         & $0.100$         & $0.135$         & $0.082$         & $0.080$         & $0.100$         & $0.111$         \\
                                    & GP                & $0.053$         & $0.031$         & $0.017$         & $1.133$         & $1.081$         & $0.938$         & $0.135$         & $0.036$         & ${0.011}$    \\ \midrule

\texttt{GridWatch} & $\mathcal{F}$-EBM & $\bm{0.006}$    & $\bm{0.006}$    & $\bm{0.005}$    & $\bm{0.065}$    & $\bm{0.042}$    & $\bm{0.010}$    & $\bm{0.008}$    & $\bm{0.005}$    & $\bm{0.005}$    \\
                                    & $\pi$-VAE         & $0.376$         & $0.330$         & $0.249$         & High            & High            & High            & High            & $0.260$         & $0.122$         \\
                                    & NP                & $0.158$         & $0.155$         & $0.160$         & $0.162$         & $0.178$         & $0.123$         & $0.163$         & $0.166$         & $0.159$         \\
                                    & EBP               & $0.301$         & $0.378$         & $0.251$         & $0.900$         & $0.725$         & $0.353$         & $0.466$         & $0.465$         & $0.408$         \\
                                    & VIP               & $3.752$         & $5.070$         & $4.296$         & $2.710$         & $2.973$         & $2.611$         & $3.961$         & $4.422$         & $3.803$         \\
                                    & GP                & $0.338$         & $0.209$         & $0.106$         & $0.938$         & $0.581$         & $0.389$         & $0.567$         & $0.283$         & $0.156$         \\ \bottomrule
\end{tabular}
}
\label{tab:sse_loss}
\end{table*}

\textit{Goodness-of-fit}. It is desirable for our model to be a good fit to the data distribution. Since evaluating generative models is highly difficult, we consider the  three methods for evaluating the fit. (1) We visually inspect the samples generated from the model. (2) As a quantitative measure, we evaluate the model's fit by computing the test power of a functional non-parametric two-sample test to compare the synthetic samples and samples from the dataset. Since all models are wrong \citep{box1976science}, we expect the test to reject the null hypothesis that the learnt model is equal to the generating process of the dataset and so, the test power will indicate the fidelity of the model. We use the test proposed by \citet{wynne2022kernel}. (3) Following \citet{tgan}, we  visually evaluate this by plotting a $2$-dimensional (dimensionally-reduced) embeddings of both the samples generated by the model and taken from the dataset. A high degree of overlap between the data and model embedding's distribution suggests a good fit to the data distribution. For dimensionality reduction methods, we apply t-SNE \citep{van2008visualizing} and PCA \citep{bryant1995principal} on both the original data and the generated synthetic samples. For the sake of brevity, we defer these figures to Appendix \ref{sec:additonal_figures}.

\textit{Predictive error}. The predictive distribution is one of the central quantities of interest for NPs and GPs. We measure its fidelity with the difference between the mean function of the model and the underlying ``true'' function. Since the underlying function is usually unknown to us (except in synthetic settings), we split the dataset into disjoint sets, where one will be used to infer the function and the other is used to evaluate the conditional distribution. Given a portion $p\in(0,1)$, we consider three methods for splitting the dataset $(X, Y)$ to create the context pair $({X}^c, {Y}^c)$ and the evaluation pair $({X}^e, {Y}^e)$ such that the context pair has $p$ portion of the original data (and $1-p$ for the evaluation dataset). As shown in Figure \ref{fig:missing_setup}, the splits were chosen either (a) uniformly without replacement, (b) selected to be between sampled points in the infer dataset, or (c) down-sampled version of the original sampled points. We report the mean of squared errors averaged over the evaluation set., i.e, $\mathbb{E}_{X,Y} \left [ \frac{1}{|X^e|}\sum_{(x_i,y_i) \in ({X}^e, {Y}^e)} (\mathbb{E}[f(x_i)] - y_i)^2 \right ]$ where the inner expectation is estimated using samples from $p(f\,|\,{Y}^c;{X}^c)$.

\begin{figure*}[!ht]
    \centering
    \begin{subfigure}[b]{.32\linewidth}
        \includegraphics[width=\linewidth]{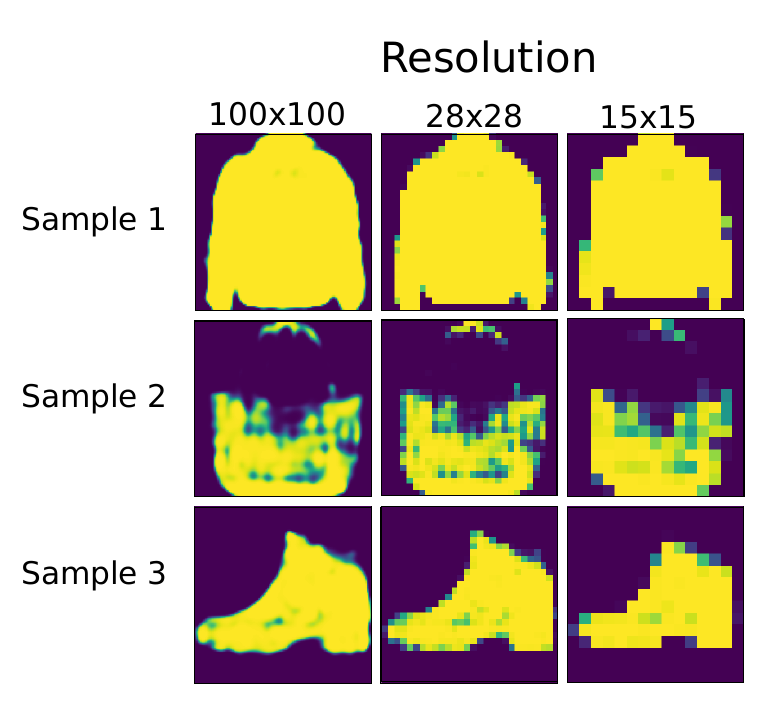}
        \caption{Scaled samples}\label{fig:mnist_scaled}
    \end{subfigure}
    \begin{subfigure}[b]{.65\linewidth}
        \includegraphics[width=\linewidth]{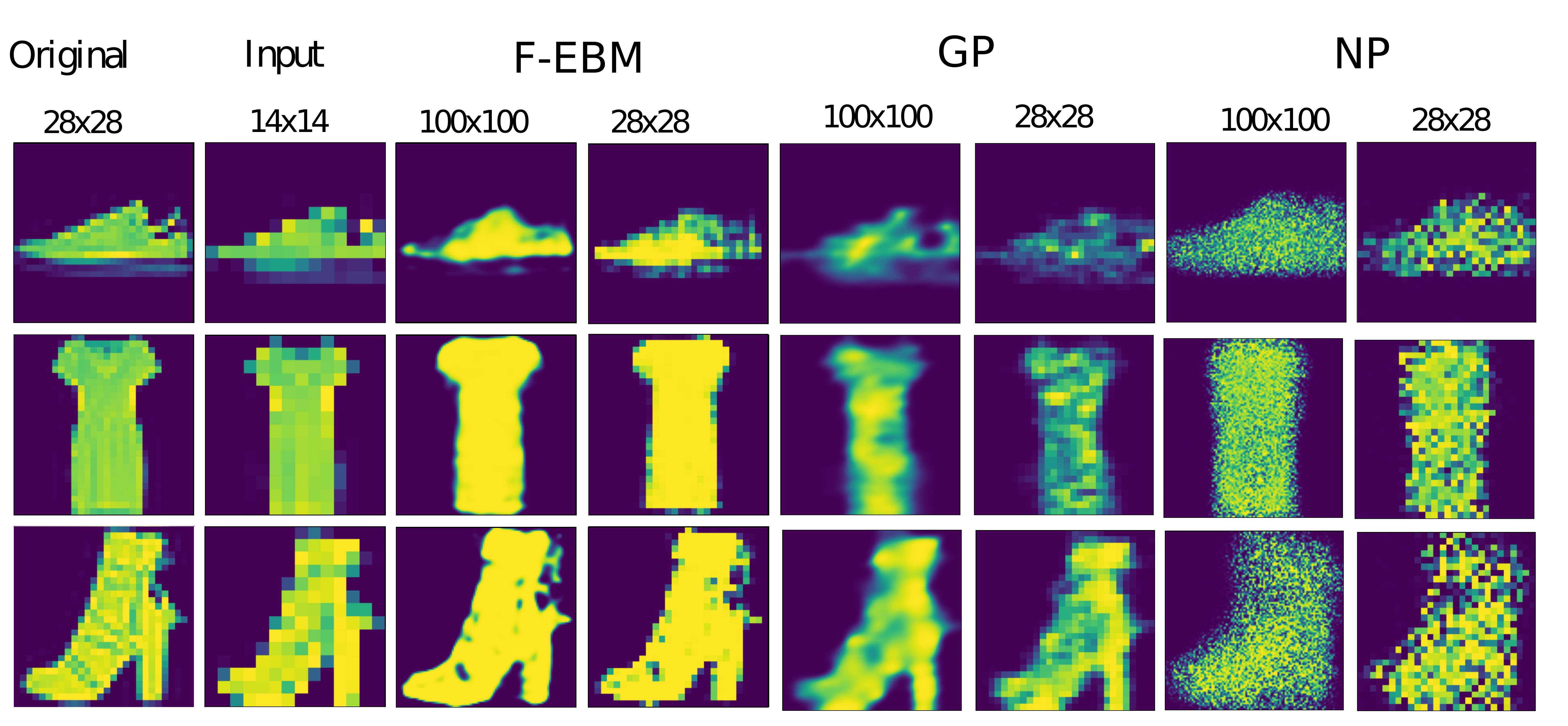}
        \caption{Upscaling Images}\label{fig:mnist_upscale}
    \end{subfigure}
    % \begin{subfigure}[b]{.25\linewidth}
    %     \includegraphics[width=\linewidth]{plots/mnist/samples.pdf}
    %     \caption{Model samples}\label{fig:mnist_samples}
    % \end{subfigure}
    \caption{Visualization of synthetic samples learnt from FashionMNIST. Figure \ref{fig:mnist_scaled} shows the three functions evaluated at different resolutions. Figure \ref{fig:mnist_upscale} displays upsampled samples generated by $\mathcal{F}$-EBM, GP and NP conditioned on the ``Input'' which is a downsampled variant the ``Original'' image.}
    \label{fig:mnist}
\end{figure*}
\textbf{Goodness-of-fit.} Table \ref{tab:samples-visualised} shows samples generated from the model and samples from the dataset. It is clear from the samples that both $\mathcal{F}$-EBM, $\pi$-VAE can capture the intricacies of each dataset better than both the NP, VIP. EBP performs well on \texttt{Quadratic}, but falls short in other datasets. For more complicated datasets, EBP, NP, and VIP can capture certain characteristics of the dataset (such as the mode) but cannot capture the tails of the data distribution. It is hard to distinguish between $\pi$-VAE and $\mathcal{F}$-EBM using only samples. Fortunately, the results of the test power of the two-sample test shown in Table \ref{tab:test_power} tells a different story. It can be seen that $\mathcal{F}$-EBM achieves the lowest test power for almost all datasets, with $\pi$-VAE being a close contender. These results suggest that $\mathcal{F}$-EBM is a good choice for generative modelling.

\textbf{Predictive Error.} Table \ref{tab:sse_loss} shows the prediction error on a range of datasets. Most competitors are designed for regression, and are competitive baselines for this problem. However, the performance of GP on \texttt{Quadratic} is noticeably poorer than others. This may be due to us not performing any pre-processing for \texttt{Quadratic}. For the GP, interpolating between regions where there are not many observe points the mean will tend towards the prior mean which will accumulate large errors in this problem (see Figure \ref{fig:missing_setup}). For other datasets where we pre-process the dataset, GP produces more competitive results. $\pi$-VAE also has high errors, particularly on the ``Middle'' split and small context pair on ``Random''. This is due to error from estimating the initial coefficients for small context pairs, which leads to high errors in the approximate posterior distribution. NPs and VIP performs well on all datasets and affirms that they are good choices for regression. EBP also performs well, particularly on the \texttt{Stocks} dataset. Overall, it can be seen that $\mathcal{F}$-EBM performs competitively across all datasets.

\subsection{Prior Choice}

\label{sec:exp_prior}

The choice of prior $p_{\theta_{prior}}$ is critical when designing generative models. For instance, in Variational Autoencoders (VAE) \citep{kingma2013auto}, a promising research direction for increasing the performance has been designing more expressive priors (for instance, see  \citet{aneja2021contrastive, tomczak2018vae}). In our case, we often found that the posterior $p_{\theta}(f|Z,X)$ could accurately reconstruct the underlying function, but sampling from a simple prior would result in a poor fit of the distribution.

In this section, we investigate the choice of prior on $\mathcal{F}$-EBM on the \texttt{Quadratic} dataset. We compare between a Gaussian prior, energy-based prior, and energy tilted prior whilst keeping other all hyperparameters equal. Specifically, we compare between $p_{\theta_{prior}}(z) \propto \exp (-\|z\|^2/2\sigma^2_0)$, $\exp (-E_\theta(z))$, and $\exp (-E_\theta(Z) - \|z\|^2/2\sigma^2_0)$ . We show the results in Table \ref{table:prior}. It can be seen that using a Gaussian prior was too simple and resulted in poor fit of the data as indicated with the high test power, but it performs the best on the predictive error tasks. As for the energy-based prior, it can be seen that the model performs significantly better at a lower dimension ($d_z=10$) than at higher dimensions ($d_z=100$). We suspect that this is due to the inherit difficulties of sampling in higher dimensional spaces for energy-based models and requires several tricks to work (see \citet{NEURIPS2019_378a063b}). The tilted energy-based prior approach was the most balanced and appeared to be able to combine the benefits of energy-based priors and a simple Gaussian prior for this example. We suspect that this can be attributed to the regularizing effect of the base distribution $p_0$ to create a distribution that is easier to sample from.

\subsection{Resolution-Independent Learning of Images}
\label{sec:generative_model}

One of the benefits of $\mathcal{F}$-EBM is its ability to learn complex functions. We demonstrate its capabilities on FashionMNIST dataset \citep{xiao2017/online}. In this setting, the inputs corresponds to the random fourier feature embedding \citep{tancik2020fourfeat} of the Cartesian coordinate of each pixel and the output is the pixel intensity. Since $Y$ is pixel data defined on $[0,1]$, we define the point-wise likelihood to be the continuous Bernoulli \citep{NEURIPS2019_f82798ec}.
It can be seen that in Figure \ref{fig:mnist_samples} that the model can produce diverse samples for shoes, shirts, and others. Unlike most generative models, our proposed approach can scale the images to arbitrary resolutions, as can be seen in Figure \ref{fig:mnist_scaled} and \ref{fig:mnist_upscale}. In Figure \ref{fig:mnist_scaled}, we demonstrate for each sample the model can upscale (or downscale) beyond the dataset's resolution, since each sample is a function that can be evaluated at any scale we desire. Figure \ref{fig:mnist_upscale} shows a sampled function from the model conditioned on a downsampled input from the dataset. The model can accurately infer the underlying function which can be used to produce high (or low) resolution images. For completeness, we show conditional samples for GP and NP. 

\section{CONCLUSION, LIMITATIONS AND FUTURE WORK}

We introduced $\mathcal{F}$-EBM, a class of models that are suitable for learning functional
data distributions. In addition, we have shown that $\mathcal{F}$-EBM can accurately perform inference over function space, which performs similarly or better than current approaches.

The most important limitations of our proposed work pertains to the choice of hyperparameters such as the kernel, truncation levels and architecture of the neural network. Although there is a large range of possible choices, we would like to emphasize that we did not perform a significant amount of hyperparameter tuning between problems and that we used the Mat\'ern kernel, and simple neural network for all our problems. We found that the most critical hyperparameters are those of the sampler in Eq.\ \ref{eq:loss}, and we believe significant gains can be made tuning this parameter. One potential avenue is to consider variational methods to tune this (similar to \citet{nijkamp2020learning}).

% Whilst its performance is strong, it is computationally more expensive than other methods. The main computational burden lies in the use of Contrastive divergence which requires us to run long MCMC chains to compute the loss function. We found that there are still gains to be had by tuning the CD algorithm and so, improving the CD algorithm itself could be a fruitful line of future work. On the other hand, there are other methods for training unnormalized distribution (such as score matching methods \citep{hyvarinen2005estimation}) that can also be considered.

An extension can be made to $\mathcal{F}$-EBM to perform conditional generative modelling by simply encoding an additional variable as input to the neural network. This extension will allow the model to learn conditional distributions that will be independent of the resolution of the data, which is usually what is applied to current models (for instance, see conditional GANs \citep{mirza2014conditional}).

\subsubsection*{Acknowledgements}
JNL gratefully acknowledges his PhD funding
from Feuer International Scholarship in AI. SJV acknowledges the support of Rheinland-Pfälzische Technische Universität and DFKI.

% \begin{thebibliography}{}
% \setlength{\itemindent}{-\leftmargin}
% \makeatletter\renewcommand{\@biblabel}[1]{}\makeatother
% \bibitem{} J.~Alspector, B.~Gupta, and R.~B.~Allen (1989).
%     \newblock Performance of a stochastic learning microchip.
%     \newblock In D. S. Touretzky (ed.),
%     \textit{Advances in Neural Information Processing Systems 1}, 748--760.
%     San Mateo, Calif.: Morgan Kaufmann.

% \bibitem{} F.~Rosenblatt (1962).
%     \newblock \textit{Principles of Neurodynamics.}
%     \newblock Washington, D.C.: Spartan Books.

% \bibitem{} G.~Tesauro (1989).
%     \newblock Neurogammon wins computer Olympiad.
%     \newblock \textit{Neural Computation} \textbf{1}(3):321--323.
% \end{thebibliography}

\bibliography{ref}
\bibliographystyle{abbrvnat}
% If you have textual supplementary material
\appendix
\onecolumn

\aistatstitle{\ourtitle: \\
Supplementary Materials}

\section{Mathematical Background}
\label{sec:math_background}
In this section, we provide some background on the mathematical formulation of stochastic processes as random variables on Hilbert spaces. An important class of stochastic process which are central to this work is the Gaussian process \citep{williams2006gaussian}.  Let $\mathcal{D} \subset \mathbb{R}^d$ be compact. Given a positive definite kernel $k:\mathcal{D}\times\mathcal{D}\rightarrow \mathbb{R}$ and a function $m:\mathcal{D}\rightarrow \mathbb{R}$ we say $x$ is a Gaussian process with mean function $m$ and covariance function $k$ if for every finite collection of points $\{s_{n}\}_{n=1}^{N}$ the random vector $(x(s_{1}),\ldots,x(s_{N}))$ is a multivariate Gaussian random variable with mean vector $(m(s_{1}),\ldots,m(s_{N}))$ and covariance matrix $k(s_{n},s_{m})_{n,m=1}^{N}$. The mean function and covariance function completely determines the Gaussian process.  We write $x\sim\mathcal{GP}(m,k)$ to denote the Gaussian process with mean function $m$ and covariance function $k$. 

Let $X$ be a real, separable Hilbert space with inner product  $\langle\cdot,\cdot\rangle$ and norm $\lVert\cdot\rVert$.  Let $\mathcal{B}(X)$ be the Borel $\sigma$-algebra on $X$.  A measure $\mu$ on $(X, \mathcal{B}(X))$ is said to be Gaussian if there exists $m \in X$ and a linear operator $C$ such that the push-forward ${l_h}_{*}\mu$ is a Gaussian measure on $(\mathbb{R}, \mathcal{B}(\mathbb{R}))$ with mean $\langle h, m\rangle$ and variance $\langle h, C h\rangle$, for all $h \in H$ where $l_h(\cdot) = \langle h, \cdot \rangle$.  The element $m\in X$ is called the \emph{mean} and $C$ is the covariance operator, which is a symmetric non-negative operator with finite trace.  The characteristic functional of a Gaussian measure $\mu$ on $X$ satisfies
$$
\widehat{\mu}(\lambda) = \int_{X} e^{i\langle \lambda, x\rangle}\mu(dx) =  e^{i\langle m, h\rangle -\frac{1}{2}\langle Ch,h\rangle}.
$$  
It is therefore uniquely determined by $m$ and $C$.  We therefore use the notation $\mu \equiv N_{m,C}$.  See \citep{DaPrato2006} for further properties of Gaussian measures on Hilbert spaces.
\\\\
Gaussian processes can be uniquely associated to a Gaussian measure on the $X = L^2(\mathcal{D})$.  Indeed, a Gaussian process $\mathcal{GP}(m, k)$ on $\mathcal{D}$, where $m \in L^2(\mathcal{D})$ and $k$ is positive-definite continuous, is characterized by a Gaussian measure on $X$ with mean $m$ and covariance operator defined by
$$
    Cf(\cdot) = \int k(\cdot, y)f(y)\,dy, \quad \forall f \in L^2(\mathcal{D}).
$$
It can be shown that the covariance $C$ is a positive trace class operator which admitting the spectral decomposition $C\cdot = \sum_{i=1}^\infty \lambda_i e_i \langle e_i, \cdot\rangle$, where $\lbrace \lbrace \lambda_i\rbrace_{i=1}^\infty, \lbrace e_i\rbrace_{i=1}^\infty \rbrace$ is the eigensystem associated with the operator $C$.  Moreover we can write the kernel as $k(x,y) = \sum_{i=1}^\infty \lambda_i e_i(x)e_i(y)$, for all $x, y \in \mathcal{D}$.
\\\\
Given a second-order stochastic process $x(\cdot)$ taking values in $L^2(\mathcal{D})$ the Karhunen-Loeve expansion provides  an important characterization.   Suppose that the point-wise covariance $k(s,t) = \mathrm{Cov}[x(s), s(t)]$ is continuous, and the mean function defined by  $t\rightarrow m(t)) = \mathbb{E}[x(t)]$ lies in $L^2(\mathcal{D})$.   Let $C$ be the non-negative definite trace class covariance operator associated to $k$, and let $\lbrace \lbrace \lambda_i\rbrace_{i=1}^\infty, \lbrace e_i\rbrace_{i=1}^\infty \rbrace$ eigensystem.   The Karhunen-Loeve expansion \citep[Theorem 11.4]{Sullivan2015} for the random process $x(\cdot)$ is given by
$$
    x(\cdot) \sim m + \sum_{i=1}^\infty \sqrt{\lambda}_i \xi_i e_i(\cdot),
$$
where $\lbrace \xi_i \rbrace_{i=1}^\infty$ are unit-variance uncorrelated random variables.  In the special case where the process $x(\cdot)$ is Gaussian then the $\xi_i$ are actually independent standard Gaussian random variables, yielding a convenient means of generating realizations of the process, when the eigensystem of the covariance is available.
 
The model proposed in Section 3 is characterized by the density of the joint distribution of $Y=(f(x_1), \ldots, f(x_M))$, for a set of evaluation points $X=(x_1, \ldots, x_M)$ in $\mathcal{X}$ defined by
\begin{align*}
       p_{\theta}(y_1,\ldots, y_M | x_1,\ldots, x_M) &\propto \int p_{\theta}(Y, z| X)\exp( -E_{\theta_{prior}}(z))p_0(dz) \\
       &= \int \prod_{i=1}^M p_{\theta}(y_i | x_i, z)\exp( -E_{\theta_{prior}}(z))p_0(dz).
\end{align*}
To show that this characterizes a stochastic process indexed by $\mathcal{X}$ we verify Kolmogorov's consistency conditions.  Firstly, note that, for any permutation $\pi_1,\ldots, \pi_M$ of $1,\ldots, M$, then 
\begin{align*}
   p_{\theta}(y_{\pi_1},\ldots, y_{\pi_M} | x_{\pi_1},\ldots, x_{\pi_M}) &= \frac{1}{Z}\int \prod_{i=1}^M p_{\theta}(y_{\pi_i} | x_{\pi_i}, z)\exp( -E_{\theta_{prior}}(z))p_0(dz), \\
   &= \frac{1}{Z}\int \prod_{i=1}^M p_{\theta}(y_{i} | x_{i}, z)\exp( -E_{\theta_{prior}}(z))p_0(dz) \\
   &= p_{\theta}(y_{1},\ldots, y_{M} | x_{1},\ldots, x_{M}),
\end{align*}
where $Z = \int\cdots\int \int \prod_{i=1}^M p_{\theta}(y_{i} | x_{i}, z)p_0(dz)\,dy_1\ldots dy_M$.

Secondly, we show that the finite dimensional distributions are consistent with respect to marginalization.  To this end, consider
\begin{align*}
\int p_{\theta}(y_1, \ldots, y_{M+1}\, | \, x_1, \ldots, x_{M+1}) \,dy_{M+1} = \int \frac{1}{Z}\int \prod_{i=1}^{M+1} p_{\theta}(y_{i} | x_{i}, z)\exp( -E_{\theta_{prior}}(z))p_0(dz)\,dy_{M+1} \\
= \int \frac{1}{Z}\int p_{\theta}(y_{{M+1}} | x_{{M+1}}, z) \prod_{i=1}^M p_{\theta}(y_{i} | x_{i}, z)\exp( -E_{\theta_{prior}}(z))p_0(dz)\,dy_{M+1}\\
=  \frac{1}{Z}\int \int p_{\theta}(y_{{M+1}} | x_{{M+1}}, z) \,dy_{M+1} \prod_{i=1}^M p_{\theta}(y_{i} | x_{i}, z)\exp( -E_{\theta_{prior}}(z))p_0(dz)\\
=  \frac{1}{Z}\int  \prod_{i=1}^M p_{\theta}(y_{i} | x_{i}, z)\exp( -E_{\theta_{prior}}(z))p_0(dz) \\
= p_{\theta}(y_1,\ldots, y_M \, | x_1, \ldots, x_M).
\end{align*}
These two conditions establish that the proposed model defines a valid stochastic process. 
\label{sec:related_work}

\section{Derivations}
\subsection{Gradients of Maximum Likelihood}

Given a latent EBM defined as
$$
p_\theta(y, z| X) = \frac{1}{C(X)} \exp (-E_\theta(y,z|X)),
$$
where $C(X) = \int \exp (-E(y,z|X)) dy dz$. The marginal is given by $p_\theta (y| X) = \int p_\theta (y, z| X) dz$. The type-II likelihood for a given dataset $(\bm{X}, \bm{Y})$ is
$$
\prod_{(x,y) \in (\bm{X},\bm{Y})} p_\theta (y | x).
$$
Maximizing the type-II likelihood is equivalent to minimizing the following loss $\mathcal{L}$:
$$
\mathcal{L}(\theta) = - \mathbb{E}_{(x,y) \sim p (\bm{X}, \bm{Y})} \log p_\theta (y | x),
$$
where the expectation w.r.t $p(\bm{X}, \bm{Y})$ is an expectation w.r.t.\ the empirical distribution. We will show that its gradient is given by
$$
\nabla  \mathcal{L}(\theta) = \mathbb{E}_{(x,y) \sim p(\bm{X}, \bm{Y})}\mathbb{E}_{Z \sim  p_\theta(z| y, x)}[\nabla E(y, Z| x)] - \mathbb{E}_{x \sim p(\bm{X})}\mathbb{E}_{y,z \sim  p_\theta(y,z|X)} \left [ \nabla E(y,z|X) \right ].
$$
To see this first, we have
\begin{align*}
    \nabla \mathcal{L}(\theta) &= - \mathbb{E}_{(x,y) \sim p(\bm{X}, \bm{Y})} \left [ \frac{ \nabla p_\theta(y | x)}{ p_\theta(y | x)} \right ].
\end{align*}
Then, we have
\begin{align*}
    \frac{ \nabla p_\theta(y | x)}{ p_\theta(y | x)} &= \frac{ \nabla \int p_\theta(y, z| x) dz}{ p_\theta(y | x)} \\
    &= \frac{ \int p_\theta(y, z| x) \nabla \log p_\theta(y, z| x) dz}{ p_\theta(y | x)} \\
    &= \int p_\theta(z| y, x) \nabla \log p_\theta(y, z| x) dz \\ 
    &= \mathbb{E}_{Z \sim  p_\theta(z| y, x)}[\nabla \log p_\theta(y, Z| x)].
\end{align*}
So we have
\begin{align}
    \nabla \mathcal{L}(\theta) &= - \mathbb{E}_{(x,y) \sim p(\bm{X}, \bm{Y})} \mathbb{E}_{Z \sim  p_\theta(z| y, x)}[\nabla \log p_\theta(y, Z| x)].
    \label{eq:grad_loss1}
\end{align}
Note that
\begin{equation}
    \nabla \log p_\theta(y, Z| x) = - \nabla E (y, Z | x) - \nabla \log C(x),
    \label{eq:grad_loss2}
\end{equation}
and
\begin{align*}
\nabla \log C(x) &=  \nabla \log  \int \exp (-E(y,z|X)) dy dz \\
&= \frac{\nabla \int \exp (-E(y,z|X)) dy dz}{\int \exp (-E(y,z|X)) dy dz} \\
&= - \frac{\int  \exp (-E(y,z|X)) \nabla E(y,z|X) dy dz }{\int \exp (-E(y,z|X)) dy dz} \\
&= - \int  p_\theta(y,z|X)\nabla E(y,z|X) dy dz \\
&= - \mathbb{E}_{y,z \sim  p_\theta(y,z|X)} \left [ \nabla E(y,z|X) \right ].
\end{align*}
Putting this into Eq.\ \ref{eq:grad_loss2} and Eq.\ \ref{eq:grad_loss1}, we have as desired.

\label{sec:derivation_gradient_ml}

\subsection{Gradients of $\mathcal{F}$-EBM}
\label{sec:derivation_gradient}
Recall that the gradient of the marginal likelihood can be written as
\begin{align*}
\nabla  \mathcal{L}(\theta) = \mathbb{E}_{(X,Y) \sim p(\bm{X}, \bm{Y})}\mathbb{E}_{Z \sim  p_\theta(Z| Y, X)}[\nabla E(Y, Z| X)] - \mathbb{E}_{X \sim p(\bm{X})}\mathbb{E}_{Y,Z \sim  p_\theta(Y,Z|X)} \left [ \nabla E(Y,Z|X) \right ].
\end{align*}
and we have for the tilted model
\begin{align*}
    {E}_\theta(Y, Z; X) &= - \log p_{\theta_{gen}}(Y|Z,X) - \log p_{\theta_{prior}}(Z) \\
    &= - \log p_{\theta_{gen}}(Y|Z,X) + E_{\theta_{prior}}(Z) - \log p_0(Z).
\end{align*}
Thus, the gradient w.r.t $\theta_{gen}$ is given by
\begin{align*}
\nabla_{\theta_{gen}}\mathcal{L}(\theta) &= \mathbb{E}_{(X,Y) \sim p(\bm{X}, \bm{Y})}\mathbb{E}_{Z \sim  p_\theta(Z\,|\,Y)} \nabla_{\theta_{gen}}E_\theta(Y, Z|X)
    - \mathbb{E}_{X \sim p(\bm{X})} \mathbb{E}_{(Y, Z) \sim p_\theta(Y, Z|X)}\nabla_{\theta_{gen}}E_\theta(Y, Z| X), \\
    &\overset{(a)}{=} - \mathbb{E}_{(X,Y) \sim p(\bm{X}, \bm{Y})}\mathbb{E}_{Z \sim  p_\theta(Z\,|\,Y)} \nabla_{\theta_{gen}}\log p_{\theta_{gen}}(Y|Z,X)
    + \mathbb{E}_{X \sim p(\bm{X})}\mathbb{E}_{(Y, Z) \sim p_\theta(Y, Z)}\nabla_{\theta_{gen}}\log p_{\theta_{gen}}(Y|Z,X), \\
    &= - \mathbb{E}_{(X,Y) \sim p(\bm{X}, \bm{Y})}\mathbb{E}_{Z \sim  p_\theta(Z\,|\,Y)} \nabla_{\theta_{gen}}\log p_{\theta_{gen}}(Y|Z,X),
\end{align*}
where (a) we use the fact that the second term is zero. To see this note that, we have for all $X$
\begin{align*}
    \mathbb{E}_{(Y, Z) \sim p_\theta(Y, Z; X)}\nabla_{\theta_{gen}}\log p_{\theta_{gen}}(Y|Z,X) &=  \int p_\theta(y,z|X)\nabla_{\theta_{gen}}\log p_{\theta_{gen}}(y|z,X)dydz \\
    &= \int p_\theta(y,z|X)\left [\nabla_{\theta_{gen}}\log p_{\theta}(y,z|X) -\nabla_{\theta_{gen}} \log p_{\theta_{prior}}(z) \right ]dydz \\
    &= \int p_\theta(y,z|X)\frac{\nabla_{\theta_{gen}}p_{\theta}(y,z;X)}{p_{\theta}(y,z|X)} dydz = \int \nabla_{\theta_{gen}}p_{\theta}(y,z|X) dydz = \nabla_{\theta_{gen}} 1 \\
    &= 0.
\end{align*}
For the gradient w.r.t $\theta_{prior}$, we have
\begin{align*}
\nabla_{\theta_{prior}}\mathrm{CD}(\theta) &= \mathbb{E}_{(X,Y) \sim p(\bm{X}, \bm{Y})}\mathbb{E}_{Z \sim  p_\theta(Z\,|\,Y)} \nabla_{\theta_{prior}}E_\theta(Y, Z|X)
    - \mathbb{E}_{X \sim p(\bm{X})}\mathbb{E}_{(Y, Z) \sim p_\theta(Y, Z |X )}\nabla_{\theta_{prior}}E_\theta(Y, Z| X), \\
    &= \mathbb{E}_{(X,Y) \sim p(\bm{X}, \bm{Y})}\mathbb{E}_{Z \sim  p_\theta(Z\,|\,Y)} \nabla_{\theta_{prior}} E_{\theta_{prior}}(Z)
    - \mathbb{E}_{Z \sim p_{\theta_{prior}}(Z)}\nabla_{\theta_{prior}} E_{\theta_{prior}}(Z).
\end{align*}

\section{Experiment details}
\label{sec:exp_settings}

The code was implemented in PyTorch \citep{paszke2019pytorch}. All the experiments were either performed on the CPU or on a RTX 2060 Super. All $\mathcal{F}$-EBM models took a maximum of 4 hours to train.

\subsection{Dataset availability}
The datasets can be found online for \texttt{Melbourne} (\url{https://tinyurl.com/rr4vu2b3}) and $\texttt{GridWatch}$ (7\url{https://www.gridwatch.templar.co.uk/}). The stocks datasets were collected by querying Investors Exchange (IEX).

\subsection{Training and design choices $\mathcal{F}$-EBM}
For all our experiments, we use the same neural architecture and kept all the training settings the same, except from the parameters of the Langevin sampler and latent dimension.
\begin{itemize}
    \item \textbf{Architecture}. For the mapping $\mu_\theta$, we use a three hidden layered neural network with $512$ hidden units and ReLU activations. We utilize skip connections between the first and second hidden layer, as well as the second and third layer. For the $E_{\theta_{prior}}(x)$, we use a similar neural network with $512$ hidden layers and ReLU activations with similar skip connections.
    \item \textbf{Kernel}. We use a Mat\'ern kernel and limit the number of estimated eigenfunction and eigenvalues to be equal to $m$. Recall that $m$ is the number of evaluation points of each function.
    \item \textbf{Optimization}. We use Adam optimizer with learning rate set to $10^{-3}$ for both $\theta_{gen}$ and $\theta_{prior}$ with all other parameters kept as default. We reduce the learning rate on plateau by a factor of $0.9$ and $0.8$ for $\theta_{gen}$ and $\theta_{prior}$ respectively until a minimum of $10^{-4}$ and $10^{-5}$ respectively. We use a batch size of $128$ and was run for $500$ epochs with early stopping. 
    \item \textbf{Langevin Dynamics.} Contrastive divergence is calculated by samples generated from stochastic gradient Langevin dynamics. For \texttt{Quadratic}, we use a variable step size that linearly interpolated starting at $10^{-2}$ to $10^{-3}$ for $100$ steps. As for the others (\texttt{Melbourne}, \texttt{Stocks}, and \texttt{Gridwatch}), we use constant step sizes of $10^{-2}$ for $100$ steps.
    \item \textbf{Latent Dimension $d_z$.} We use $d_z=100$ for \texttt{Quadratic} and $d_z=20$ for the others (\texttt{Melbourne}, \texttt{Stocks}, and \texttt{Gridwatch}). 
\end{itemize}

\subsection{Pre-processing}
\label{sec:preprocessing}
We detail the pre-processing applied to each dataset.
\begin{itemize}
    \item \texttt{Quadratic}. We do not perform any preprocessing for this dataset.
    \item \texttt{Stocks}. We normalize the dataset to have zero mean and unit variance.
    \item \texttt{Gridwatch}. We normalize each sample to have zero mean and unit variance.
    \item \texttt{Melbourne}. We normalize the dataset to have zero mean and unit variance.
\end{itemize}

\subsection{Baselines}
\label{sec:baseline}
\begin{itemize}
    \item \textbf{Gaussian process}. We use the same kernel as $\mathcal{F}$-EBM, with its parameters obtained from maximizing the marginal likelihood.
    
    \item \textbf{Neural process}. We use a modified implementation of neural processes available online: \url{https://github.com/EmilienDupont/neural-processes}. We set the dimension of the representation of the context points to be $512$ and the dimension of the latent variable to be $512$. The encoder and decoder was a $3$-layer neural network with $512$ hidden units with ReLU activations with the same skip connection. We trained the model up to $1000$ epochs with early stopping.
    
    \item \textbf{$\pi$-VAE}.
    We set the intermediate dimension of the encoder and decoder networks to $100$ with ReLU activation, and the dimension of the latent variable to $5$. For each dataset, the model is trained with early stopping with patience $100$. We use the Adam optimizer with a learning rate set to $5\times 10^{-3}$. As feature map $\Phi(\cdot)$ we use a basis representation given by $\Phi(\cdot)=(\phi_1(\cdot), \ldots, \phi_B(\cdot))$ where $\{\phi_i(\cdot)\}_{i=1}^B$ is a set of $B$ basis functions. For the basis, we compare a Mat\'ern kernel induced basis ($m$ estimated basis functions) with a FPCA based basis ($0.99$ explained variability) and consequently choose the Matérn kernel induced basis for all datasets except Melbourne. The initial coefficients for the interpolation are estimated from the context pairs via least squares.
    \item \textbf{Energy based process (EBP)}. We implement the Gaussian energy-based process \citep[See 3.2]{yang2020energy}. This is a Gaussian process with a kernel $k(x,x') = \phi_\theta(x)^\top \phi_\theta(x')$. Its marginal likelihood is tractable and $\phi_\theta$ be directly optimized \citep[Section 3.2 \& Appendix B.1]{yang2020energy}. 
    \item \textbf{Variational Implicit processes (VIP)}. We use the neural sampler VIP \citep[Example 1]{ma2019variational} is a $3$-layered neural network with $512$ hidden units, with Sigmoid Linear Unit (SiLU) activations and skip connections. The key difference between our implementation and the approach of \citet{ma2019variational} is in the wake phase. We obtain the parameters of the neural sampler by optimizing the marginal likelihood of the optimal GP from the sleep phase. In \citet{ma2019variational}, they instead optimized an approximation of the marginal log likelihood.
\end{itemize}

\section{Approximating Eigenfunctions}

\label{sec:interpolation}

\begin{table}[]
\centering
\begin{tabular}{@{}ccccccccccc@{}}
\toprule
\multicolumn{1}{l}{} & \multicolumn{1}{l}{} & \multicolumn{3}{c}{Downsample}                      & \multicolumn{3}{c}{Middle}                          & \multicolumn{3}{c}{Random}                          \\ \midrule
Kernel               & Test Power           & $p=\frac{1}{4}$ & $p=\frac{1}{3}$ & $p=\frac{1}{2}$ & $p=\frac{1}{4}$ & $p=\frac{1}{2}$ & $p=\frac{3}{4}$ & $p=\frac{1}{4}$ & $p=\frac{1}{2}$ & $p=\frac{3}{4}$ \\ \midrule
Gaussian             & 0.09                 & 1.170           & 1.160           & $1.089$         & $1.169$         & $1.126$         & $1.049$         & $1.116$         & $1.114$         & $1.085$         \\
Mat\'{e}rn               & 0.09                  & 1.175           & 1.166           & 1.094           & 1.173           & 1.137           & 1.050           & 1.177           & 1.119           & 1.086           \\ \midrule
Gaussian*            & 0.10                 & 1.150           & 1.148           & 1.077           & 1.160           & 1.122           & 1.038           & 1.146           & 1.101           & 1.082           \\
Mat\'{e}rn*              & 0.10                 & 1.184           & 1.180           & 1.105           & 1.197           & 1.140           & 1.045           & 1.175           & 1.127           & 1.096           \\ \midrule GP         & 0.08                 & 1.062           & 1.054           & 1.038           & 1.075           & 1.046           & 1.005           & 1.079           & 1.042           & 1.041           \\ \bottomrule
\end{tabular}
\caption{Performance of the proposed method using kernel ridge regression (indicated by the lack of *) and neural network (indicated by the *) on noiseless Gaussian process dataset used in Section \ref{sec:calibration}.}
\label{tab:nn_interpolation}
\end{table}
Nystr\"{o}m method produces an estimator for the eigenvalues and eigenfunctions that solve the problem
\begin{equation}
    \lambda_i e_i(t) = \int_\mathcal{X} k(t, x)e_i(x)dp(x) \approx \frac{1}{l}\sum_{j=1}^l k(t, x_j)e_i(x_j),
\end{equation}
for some $X:=\{x_i\}_{i=1}^l$.

Recall that  $K(X, X) = (k(x_i,x_j)_{ij}) \in \mathbb{R}^{l \times l}$ is the gram matrix, and $\hat{\bm{e}}_i(X) = [\hat{e}_i(x_1), \ldots, \hat{e}_i(x_l)]^\top \in \mathbb{R}^l$. The Nystr\"{o}m's eigenfunction estimator is given by
\begin{align}
    \tilde{e}_i(t) &:= \frac{1}{l\hat{\lambda}_{i}}\sum_{j=1}^l k(t, x_j)\hat{e}_i(x_j),
    \label{eq:nysestimator}
\end{align}
where the eigensystem $\{\hat{\lambda}_i, \hat{\bm{e}}_i(X)\}_{i=1}^l$ are the eigenvalues and eigenvectors of $\frac{1}{l} K({X}, {X})$ with its eigenvectors $v$ scaled such that $\|v\|^2_2 = l$. The deviation of the standard normalization factor of one is to ensure that
$$
\int \hat{e}_i(x) \hat{e}_i(x) p(x) dx \approx \frac{1}{l}\sum_{j=1}^l \hat{e}_i (x_j) \hat{e}_i (x_j) = \frac{1}{l}\|\hat{\bm{e}}_i(X)\|^2_2 = 1. 
$$
We found that the estimator in Eq. \ref{eq:nysestimator} produced accurate estimates for large eigenvalues, but for the smaller eigenvalues, it was a poor fit. Instead, one can treat this as a regression problem with inputs $X$ and $\{\hat{\bm{e}}_i(X)\}_{i=1}^l$ and utilize effective regression algorithms. For instance, for the majority of this work, we use kernel ridge regression. However, other models can be used instead, such as neural networks with least squares minimization. In Table \ref{tab:nn_interpolation}, we compare the performance between using neural networks trained by minimizing least squares and kernel ridge regression and found negligible differences on the noiseless Gaussian process dataset used in Section \ref{sec:calibration}.

\section{Additional Figures}
\label{sec:additonal_figures}

\begin{figure*}[h]
\tiny
    \centering
    \begin{center}
        \includegraphics[width=0.7\linewidth]{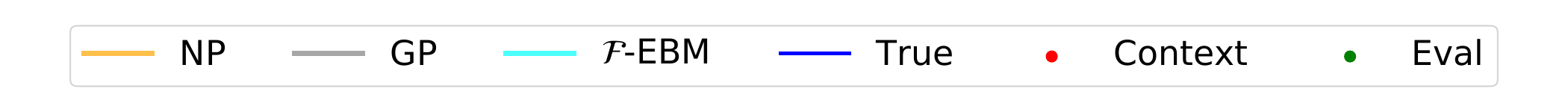}
    \end{center}
    \vspace{-8mm}
    \begin{subfigure}[b]{.25\linewidth}
        \includegraphics[width=\linewidth]{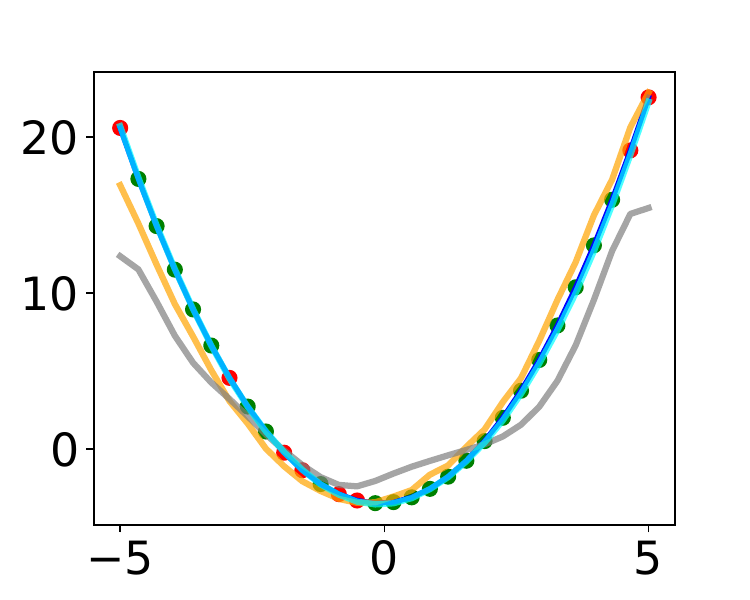}
        \caption{Random}\label{fig:random}
    \end{subfigure}
    \begin{subfigure}[b]{.25\linewidth}
        \includegraphics[width=\linewidth]{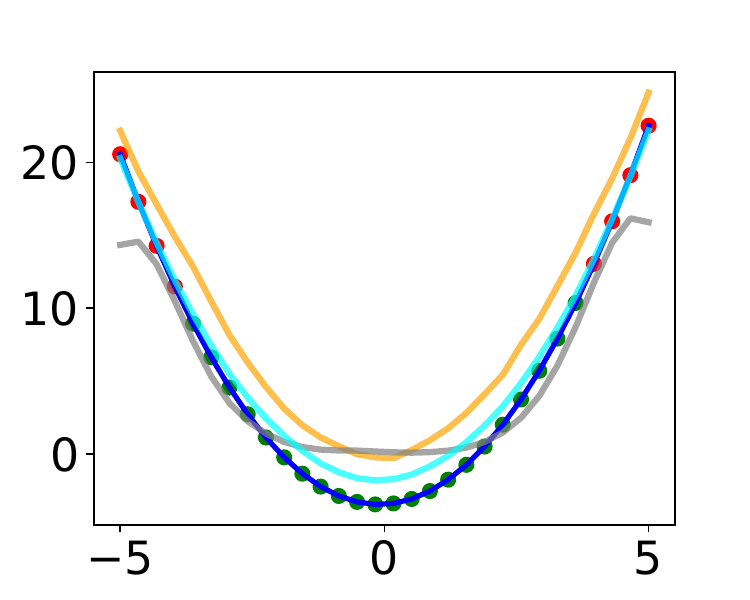}
        \caption{Middle}\label{fig:middle}
    \end{subfigure}
    \begin{subfigure}[b]{.25\linewidth}
        \includegraphics[width=\linewidth]{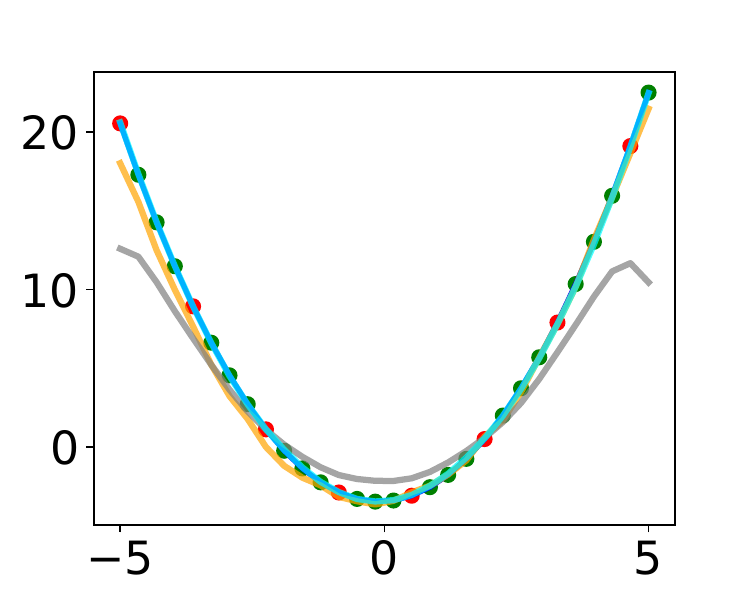}
        \caption{Down-sample}\label{fig:downsample}
    \end{subfigure}
    \caption{A visualization of the three methods used for splitting evaluations $(X,Y)$ of a quadratic function with $p=\frac{1}{4}$. The context pair $({X}^c,{Y}^c)$ shown in \textcolor{red}{red} used to compute the mean function for various models ($\mathcal{F}$-EBM, GP and NP) which is evaluated on the evaluation pair $({X}^e,{Y}^e)$ shown in \textcolor{green}{green}.}
    \label{fig:missing_setup}
\end{figure*}

\begin{table}[ht!]
\centering
\caption{The PCA embeddings of $100$ samples of both samples generated from the model (\textcolor{blue}{blue}) and samples from the data (\textcolor{red}{red})}
\begin{tabular}{@{}cccccc@{}}
\toprule
 & $\mathcal{F}$-EBM (ours) & $\pi$-VAE & NP & GP\\ \midrule
\texttt{Quadratic} & \includegraphics[width=.2\linewidth,valign=m]{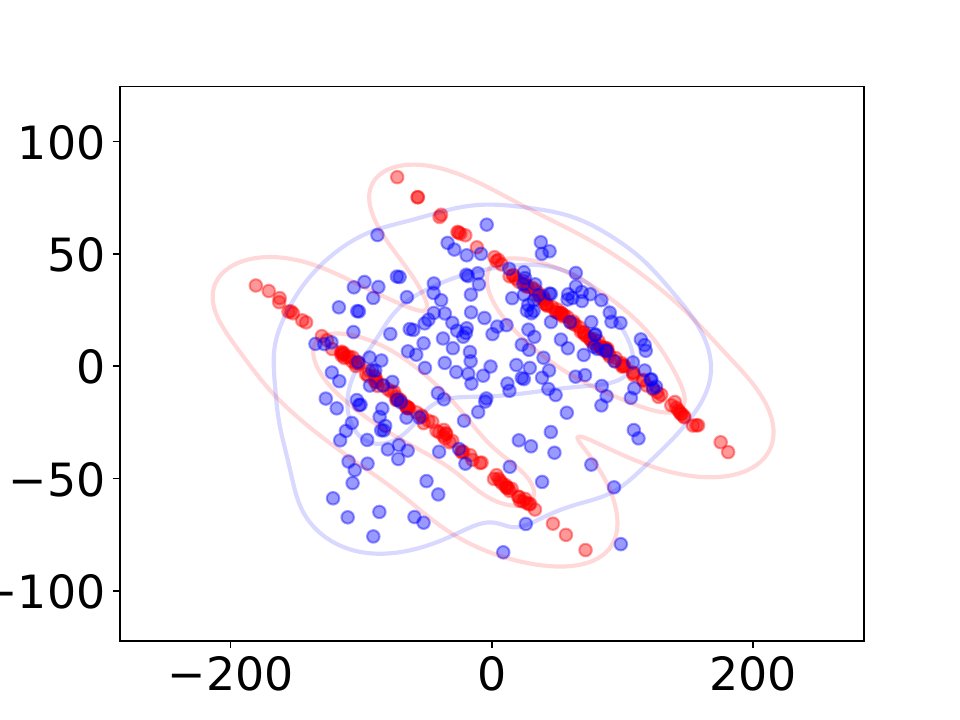} & \includegraphics[width=.2\linewidth,valign=m]{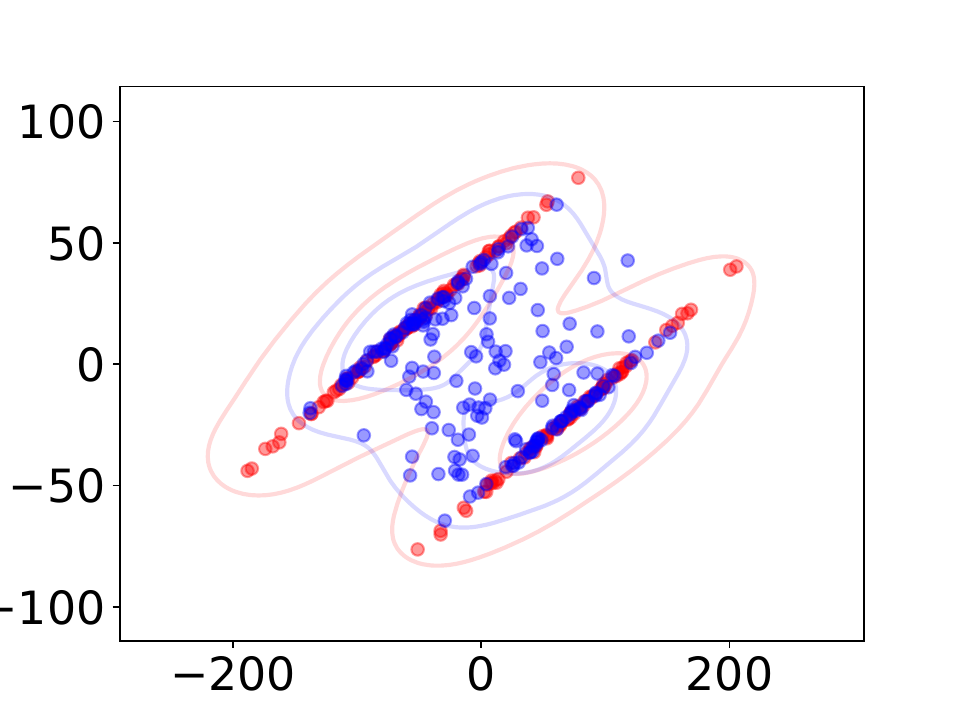} & \includegraphics[width=.2\linewidth,valign=m]{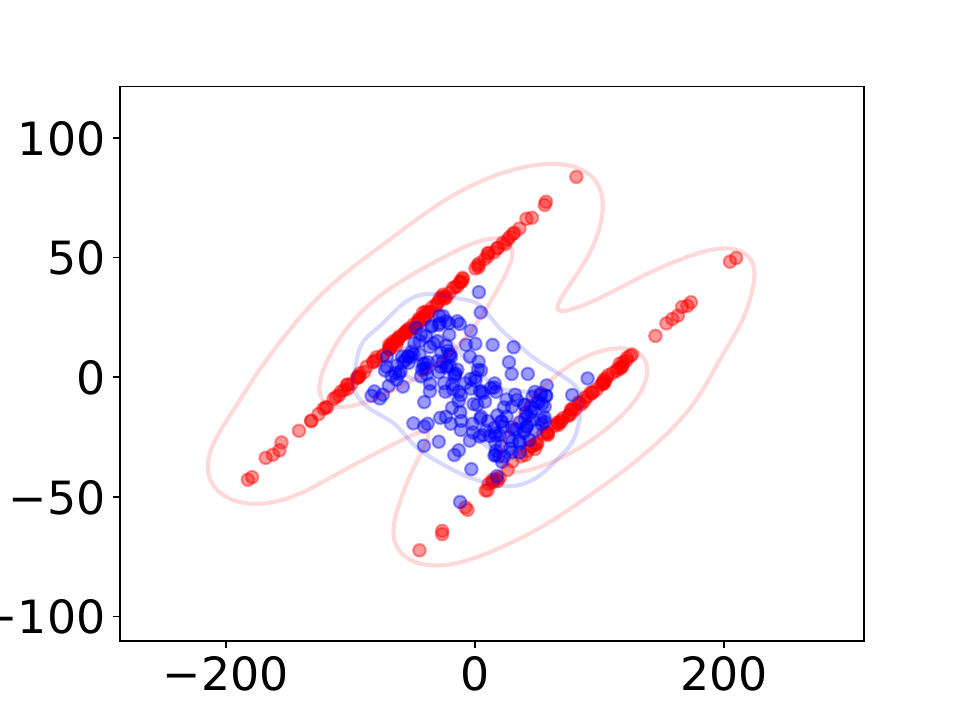} & \includegraphics[width=.2\linewidth,valign=m]{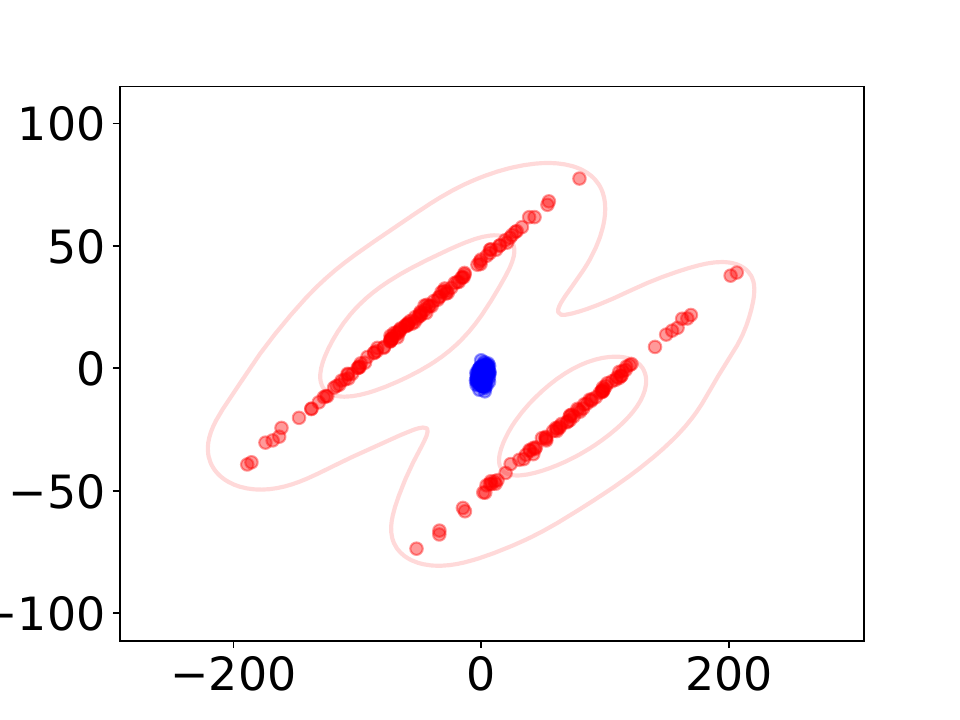}\\ 
\texttt{Melbourne}  & \includegraphics[width=.2\linewidth,valign=m]{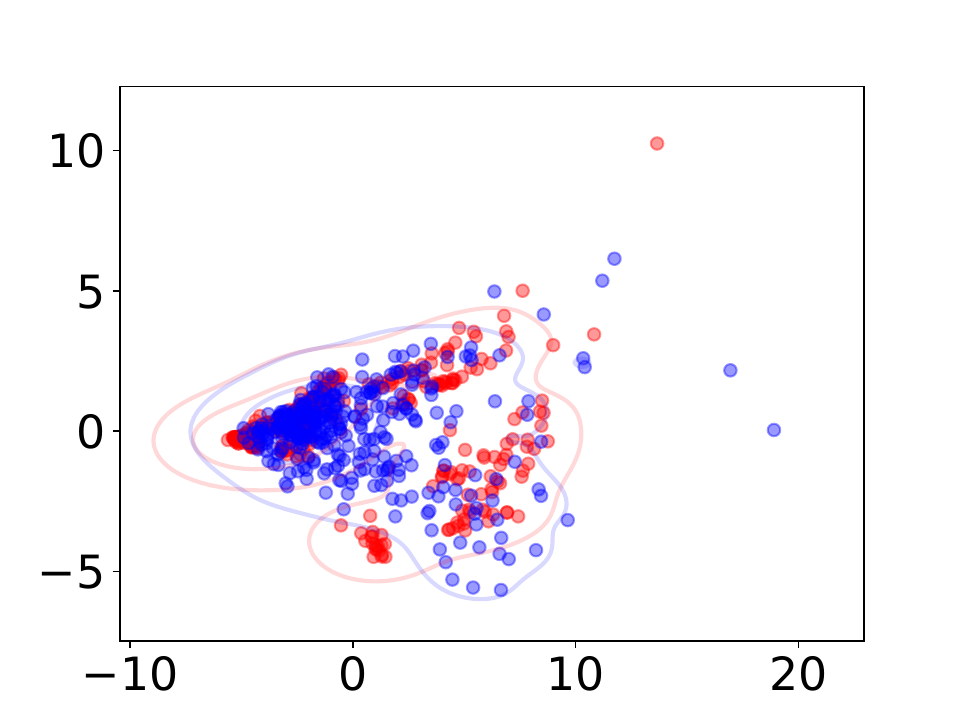} & \includegraphics[width=.2\linewidth,valign=m]{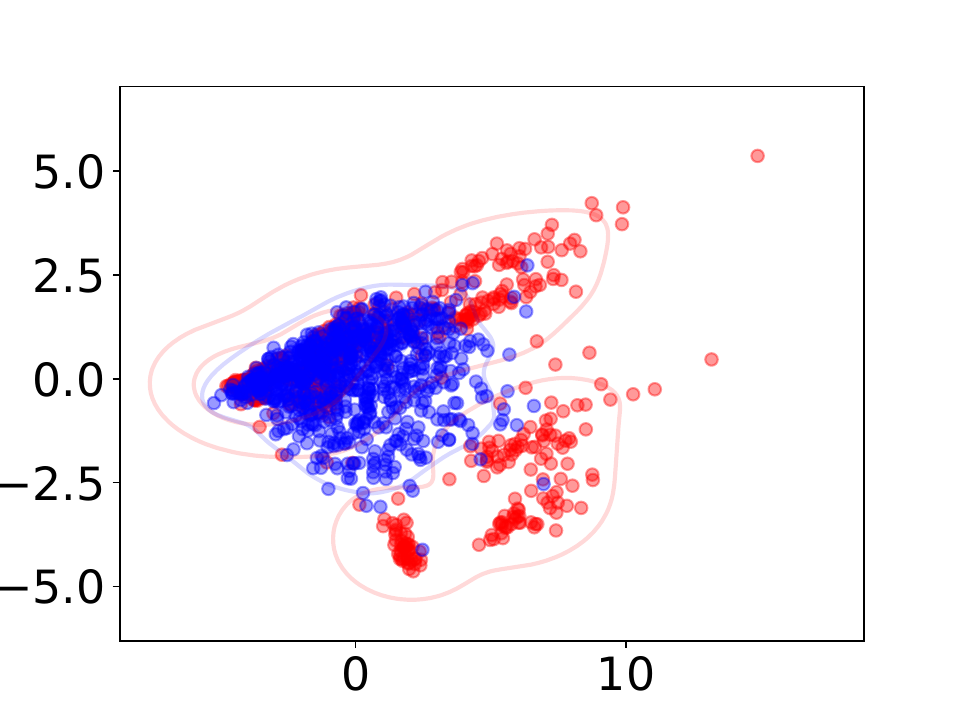} & \includegraphics[width=.2\linewidth,valign=m]{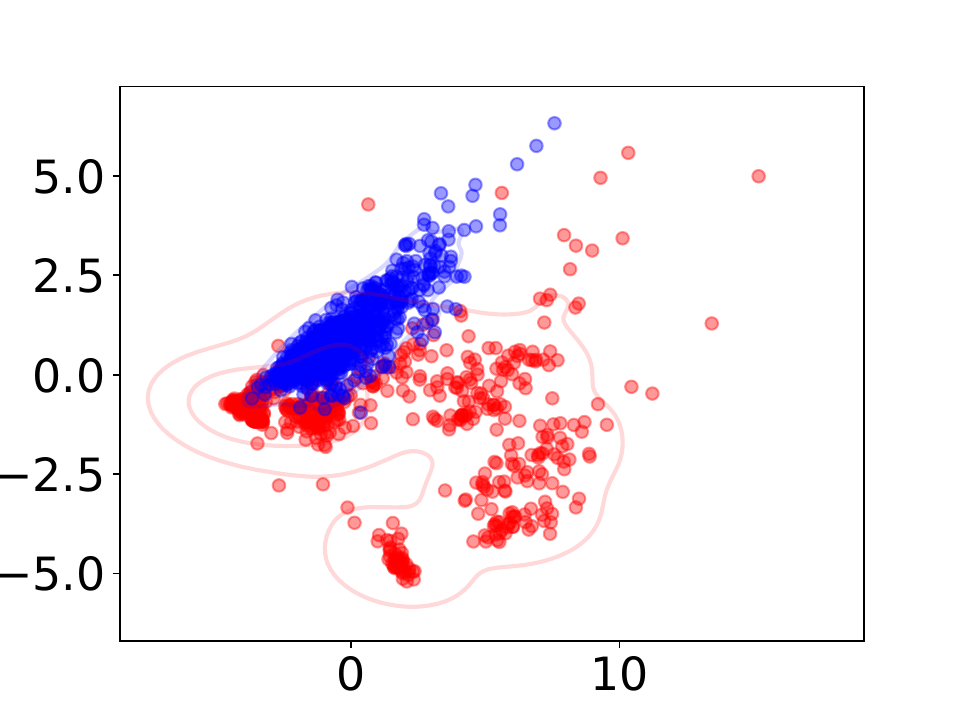}& \includegraphics[width=.2\linewidth,valign=m]{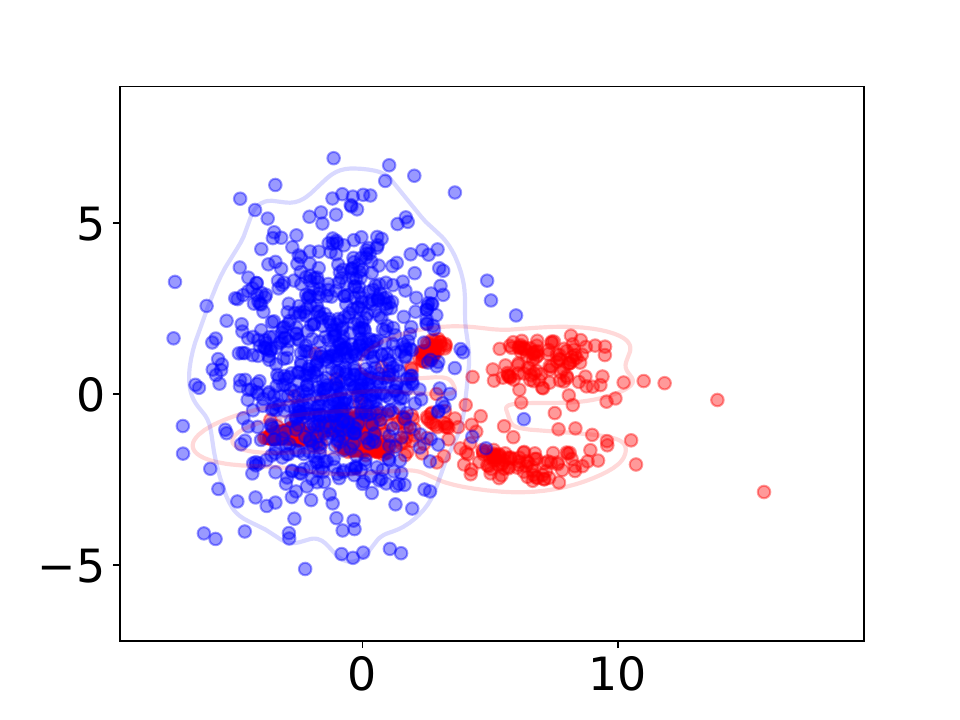}\\
\texttt{Stocks} & \includegraphics[width=.2\linewidth,valign=m]{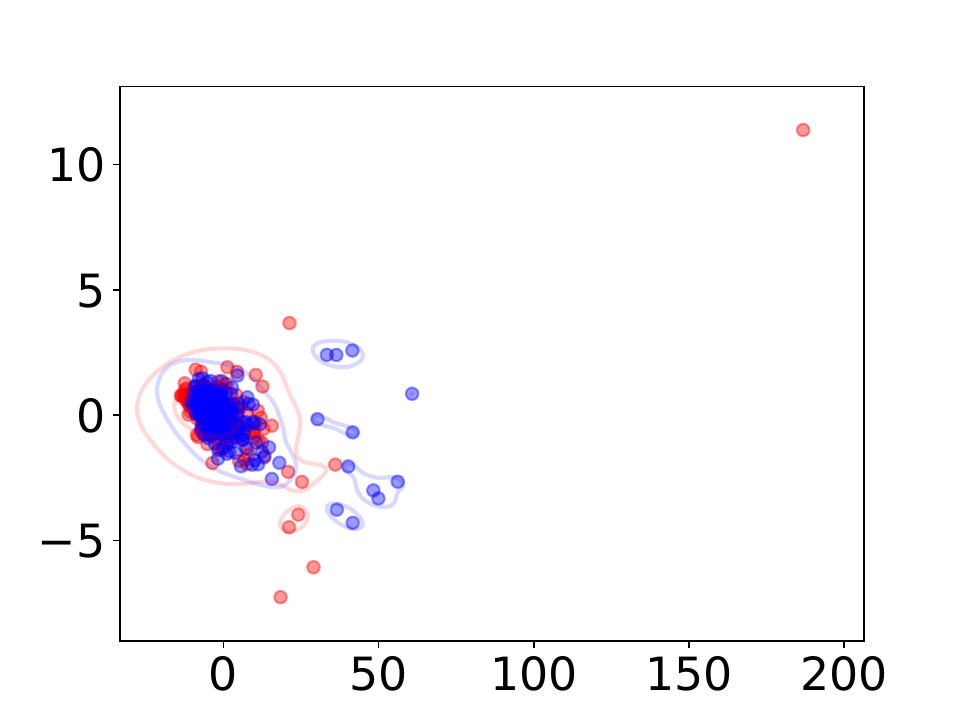} & \includegraphics[width=.2\linewidth,valign=m]{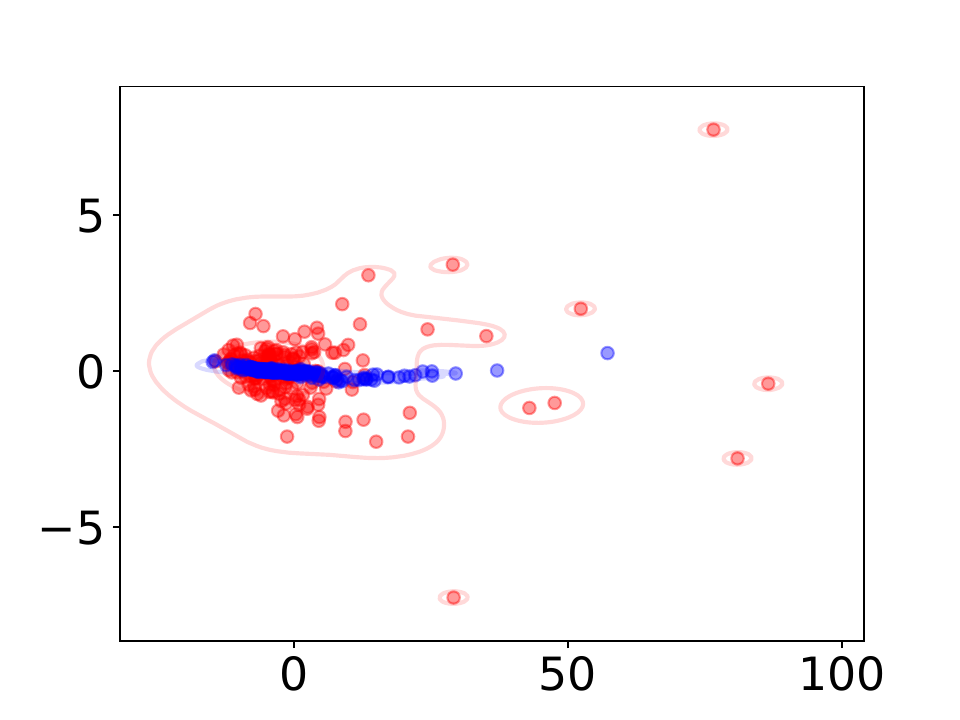} & \includegraphics[width=.2\linewidth,valign=m]{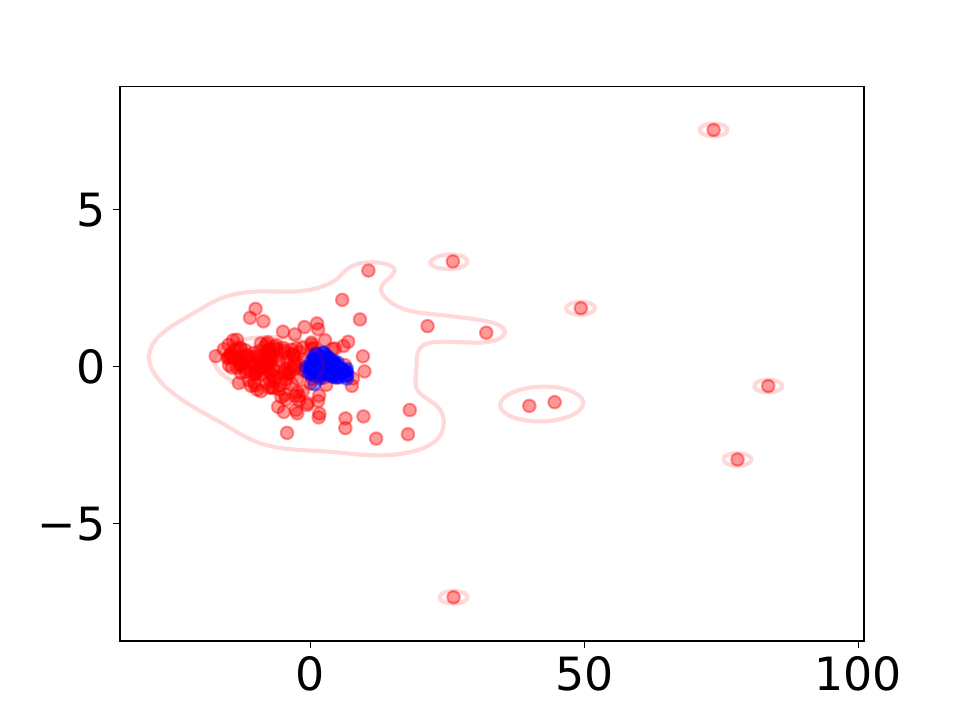}& \includegraphics[width=.2\linewidth,valign=m]{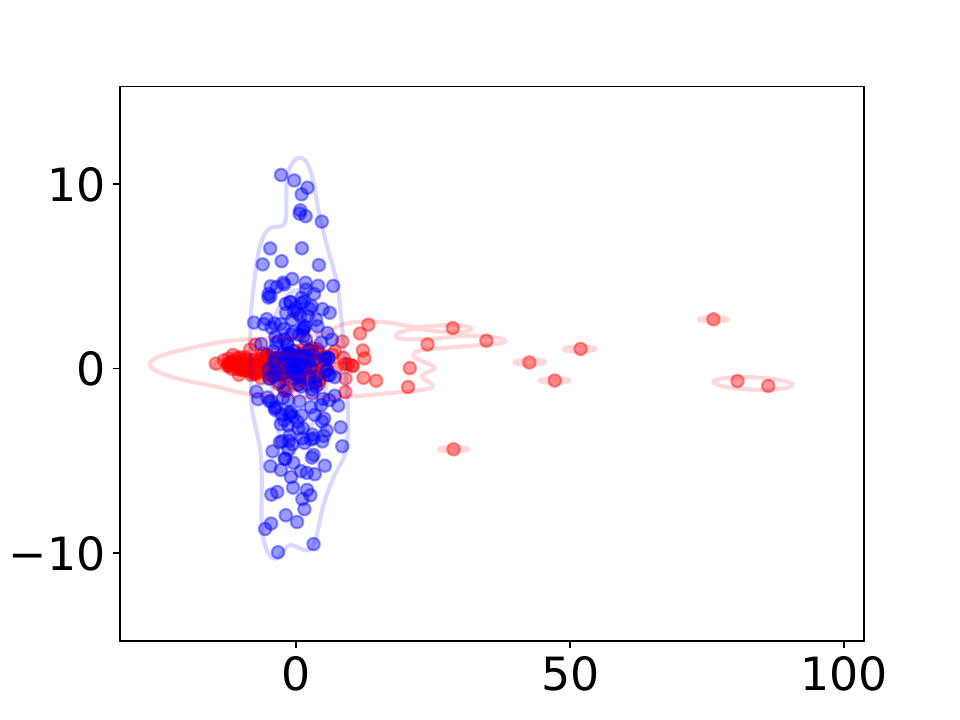}\\
\texttt{GridWatch}  & \includegraphics[width=.2\linewidth,valign=m]{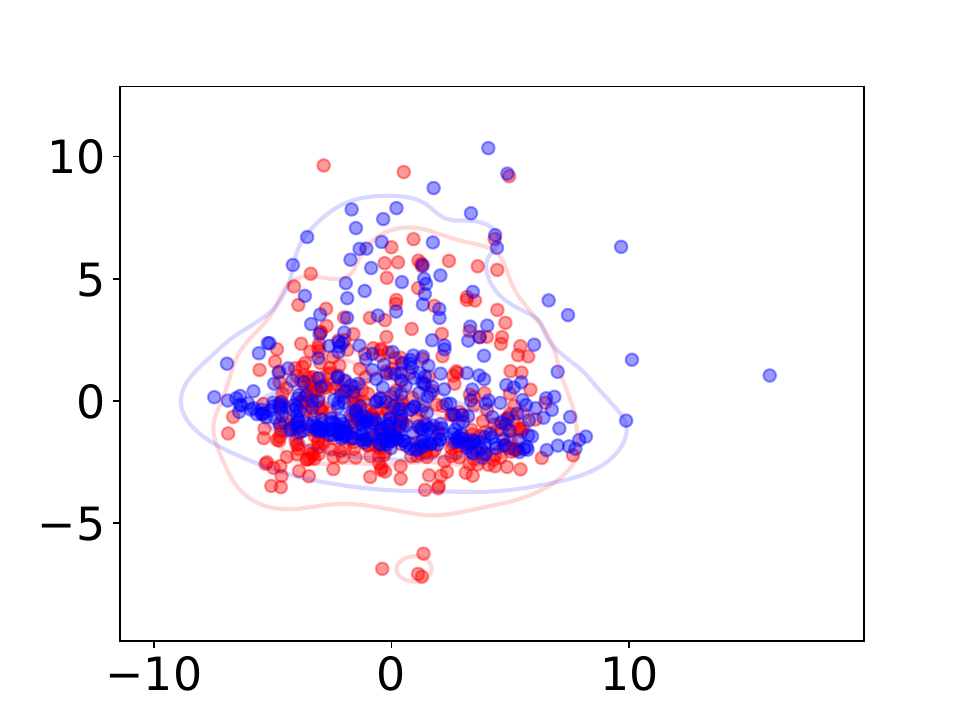} & \includegraphics[width=.2\linewidth,valign=m]{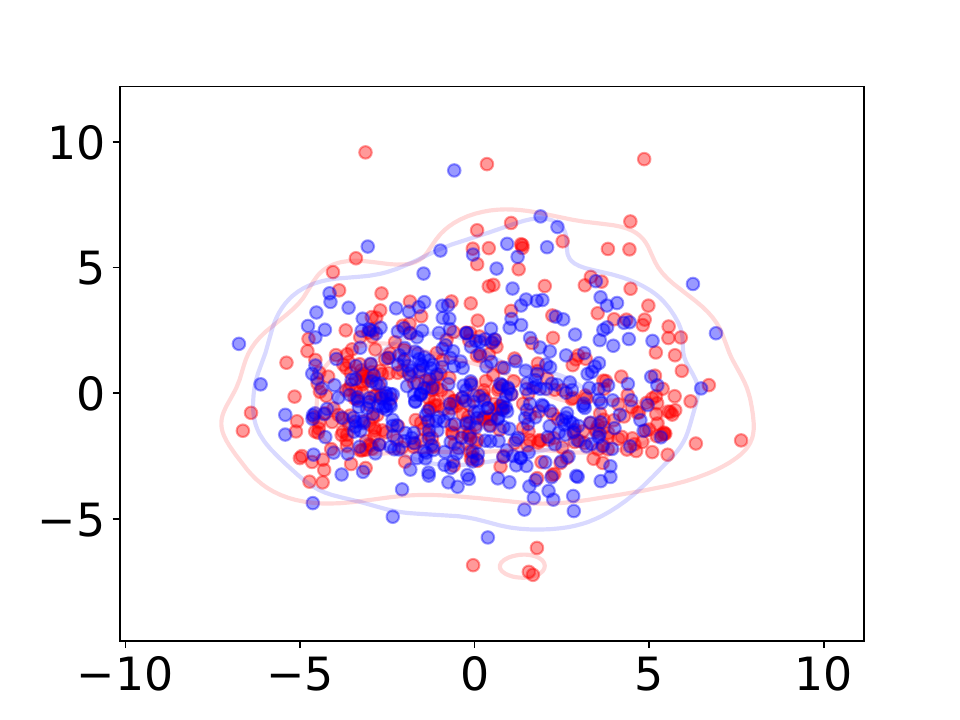} & \includegraphics[width=.2\linewidth,valign=m]{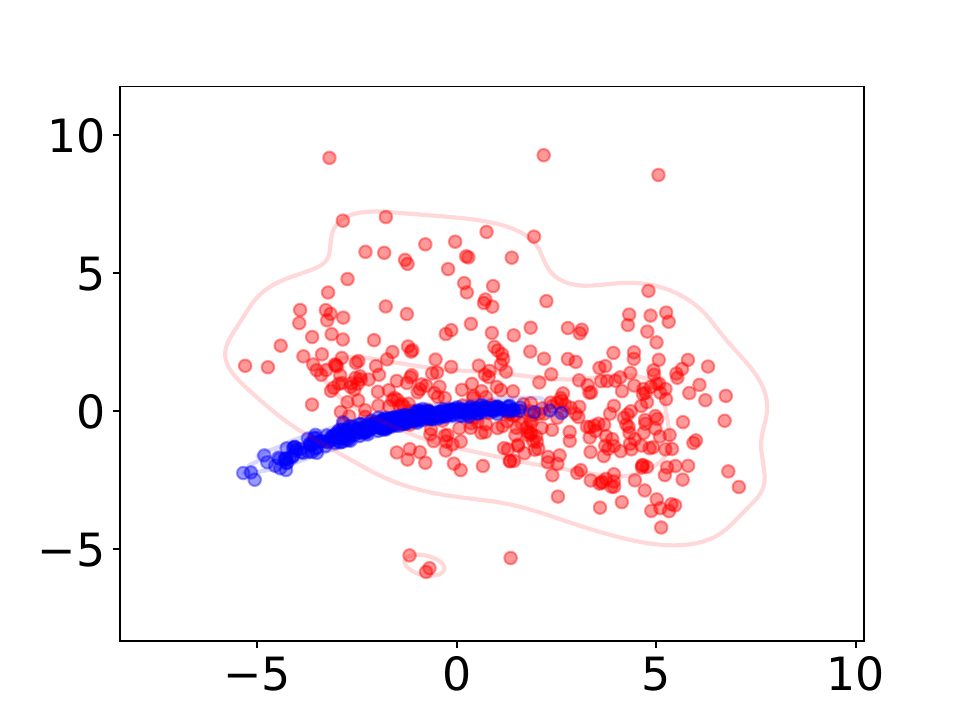} & \includegraphics[width=.2\linewidth,valign=m]{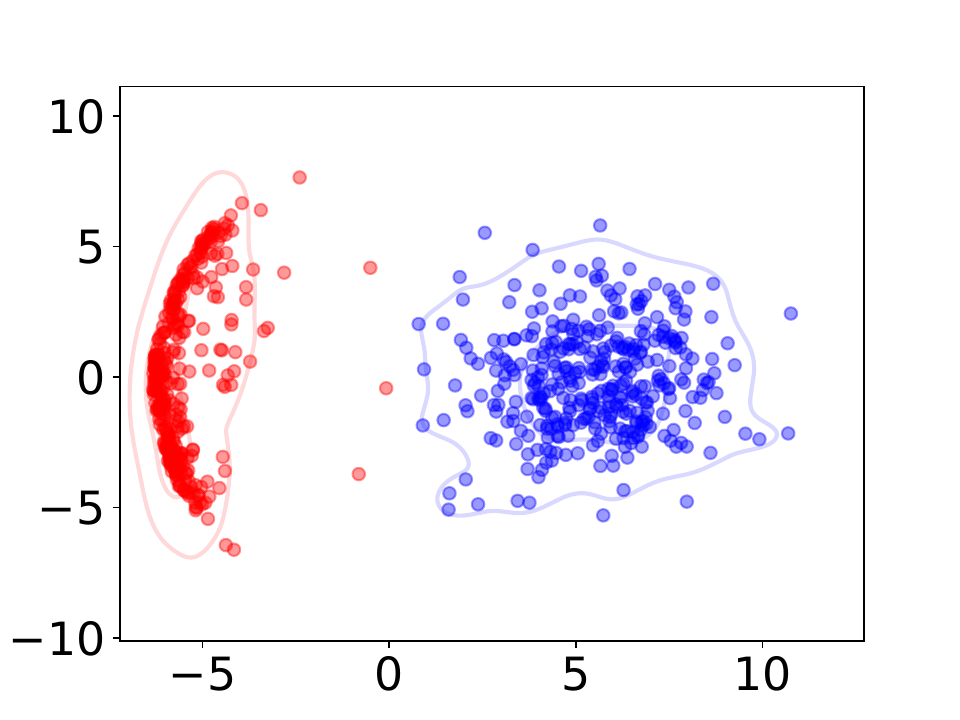}\\ \bottomrule
\end{tabular}
\label{tab:pca-samples-visualised}
\end{table}

\begin{table*}[!ht]
\centering
\caption{The t-SNE embeddings of $100$ samples of both samples generated from the model (\textcolor{blue}{blue}) and samples from the data (\textcolor{red}{red}).}
\makebox[\textwidth]{
\begin{tabular}{@{}ccccc@{}}
\toprule
\multicolumn{1}{l}{} & $\mathcal{F}$-EBM (ours)
& $\pi$-VAE
& NP                                                                                                                            & GP     \\ \midrule
\texttt{Quadratic}           & \includegraphics[width=.2\linewidth,valign=m]{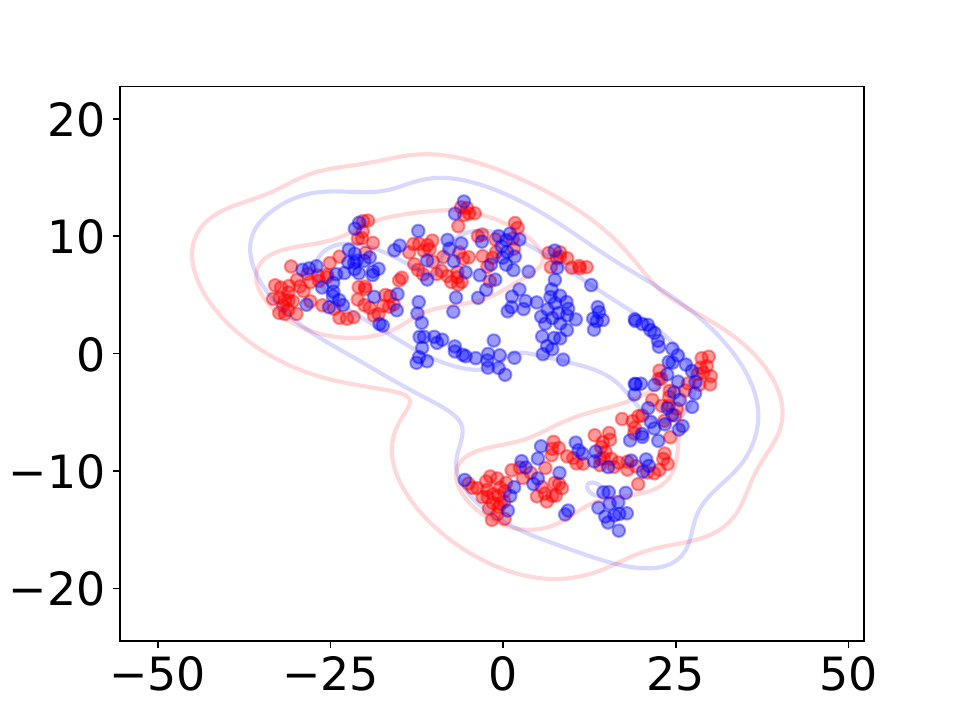}  &
\includegraphics[width=.2\linewidth,valign=m]{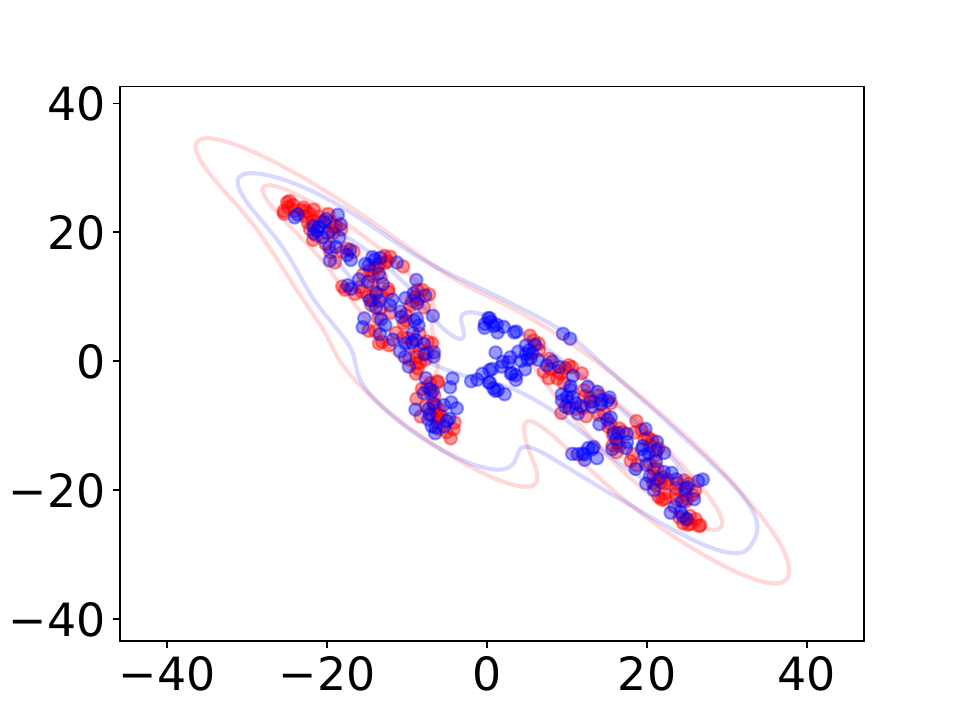}
& \includegraphics[width=.2\linewidth,valign=m]{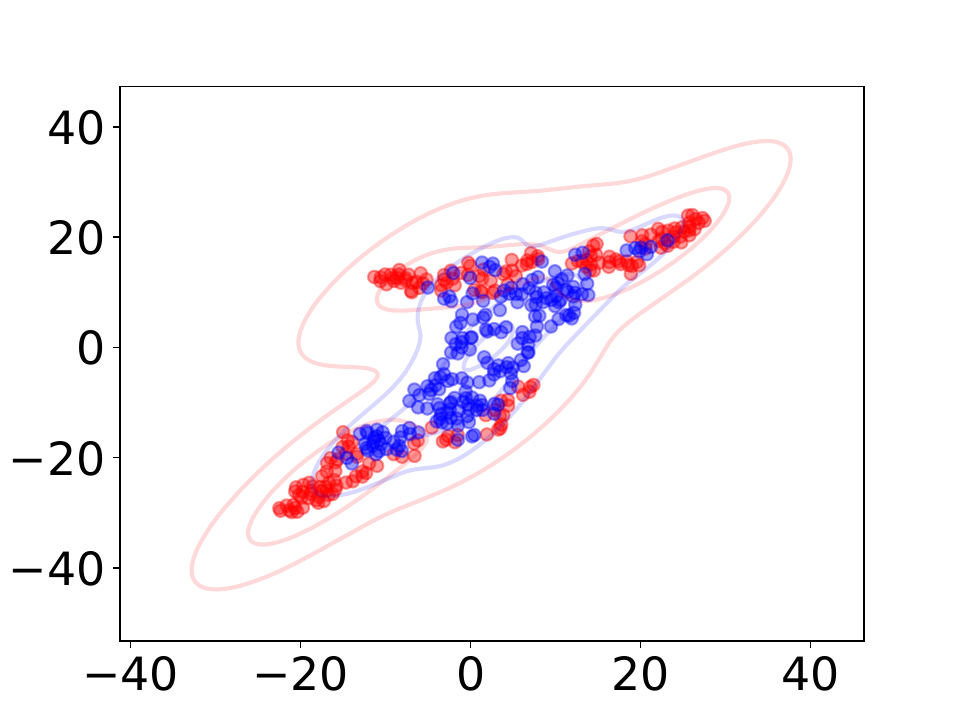}  & \includegraphics[width=.2\linewidth,valign=m]{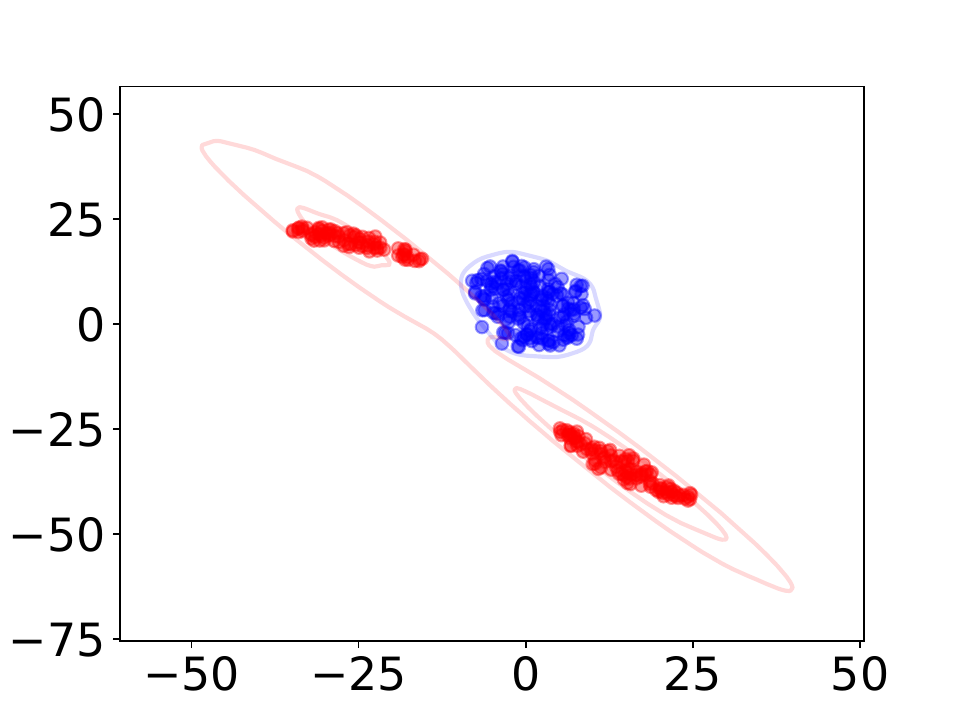}  \\
\texttt{Melbourne}          & \includegraphics[width=.2\linewidth,valign=m]{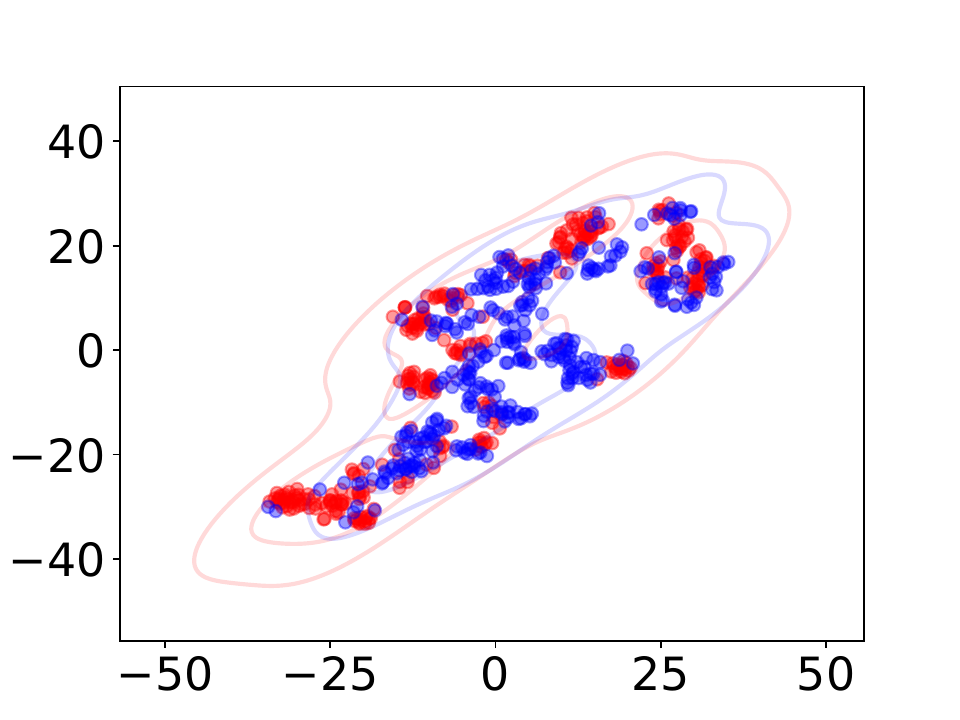} &
\includegraphics[width=.2\linewidth,valign=m]{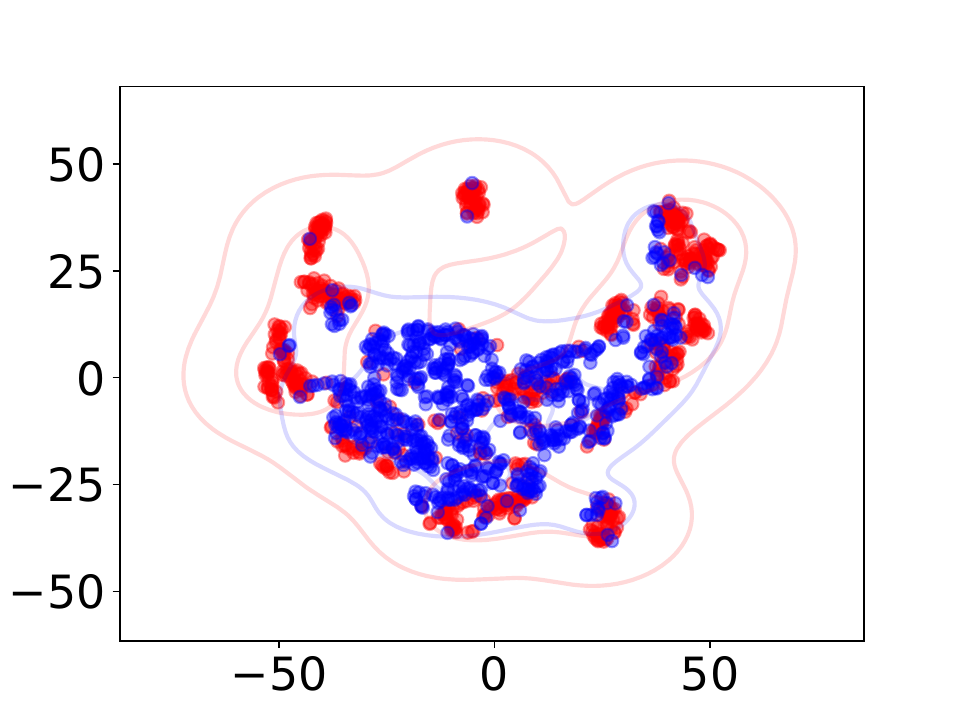}
&\includegraphics[width=.2\linewidth,valign=m]{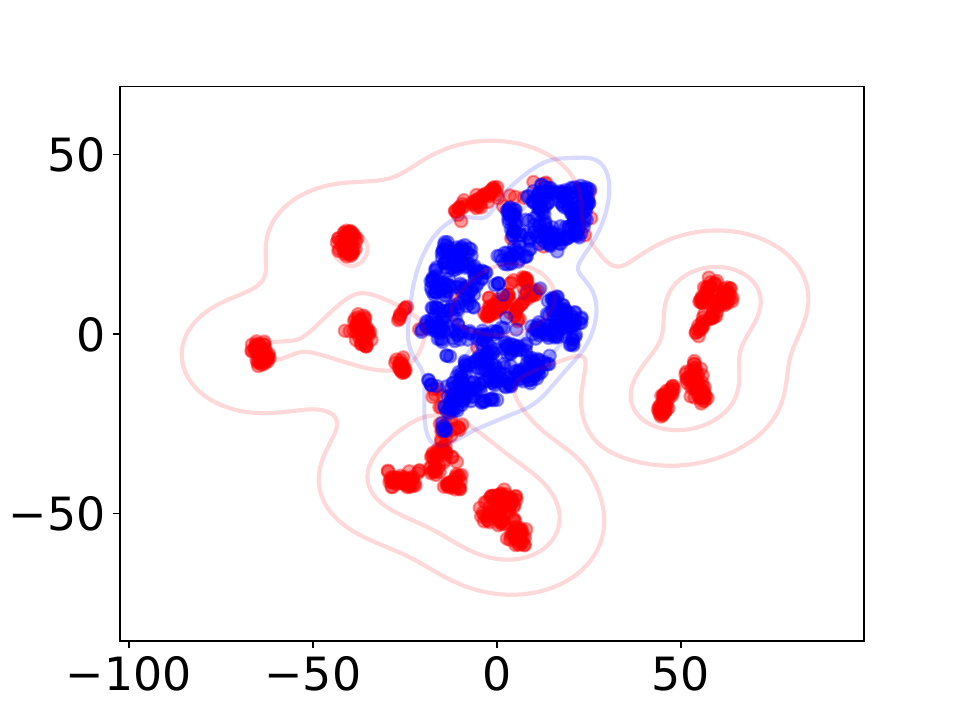} & \includegraphics[width=.2\linewidth,valign=m]{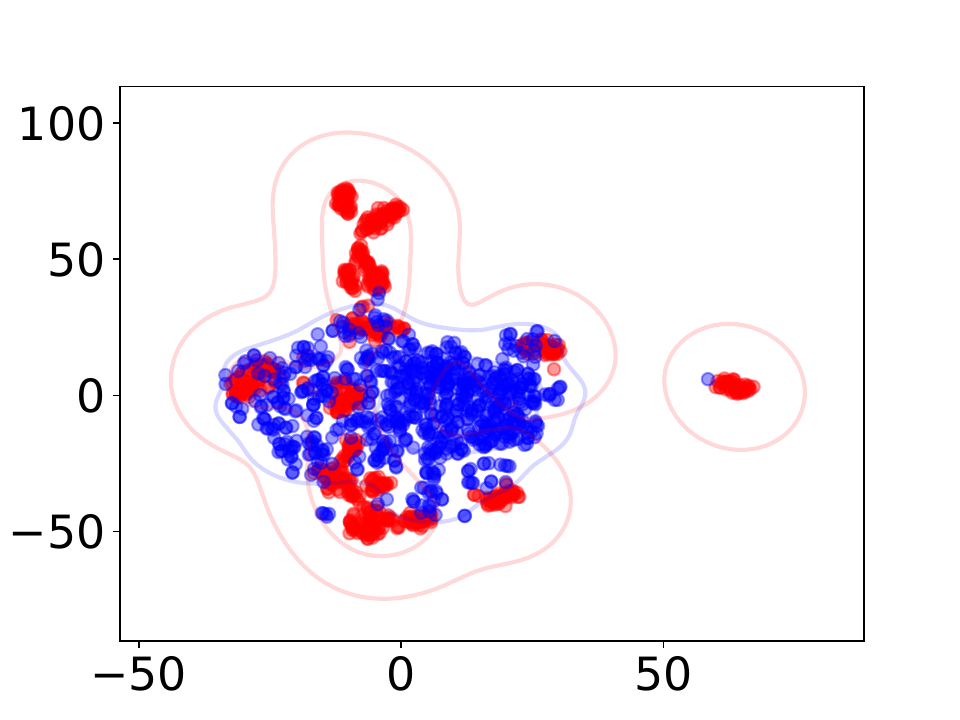} \\
\texttt{Stocks}                 & \includegraphics[width=.2\linewidth,valign=m]{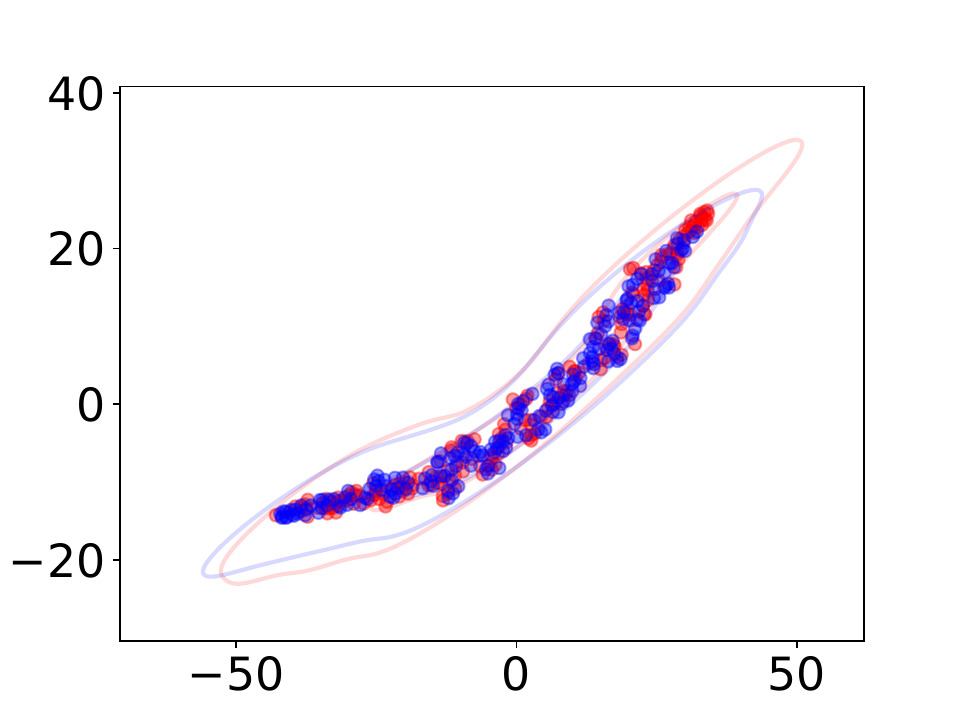}        &
\includegraphics[width=.2\linewidth,valign=m]{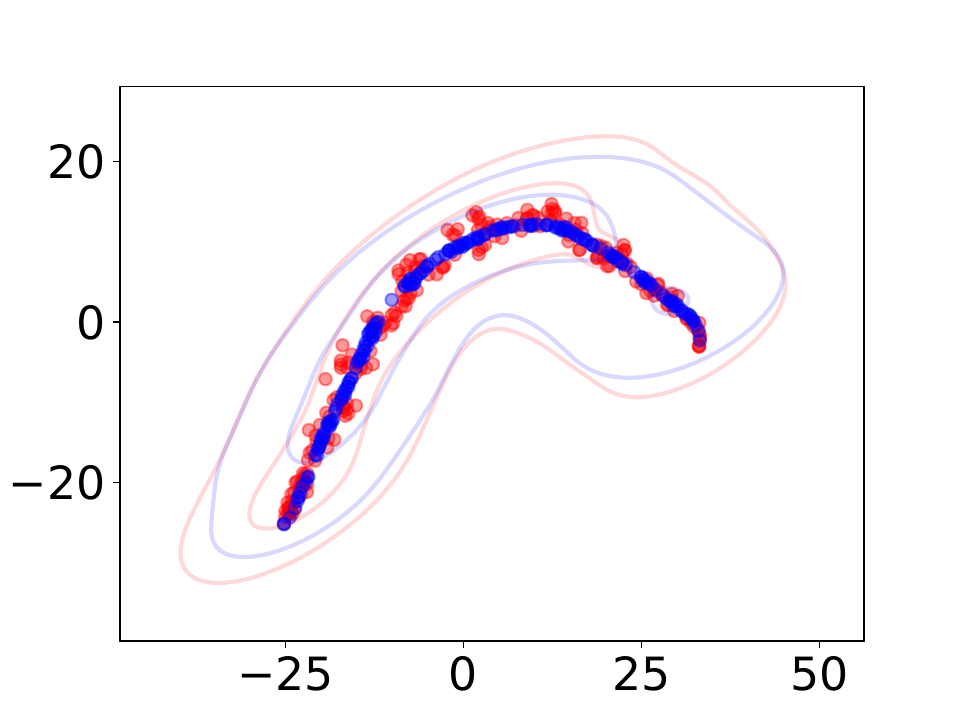}
&\includegraphics[width=.2\linewidth,valign=m]{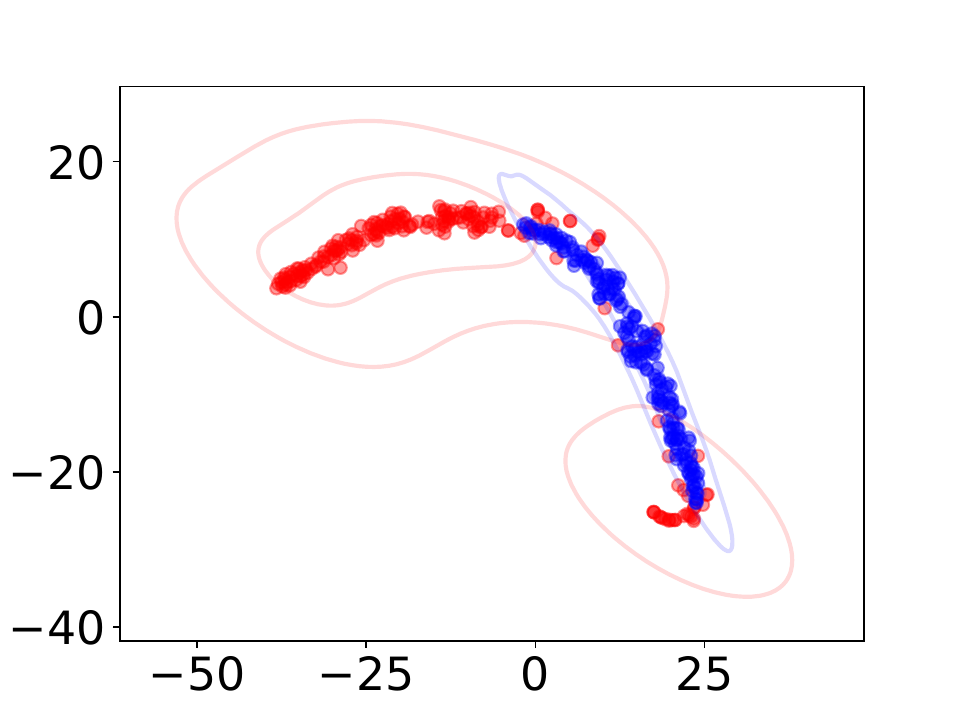}        & \includegraphics[width=.2\linewidth,valign=m]{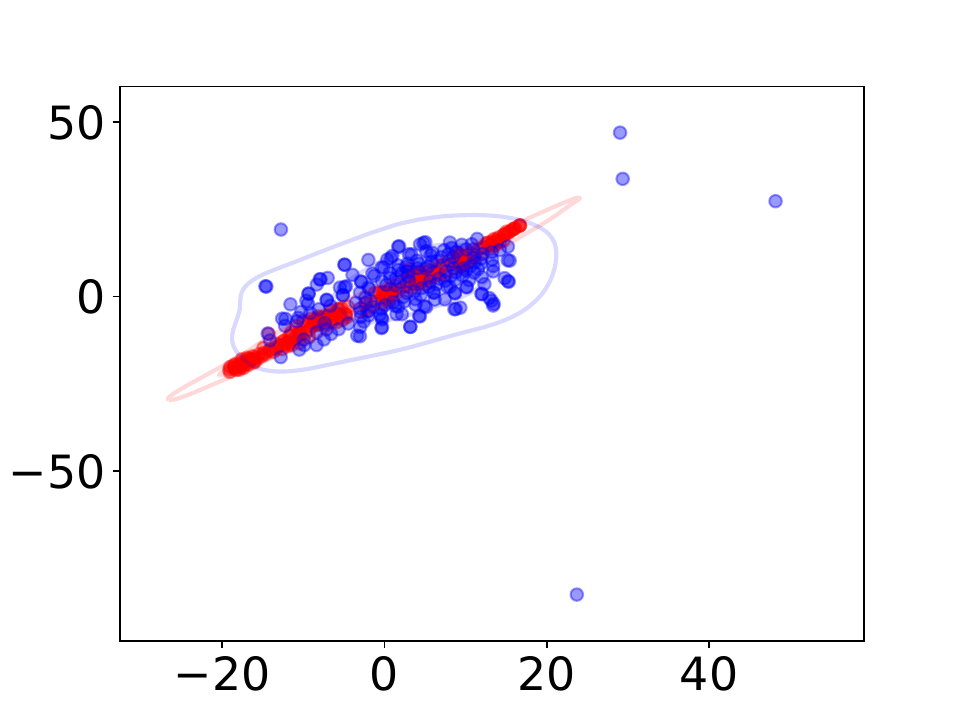}        \\
\texttt{GridWatch}            & \includegraphics[width=.2\linewidth,valign=m]{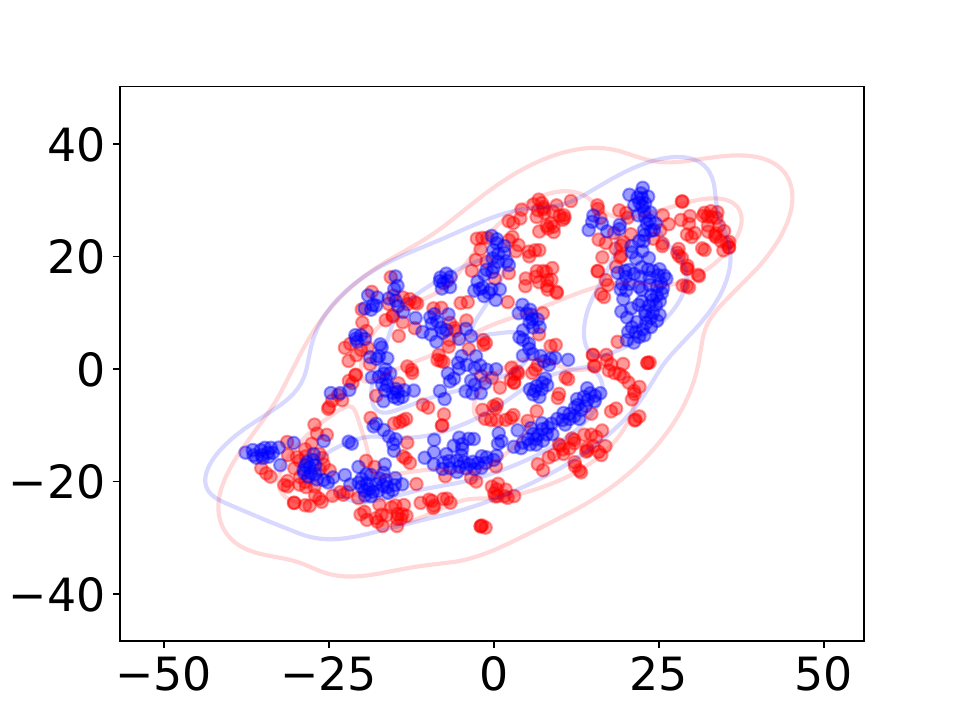}   & \includegraphics[width=.2\linewidth,valign=m]{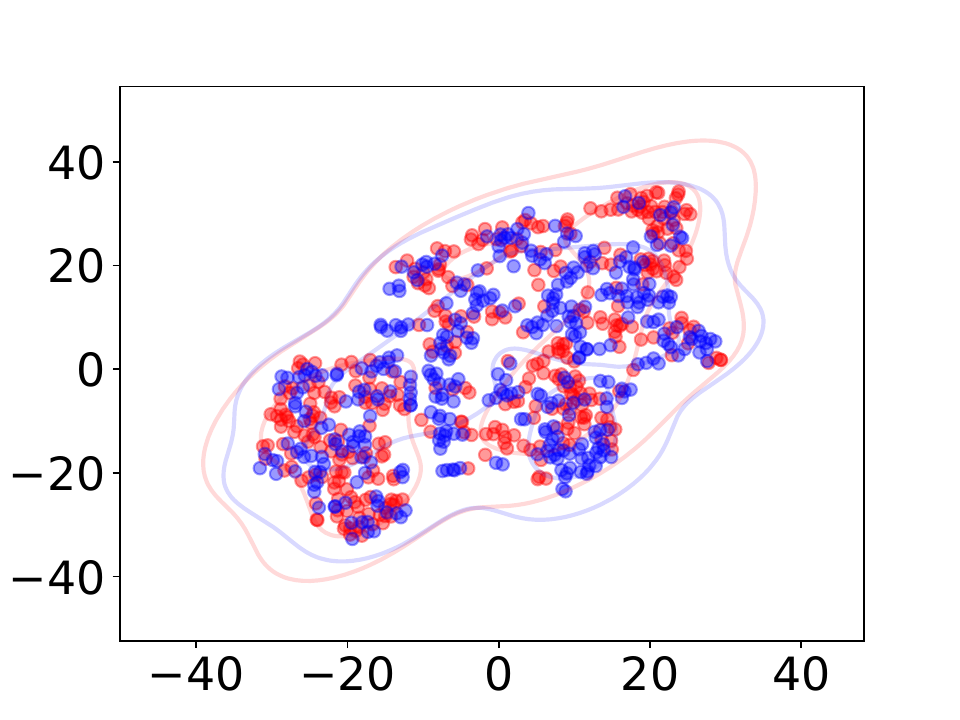}
&\includegraphics[width=.2\linewidth,valign=m]{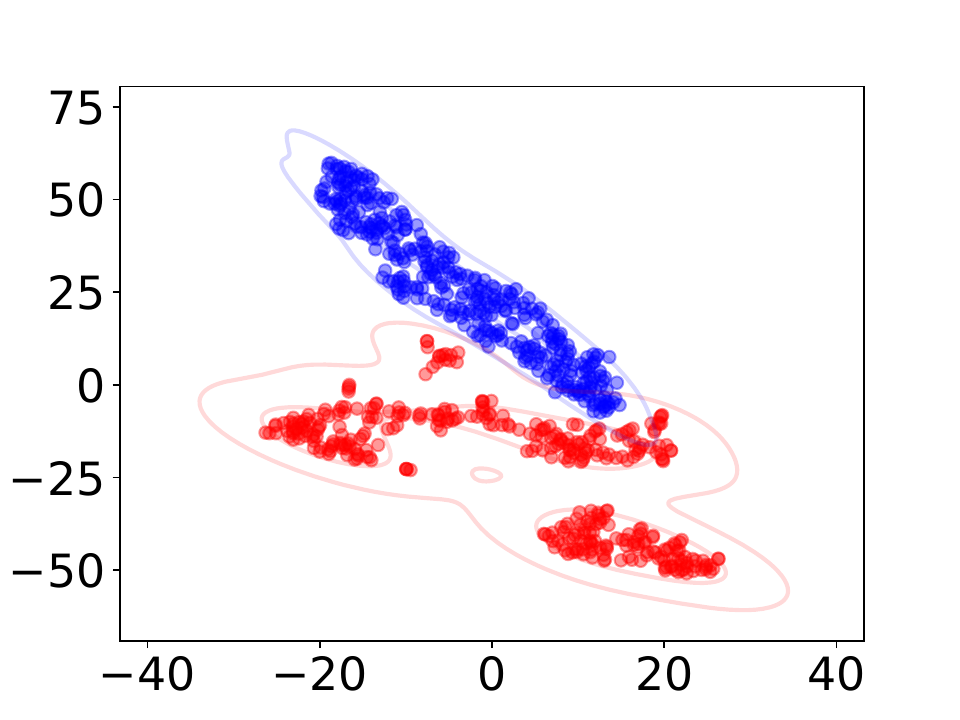}   & \includegraphics[width=.2\linewidth,valign=m]{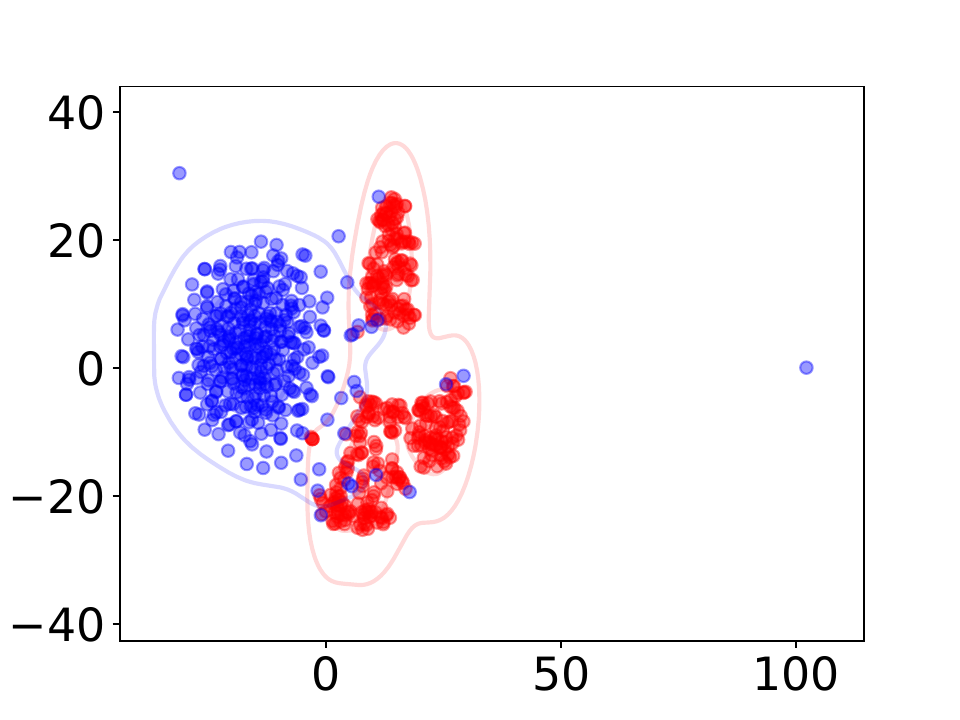}   \\ \bottomrule
\end{tabular}}
\label{tab:tsne-samples-visualised}
\end{table*}

\begin{figure}
    \includegraphics[width=\linewidth]{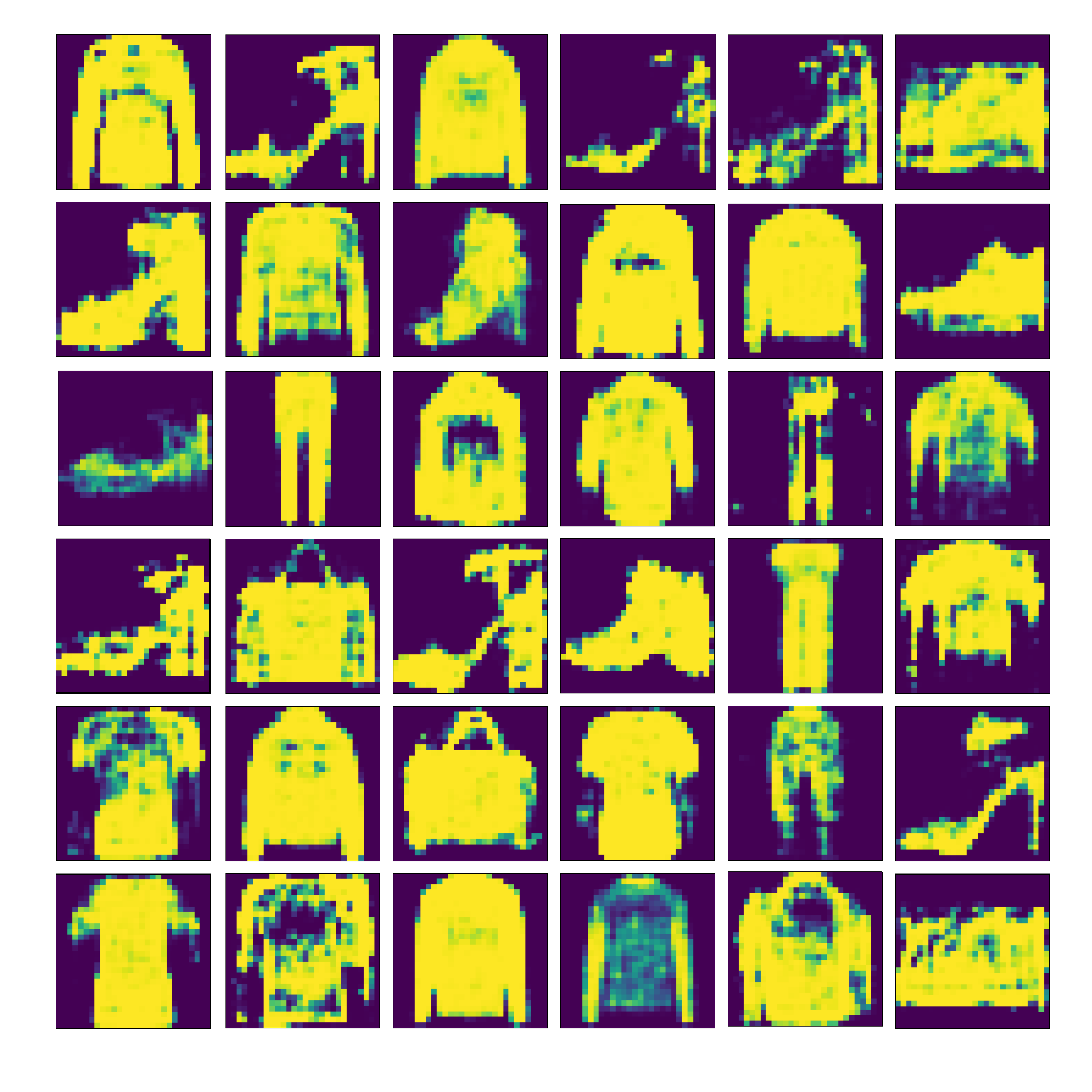}
    \caption{Samples from the $\mathcal{F}$-EBM.}
    \label{fig:mnist_samples}
\end{figure}
\end{document}